\documentclass{article}

\usepackage{microtype}
\usepackage{graphicx}
\usepackage{amsmath,amssymb}
\usepackage{color}
\usepackage{booktabs} 

\usepackage{caption}
\usepackage{subcaption}

\usepackage{hyperref}
\usepackage{color}

\newcommand{\Oh}{\mathcal{O}}

\newenvironment{itemize*}%
  {\begin{itemize}%
    \setlength{\itemsep}{0pt}%
    \setlength{\parskip}{0pt}}%
  {\end{itemize}}
  
\newenvironment{enumerate*}%
  {\begin{enumerate}%
    \vskip 0.1in%
    \setlength{\itemsep}{0pt}%
    \setlength{\parskip}{0pt}}%
  {\end{enumerate}%
   \vskip -0.1in}

\usepackage[accepted]{icml2019}

\icmltitlerunning{Understanding the Impact of Entropy on Policy Optimization}

\begin{document}

\twocolumn[
\icmltitle{Understanding the Impact of Entropy on Policy Optimization}



\icmlsetsymbol{equal}{*}

\begin{icmlauthorlist}
\icmlauthor{Zafarali Ahmed}{mila,intern}
\icmlauthor{Nicolas Le Roux}{mila,google}
\icmlauthor{Mohammad Norouzi}{google}
\icmlauthor{Dale Schuurmans}{google,uofa}
\end{icmlauthorlist}

\icmlaffiliation{mila}{Mila, McGill University, Montr{\'e}al, Canada}
\icmlaffiliation{intern}{Work done while at Google Research}
\icmlaffiliation{google}{Google Research}
\icmlaffiliation{uofa}{University of Alberta}

\icmlcorrespondingauthor{Zafarali Ahmed}{zafarali.ahmed@mail.mcgill.ca}

\icmlkeywords{Policy gradient, Entropy regularization, Optimization landscape, Reinforcement Learning, ICML}

\vskip 0.3in
]



\printAffiliationsAndNotice{}  

\begin{abstract}
Entropy regularization is commonly used to improve policy optimization in reinforcement learning. It is believed to help with \emph{exploration} by encouraging the selection of more stochastic policies. In this work, we analyze this claim using new visualizations of the optimization landscape based on randomly perturbing the loss function. We first show that even with access to the exact gradient, policy optimization is difficult due to the geometry of the objective function. We then qualitatively show that in some environments, a policy with higher entropy can make the optimization landscape smoother, thereby connecting local optima and enabling the use of larger learning rates. This paper presents new tools for understanding the optimization landscape, shows that policy entropy serves as a regularizer, and highlights the challenge of designing general-purpose policy optimization algorithms.
\end{abstract}

\section{Introduction}
Policy optimization is a family of reinforcement learning (RL) algorithms aiming to directly optimize the parameters of a policy by maximizing discounted cumulative rewards.
This often involves a difficult non-concave maximization problem, even when using a simple policy with a linear state-action mapping. 

Contemporary policy optimization algorithms build upon the \textsc{Reinforce} algorithm \citep{williams1992simple}. These algorithms involve estimating a noisy gradient of the optimization objective using Monte-Carlo sampling to enable stochastic gradient ascent. This estimate can suffer from high variance and several solutions have been proposed to address what is often seen as a major issue  \citep{konda2000actor,greensmith2004variance,schulman2015high,tucker2018mirage}. 

However, in this work we show that noisy estimates of the gradient are not necessarily the \emph{main} issue: The optimization problem is difficult because of the geometry of the landscape. Given that ``high variance'' is often the reason given for the poor performance of policy optimization, it raises an important question: \textit{How do we study the effects of different policy learning techniques on the underlying optimization problem?}

An answer to this question would guide future research directions and drive the design of new policy optimization techniques. Our work makes progress toward this goal by taking a look at one such technique: \emph{entropy regularization}.

In RL, exploration is critical to finding good policies during optimization: If the optimization procedure does not sample a large number of diverse state-action pairs, it may converge to a poor policy. To prevent policies from becoming deterministic too quickly, researchers use entropy regularization \citep{williams1991function,mnih2016asynchronous}. Its success has sometimes been attributed to the fact that it ``encourages exploration'' \citep{mnih2016asynchronous,schulman2017equivalence,schulman2017proximal}. Contrary to Q-learning \citep{watkins1992q} or Deterministic Policy Gradient \citep{silver2014deterministic} where the exploration is handled separately from the policy itself, direct policy optimization relies on the stochasticity of the policy being optimized for the exploration. However, policy optimization is a pure maximization problem and any change in the policy is reflected in the objective. Hence, any strategy, such as entropy regularization, can only affect learning in one of two ways: either it reduces the noise in the gradient estimates or it changes the optimization landscape.

In this work we investigate some of these questions by controlling the entropy of policies and observing its effect on the geometry of the optimization landscape. This work contains the following contributions:
\begin{itemize}
    \item{We show experimentally that the difficulty of policy optimization is strongly linked to the geometry of the objective function.}
    \item{We propose a novel visualization of the objective function that captures local information about gradient and curvature.}
    \item{We show experimentally that policies with higher entropy induce a smoother objective that connects solutions and enable the use of larger learning rates.}
\end{itemize}

\section{Approach}
We take here the view that improvements due to entropy regularization might be attributed to having a better objective landscape. In Section~\ref{approach_obj_funcs} we introduce tools to investigate landscapes in the context of general optimization problems. In Section~\ref{appr:pertubations} we propose a new visualization technique to understand high dimensional optimization landscapes. We will then explain the RL policy optimization problem and entropy regularization in Section~\ref{approach_policy_opt}.

\subsection{Understanding the Landscape of Objective Functions}
\label{approach_obj_funcs}
We explain our experimental techniques by considering the general optimization problem and motivating the relevance of studying objective landscapes. We are interested in finding parameters $\theta\in\mathbb{R}^n$ that maximize an objective function, $\Oh : \mathbb{R}^n\rightarrow \mathbb{R}$, denoted $\theta^*=\arg\max_\theta~\Oh(\theta)$. The optimization algorithm takes the form of gradient ascent: $\theta_{i+1}=\theta_i+\eta_i\nabla_\theta \Oh$, where $\eta_i$ is the learning rate, $\nabla_\theta \Oh$ is the gradient of $\Oh$ and $i$ is the iteration number.

\emph{Why should we study objective landscapes?} The ``difficulty'' of this optimization problem is given by the properties of $\Oh$. For example, $\Oh$ might have kinks and valleys making it difficult to find good solutions from different initial parameters \citep{li2017visualizing}. Similarly, if $\Oh$ contains very flat regions, optimizers like gradient ascent can take a very long time to escape them \citep{dauphin2014identifying}. Alternatively, if the curvature of $\Oh$ changes rapidly with every $\theta_i$, then it will be difficult to choose a stepsize. 

In the subsequent subsections, we describe two effective techniques for visualization of the optimization landscapes.

\subsubsection{Linear Interpolations}
\label{appr:interpolations}
One approach to visualize an objective function is to interpolate $\theta$ in the 1D subspace between two points $\theta_0$ and $\theta_1$ \citep{chapelle2010gradient,goodfellow2014qualitatively} by evaluating the objective at $\Oh((1-\alpha)\theta_0 + \alpha\theta_1)$ for $0\leq\alpha\leq1$. 
Such visualizations can tell us about the existence of valleys or monotonically increasing paths of improvement between the parameters. Typically $\theta_0$ and $\theta_1$ are initial parameters or solutions obtained through the optimization.

Though this technique provides interesting visualizations, conclusions are limited to the 1D slice: \citet{draxler2018essentially} show that even though the local optima are isolated in the 1D slice, these local optima can be connected by a manifold of equal value.
Hence, we must be careful to conclude general properties about the landscape using this visualization. In the next section, we describe a new visualization technique that, together with linear interpolations, can serve as a powerful tool for landscape analysis.

\subsubsection{Objective Function Geometry using Random Perturbations}
\label{appr:pertubations}
To overcome some of the limitations described in Section~\ref{appr:interpolations}, we develop a new method to locally characterize the properties of $\Oh$. In particular, we use this technique to (1) classify points in the parameter space as local optimum, saddle point, or flat regions; and (2) measure curvature of the objective during optimization.

To understand the local geometry of $\Oh$ around a point $\theta_0$ we sample directions $d$ uniformly at random on the unit ball. We then probe how $\Oh$ is changing along the sampled direction by evaluating at a pair of new points: $\theta^+_d=\theta_0 + \alpha d$ and $\theta^-_d=\theta_0 - \alpha d$ for some value $\alpha$. After collecting multiple such samples and calculating the change for each pair with respect to the initial point, $\Delta^{\Oh^+}_d = \Oh(\theta^+_d) - \Oh(\theta_0)$ and $\Delta^{\Oh^-}_d = \Oh(\theta^-_d)-\Oh(\theta_0)$, we can then classify a point $\theta_0$ according to:
\begin{enumerate}
    \item{If $\Delta^{\Oh+}_d < 0$ and $\Delta^{\Oh-}_d < 0$ for all $d$, $\theta_0$ is a local maximum.}
    \item{If $\Delta^{\Oh+}_d > 0$ and $\Delta^{\Oh-}_d > 0$ for all $d$, $\theta_0$ is a local minimum.}
    \item{If $\Delta^{\Oh+}_d \approx - \Delta^{\Oh-}_d$, $\theta_0$ is in an almost linear region.}
    \item{If $\Delta^{\Oh+}_d \approx \Delta^{\Oh-}_d \approx 0$, $\theta_0$ is in an almost flat region.}
\end{enumerate}
In practice, since we only sample a finite number of directions, we can only reason about the probability of being in such a state. Further, we can also observe a combination of these pairs, for instance in the case of saddle points\footnote{This list is not exhaustive and one can imagine detecting many more scenarios}.

\begin{figure}[t]
    \vskip 0.2in
    \begin{center}
    \begin{subfigure}[b]{0.235\textwidth}
        \centering
        \includegraphics[trim={0.75cm 0.75cm 0.75cm 0.75cm},clip,width=\textwidth]{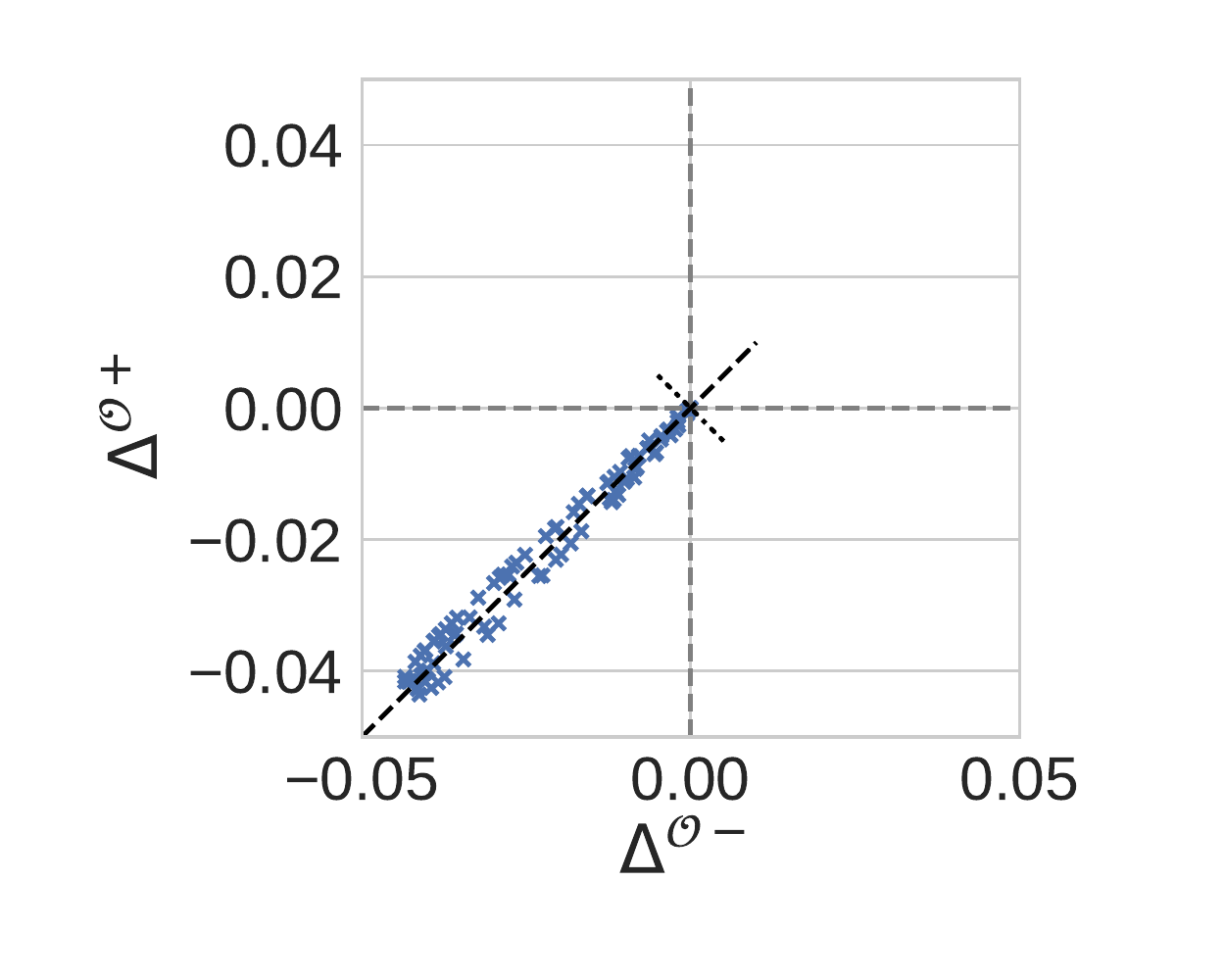}
        \caption{Apparent local optimum}
    \end{subfigure}
    \begin{subfigure}[b]{0.235\textwidth}
        \centering
        \includegraphics[trim={0.75cm 0.75cm 0.75cm 0.75cm},clip,width=\textwidth]{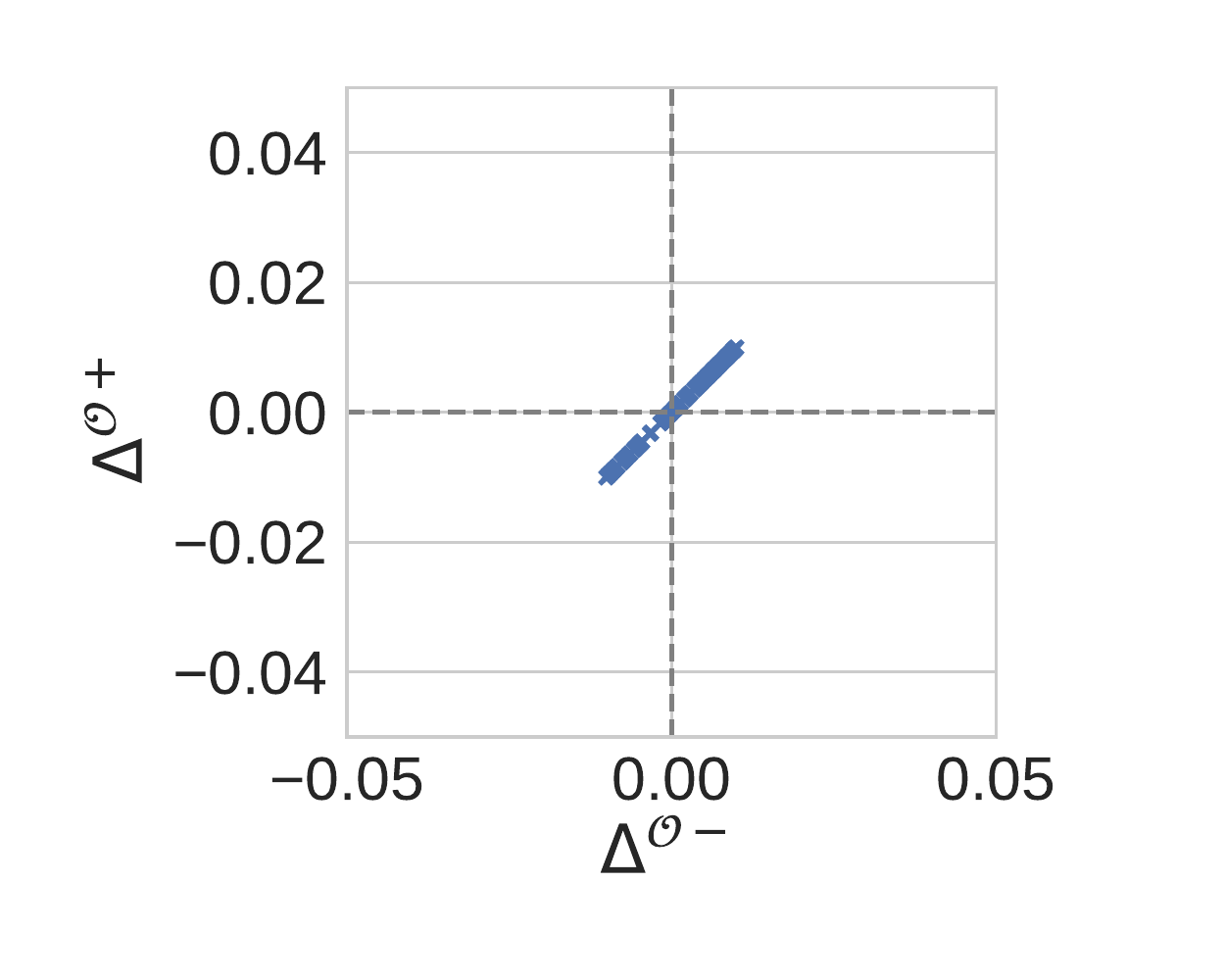}
        \caption{Saddle point}
    \end{subfigure}
    \vskip -0.1in
    \caption{\small \textbf{Demonstration of random perturbation technique.} 
    (a) If all perturbations are strictly negative, it implies that the point is likely a local optima. Given that some perturbations evaluate to 0, suggests some flat directions alluding to the connected manifold (b) If both perturbations are positive or both are negative it implies that the point is a saddle. See Section~\ref{approach_obj_funcs} for detailed explanation.}
    \label{fig:random_pertubations_examples}
    \end{center}
    \vskip -0.2in
\end{figure}

As an example, consider $\Oh(\theta)=-(1-\theta_0\theta_1)^2$ that has a saddle point and a manifold of local optima \citep{goodfellow2014qualitatively}. The proposed technique can distinguish between the local optimum at $\theta=(-0.5, -2)$ and saddle point at $\theta=(0,0)$ (Figure~\ref{fig:random_pertubations_examples}). Other examples on simple quadratics are shown in Figure~\ref{sfig:LM_example_visualizations}~and~\ref{sfig:linear_example_visualizations}.

\begin{figure}[t]
    \vskip 0.2in
    \begin{center}
    \begin{subfigure}[b]{0.235\textwidth}
        \centering
        \includegraphics[trim={0.55cm 0.55cm 0.55cm 0.55cm},clip,width=\textwidth]{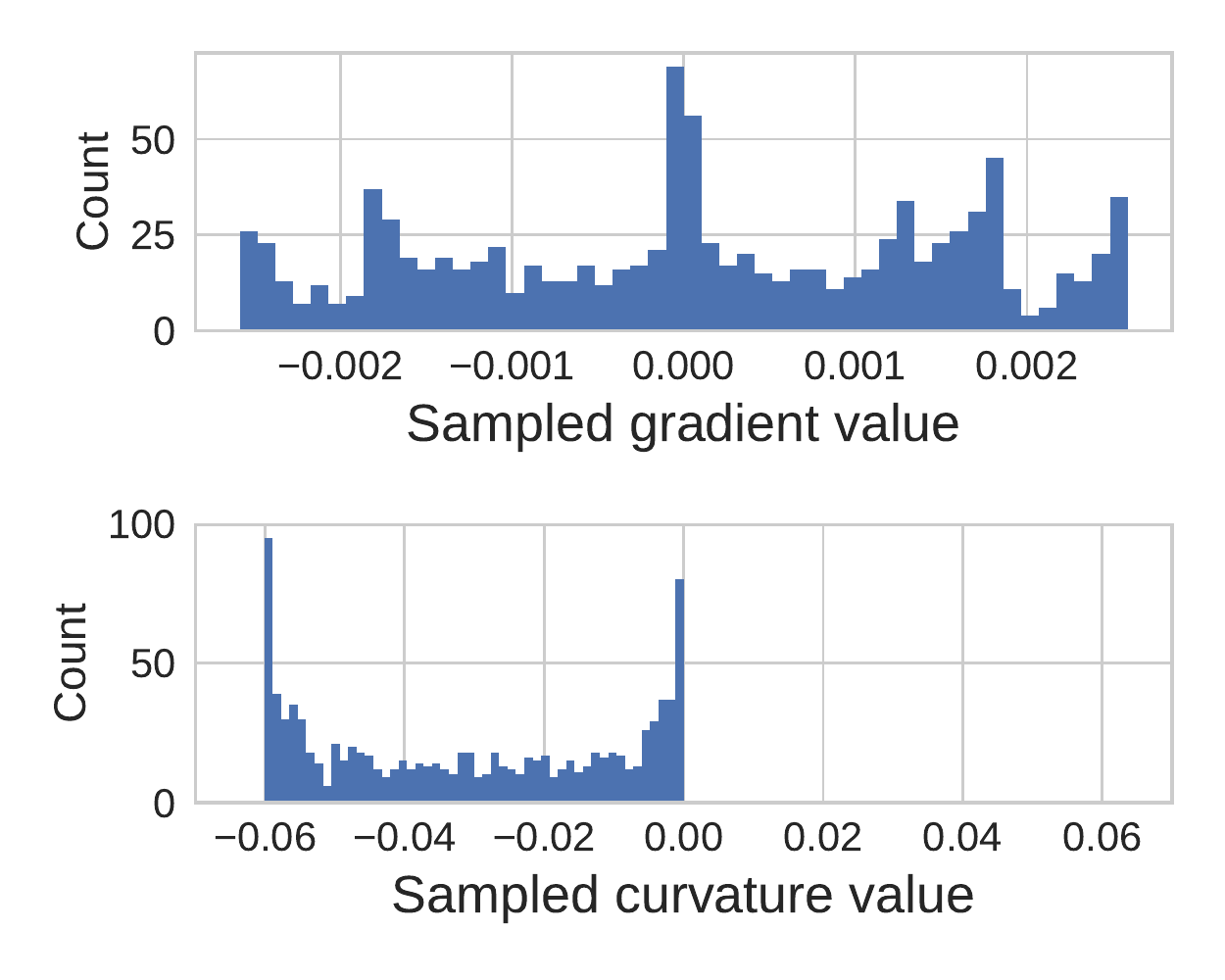}
        \caption{Apparent local optimum}
    \end{subfigure}
    \begin{subfigure}[b]{0.235\textwidth}
        \centering
        \includegraphics[trim={0.55cm 0.55cm 0.55cm 0.55cm},clip,width=\textwidth]{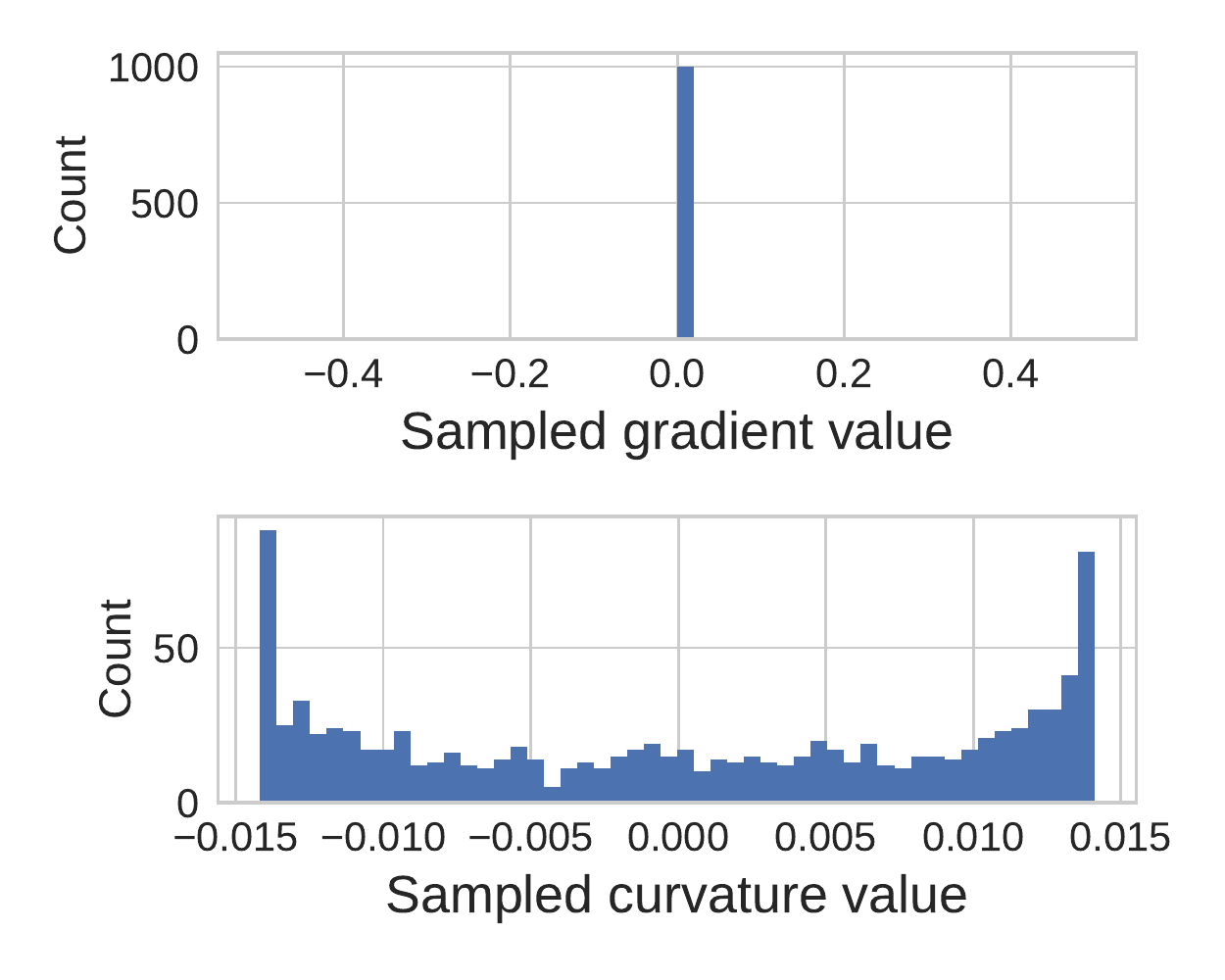}
        \caption{Saddle point}
    \end{subfigure}
    \vskip -0.1in
    \caption{\small \textbf{Demonstration of projecting random perturbations.} 
    (a) At the apparent local optimum, there is a tiny gradient and all curvatures are zero or negative. (b) At the saddle point, there is no gradient and curvatures are both positive and negative.}
    \label{fig:projections_example}
    \end{center}
    \vskip -0.2in
\end{figure}

Our method captures a lot of information about the local geometry. To summarize it, we can go one step further and disentangle information about gradient and curvature. If we assume $\Oh$ is locally quadratic, i.e., $\Oh(\theta)\approx a^T\theta + \frac{1}{2}\theta^T H \theta$, where $a$ is the linear component and $H$ is a symmetric matrix (i.e., Hessian), then: 
\begin{align}
\label{eqn:projected_final_answer}
\Delta^{\Oh+}_d - \Delta^{\Oh-}_d &= 2\alpha\nabla \Oh(\theta_0)^Td \;,\\
\Delta^{\Oh+}_d + \Delta^{\Oh-}_d &= \alpha^2d^T Hd \; .
\end{align}
The derivation is done in Appendix~\ref{asec:obj_vis}. Therefore, projections of our scatter plots capture information about the components of the gradient and Hessian in the random direction $d$. By repeatedly sampling many directions we eventually recover how the gradient and curvature vary in many directions around $\theta_0$. We can use a histogram to describe the density of these curvatures (Figure~\ref{fig:projections_example}). In particular, the maximum and minimum curvature values obtained from this technique are close to the maximum and minimum eigenvalues of $H$. This curvature spectrum is related to eigenvalue spectra which have been used before to analyze neural networks \citep{le1991eigenvalues,dauphin2014identifying}. 

While this work analyzes models with hundreds of parameters, a technique based on random perturbations is likely to miss some direction in higher dimensions. This is particularly problematic if these dimensions are the ones an optimizer would follow. We consider this limitation in the context of stochastic gradient methods, where $\nabla_\theta\Oh$ is a noisy estimate of the true gradient. At points where the objective does not change much, the true gradient is small compared to the noise introduced by the stochasticity. Therefore, the direction followed by the stochastic gradient method is dominated by noise: to escape quickly, there must be many directions to follow that increase the objective.

We give an example of the behaviour of our technique in various toy settings in Section~\ref{sec:appendix_vis} and the methods are summarized in Figure~\ref{fig:method}.

Equipped with these tools, we now describe specific details about the RL optimization problem in the next section.

\subsection{The Policy Optimization Problem}
\label{approach_policy_opt}
In policy optimization, we aim to learn parameters, $\theta$, of a policy, $\pi_\theta(a|s)$, such that when acting in an environment the sampled actions, $a$, maximize the discounted cumulative rewards, i.e., $\Oh_{ER}(\theta)=\mathbb{E}_{\pi_\theta}[\sum_{t=1}^\infty \gamma^t r_t]$, where $\gamma$ is a discount factor and $r_t$ is the reward at time $t$. The gradient is given by the policy gradient theorem \citep{sutton2000policy} as: $\nabla_\theta \Oh_{ER}(\theta) = \int_{s}d^{\pi_\theta}(s) \int_a\nabla_\theta\pi_\theta(a|s)Q^{\pi_\theta}(s,a) \text{d}a\text{d}s$
where $d^\pi$ is the stationary distribution of states and $Q^\pi(a_t,s_t)$ is the expected discounted sum of rewards starting state $s$, taking action $a$ and then sampling actions according to the policy, $a\sim\pi(\cdot|s)$. 

One approach to prevent premature convergence to a deterministic policy is to use entropy regularization \citep{schulman2017equivalence}. This is often done by augmenting the rewards with an entropy term, $\mathbb{H}(\pi(\cdot|s_t))=\mathbb{E}_{a\sim\pi(\cdot|s_t)}[-\log \pi(a|s_t)]$, weighted by $\tau$, i.e. $r_t^\tau = r_t + \tau\mathbb{H}(\pi(\cdot|s_t))$, and results in a slightly different gradient: 
\begin{align}
\label{eqn:policy_gradient_thm_entropy}
&\nabla_\theta \Oh_{ENT}(\theta) = \int_{s}d^{\pi_\theta}(s)\int_a  \pi(a|s)\Big[
\nonumber \\
&Q^{\tau,\pi_\theta}(s,a)\nabla_\theta\log\pi(a|s) + \tau\nabla_\theta\mathbb{H}(\pi(\cdot|s)) \Big]\text{d}a\text{d}s
\end{align}
where $Q^{\tau,\pi_\theta}(s,a)$ is the expected discounted sum of entropy-augmented rewards. $Q^{\tau,\pi_\theta}$ can be calculated exactly if the dynamics of the environment are known (\citet{sutton2000policy}, Appendix~\ref{asec:derivation_exact_pg}) or estimated by executing $\pi_\theta$ in the environment (\citet{williams1992simple}, Appendix~\ref{asec:reinforce}). It is noteworthy that both $\Oh_{ER}$ and $\nabla \Oh_{ENT}$ depend on $\pi_\theta$: Therefore, any change in the policy will change the experience distribution, $d^{\pi_\theta}$ and be reflected in both the objective and gradient.

\section{Results}
\label{sec:results}
Now that we have the tools to investigate objective landscapes from Section~\ref{approach_obj_funcs}, we return to questions related to entropy and policy optimization. Firstly, in Section~\ref{sec:high_variance_issue}, we use environments with no gradient estimation error to provide evidence for policy optimization being difficult due to the geometry of the objective function. Our main contribution is in Section~\ref{sec:continuous_control_results}, where we observe the smoothing effect of stochastic policies on the optimization landscape in high dimensional environments. Put together, these results should highlight the difficulty and environment-dependency of designing optimization techniques that are orthogonal to variance reduction of the gradient estimate.

\subsection{Entropy Helps Even with the Exact Gradient}
\label{sec:high_variance_issue} 
The high variance in gradient estimates due to using samples from a stochastic policy in a stochastic environment is often the reason given for poor optimization. To emphasize that policy optimization is difficult even if we solved the high variance issue, we conduct experiments in a setting where the optimization procedure has access to the exact gradient. We then link the poor optimization performance to visualizations of the objective function. Finally, we show how having an entropy augmented reward and, in general, a more stochastic policy changes this objective resulting in overall improvement in the solutions found.

\subsubsection{Experimental Setup: Environments with No Variance in the Gradient Estimate}

\begin{figure}[t]
    \begin{minipage}[c]{0.6\columnwidth}
    \caption{\label{sfig:gridworld}Gridworld used in the experiments in Section~\ref{sec:high_variance_issue}. There are two locally optimal policies: always going right and always going bottom.}
  \end{minipage}\hfill
  \begin{minipage}[c]{0.4\columnwidth}
    \includegraphics[width=\textwidth]{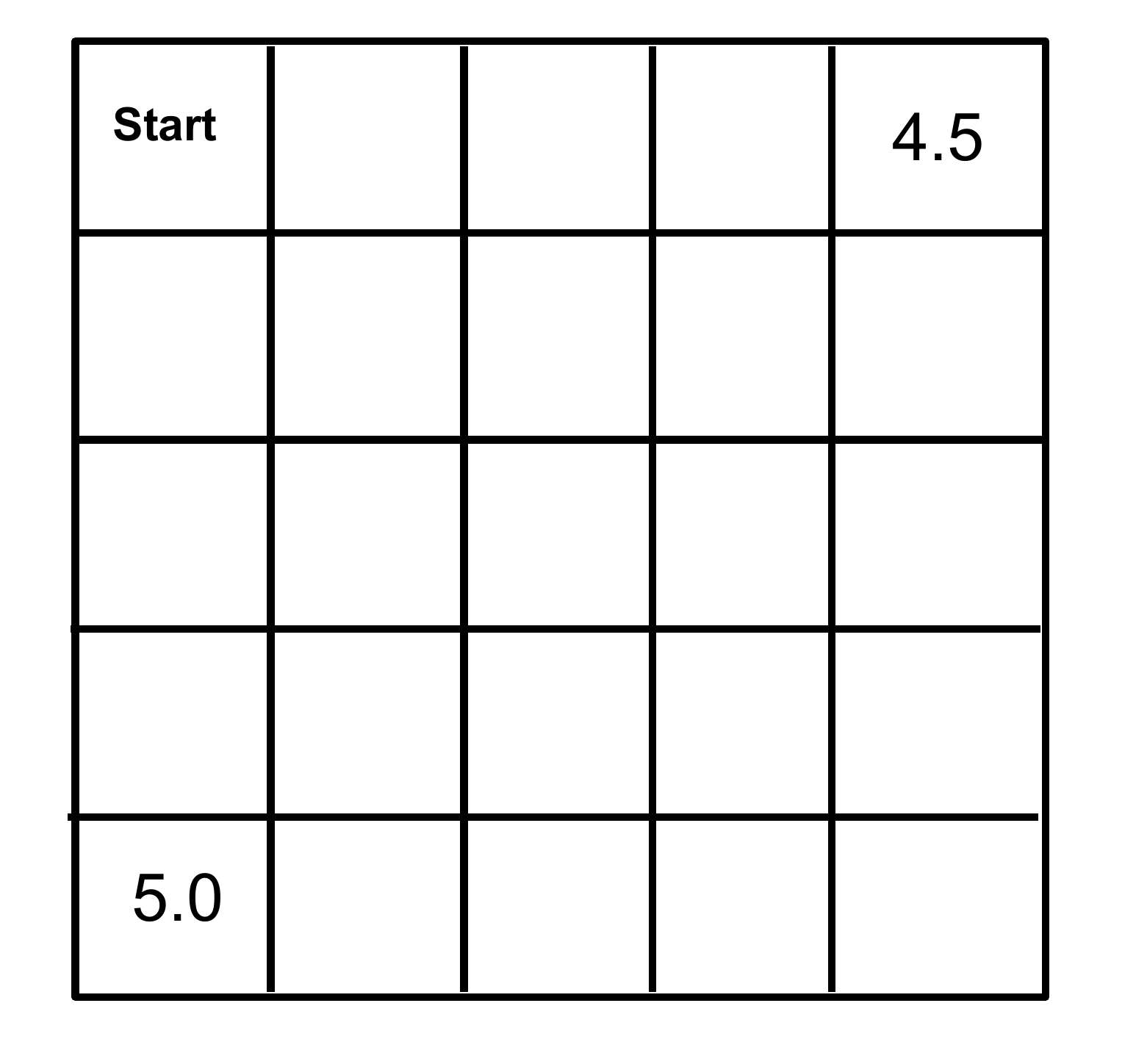}
  \end{minipage}
\end{figure}

To investigate if mitigating the high variance problem is the key to better optimization, we set our experiment in an environment where the gradient can be calculated exactly. In particular, we replace the integrals with summations and use environment dynamics to calculate Equation~\ref{eqn:policy_gradient_thm_entropy} resulting in no sampling error. 
We chose a $5\times5$ Gridworld with one suboptimal and one optimal reward at the corners (Figure~\ref{sfig:gridworld}). 
Our agent starts in the top left corner and has four actions parameterized by a categorical distribution $\pi(a|s_t)\propto \exp(\theta^Ts_t)$ and states are given by their one-hot representation. As such there are two locally optimal policies: go down, $\pi_\text{opt}$ and go right, $\pi_\text{sub}$. 
We refer to the case where the entropy weight $\tau=0$ as the \emph{true objective}.

\subsubsection{Are Poor Gradient Estimates the Main Issue with Policy Optimization?}
\begin{figure}[t]
    \vskip 0.2in
    \begin{center}
    \begin{subfigure}[b]{0.24\textwidth}
        \centering
        \includegraphics[trim={0.75cm 0.75cm 0.75cm 0.75cm},clip,width=\textwidth]{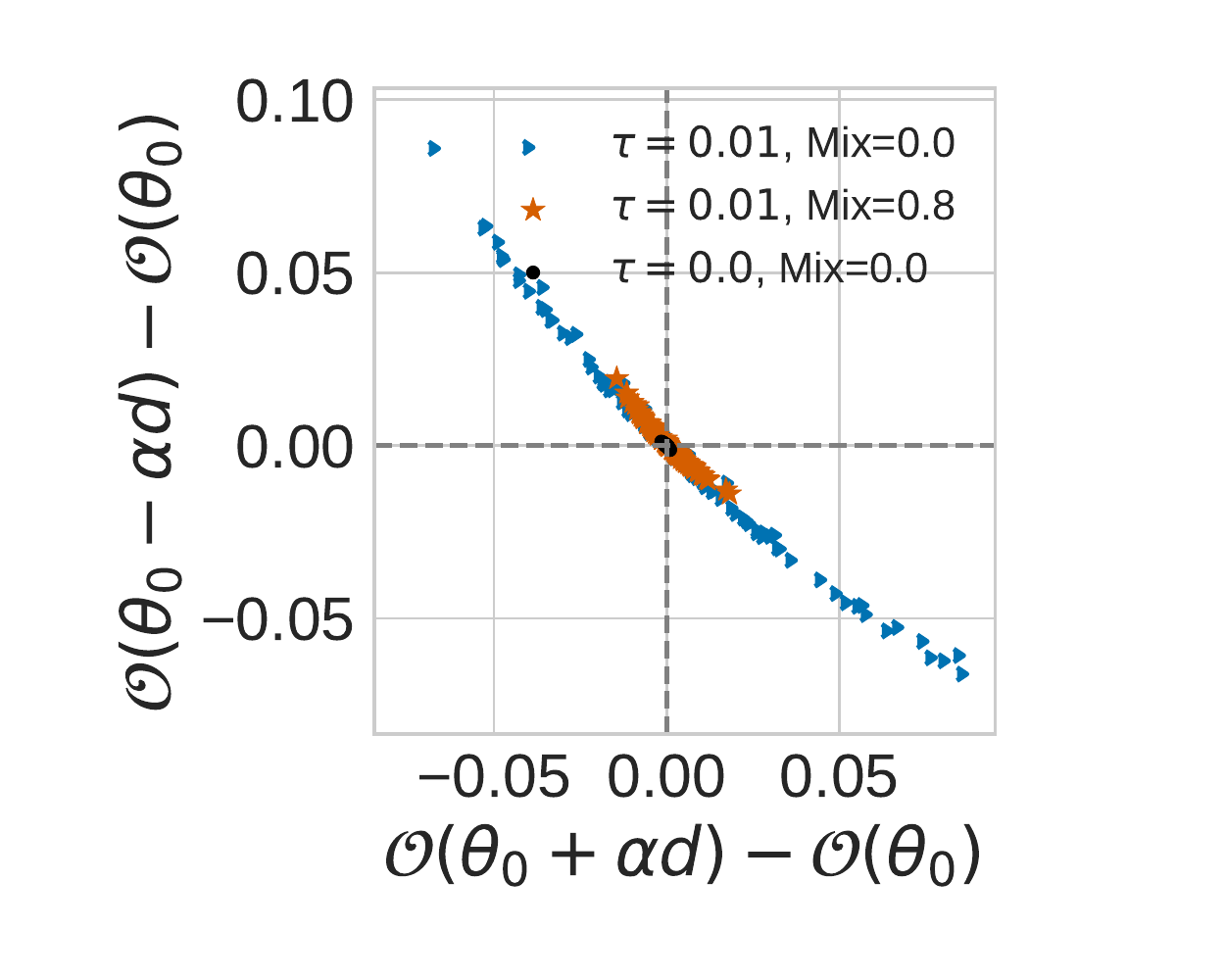}
        \caption{}
    \end{subfigure}
    \begin{subfigure}[b]{0.23\textwidth}
        \centering
        \includegraphics[trim={0.75cm 0.75cm 0.75cm 0.75cm},clip,width=\textwidth]{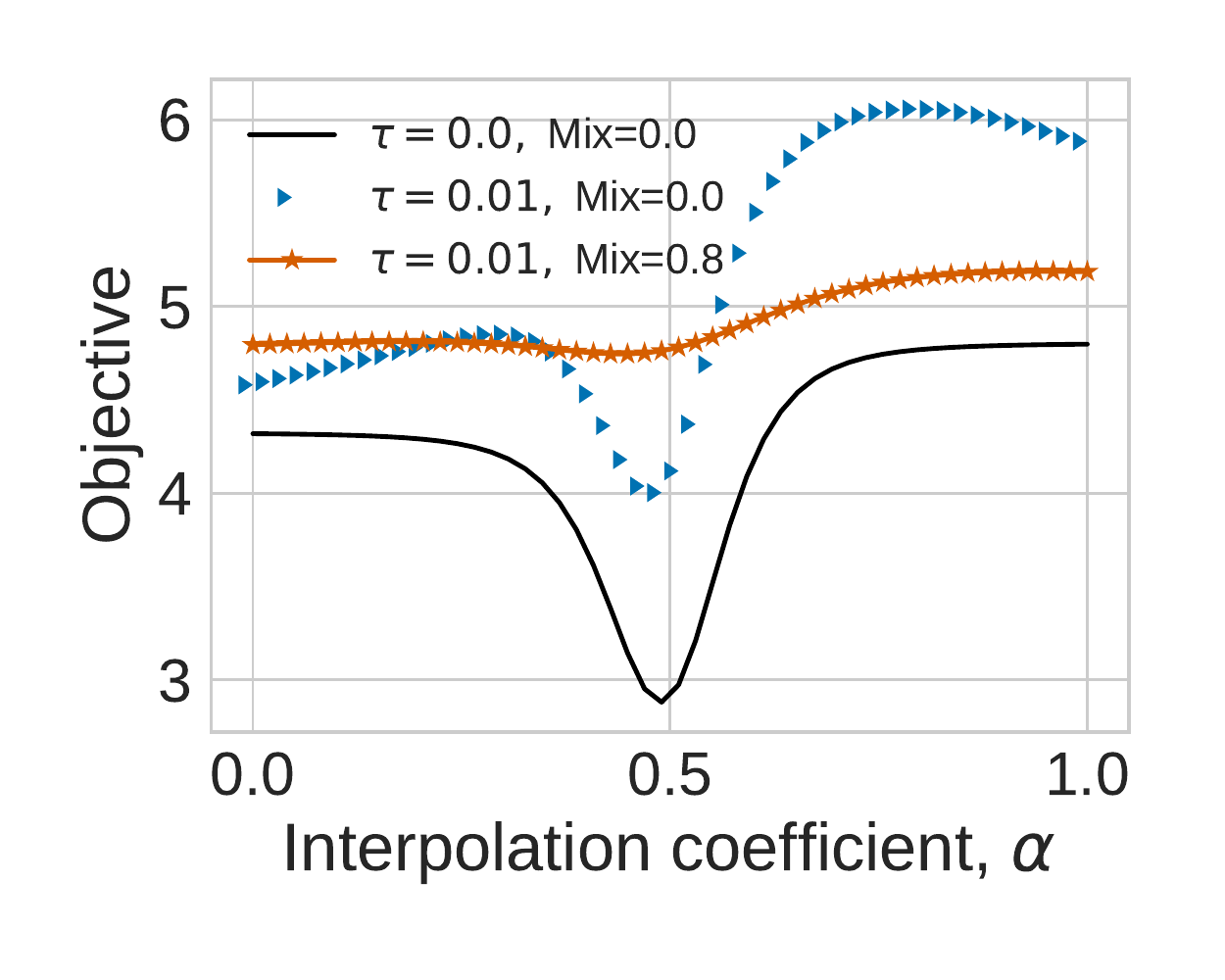}
        \caption{}
    \end{subfigure}
    \vskip -0.1in
    \caption{
    {\small \textbf{Objective function geometry around solutions in the Gridworld.} (a) Scatter plot for the change in objective for different random directions. Without entropy, no sampled directions show improvement in the objective (black circles). However, with either entropy or a more stochastic policy (blue triangle and orange star) many directions give positive improvement. (b) Linear interpolation between two solution policies for sub-optimal and optimal rewards are separated by a valley of poor solutions in the true objective (black). Using a stochastic policy and entropy regularized objective connects these local optima (orange stars) in this 1D slice. See Section~\ref{sec:high_variance_issue} for a detailed explanation.}}
    \label{fig:difficulty_exact}
    \end{center}
    \vskip -0.2in
\end{figure}

After running exact gradient ascent in the Gridworld starting from different random initializations of $\theta_0$,  we find that about $25\%$ of these initializations led to a sub-optimal final policy: there is some inherent difficulty in the geometry of the optimization landscape independent of sampling noise. To get a better understanding of this landscape, we analyze two solutions that parameterize policies that are nearly deterministic for their respective rewards $\theta_\text{sub}$ and $\theta_\text{opt}$. The objective function around $\theta_\text{sub}$ has a negligible gradient and small strictly negative curvature values indicating that the solution is in a very flat region (Figure~\ref{fig:difficulty_exact}a, black circles). 
On a more global scale, $\theta_\text{sub}$ and $\theta_\text{opt}$ are located in flat regions separated by a sharp valley of poor solutions (Figure~\ref{fig:difficulty_exact}b, black circles).

These results suggest that at least some of the difficulty in policy optimization comes from the flatness and valleys in the objective function independent of poor gradient estimates. In the next sections, we investigate effect of entropy regularization on the objective function.

\subsubsection{Why Does Using Entropy Regularization Find Better Solutions?}
\label{sec:entropy_better_solutions_gridworld} 
Our problem setting is such that an RL practitioner would intuitively think of using entropy regularization to encourage the policy to ``keep exploring'' even after finding $R_\text{sub}$.  Indeed, including entropy ($\tau>0$) and decaying it  during optimization, reduces the proportion of sub-optimal solutions found by the optimization procedure to 0 (Figure~\ref{sfig:entropy_solutions})\footnote{We classify a policy as \emph{optimal} if it achieves a return greater than that of the deterministic policy reaching $R_\text{sub}$.}. We explore reasons for the improved performance in this section.

We see in Figure~\ref{fig:difficulty_exact}a (orange stars) that augmenting the objective with entropy results in many directions of positive improvement at $\theta_\text{sub}$: it is no longer a flat solution.

We also show that interpolating a stochastic\footnote{This is done by setting the policy being evaluated to $(1-\text{Mix})\pi_{(1-\alpha)\theta_0 + \alpha\theta_1}(a|s) + \frac{\text{Mix}}{|A|}$ so that every action has a minimum probability of $\frac{\text{Mix}}{|A|}$.} version of the policy can connect the two local optima in this slice: A smooth, and in some cases monotonically increasing, path now appears in this augmented objective to a region with high value in the true objective (Figure~\ref{fig:difficulty_exact}b orange stars,~\ref{sfig:difficulty_exact}). This means that if we knew a good direction a priori, a line search would find a better solution.

This section provided a different and more accurate interpretation for entropy regularization by connecting it to changes in objective function geometry. Given that entropy regularization encourages our policies to be more stochastic, we now ask \emph{What is it about stochastic policies that helps learning?} In the next section, we explore two possible reasons for this in a more realistic high dimensional continuous control problem.

\subsection{More Stochastic Policies Induce Smoother Objectives in Some Environments}
\label{sec:continuous_control_results}
In Section~\ref{sec:entropy_better_solutions_gridworld} we saw that entropy and, more generally, stochasticity can induce a ``nicer'' objective to optimize. Our second experimental setup allows us to answer questions about the optimization implications of stochastic policies. We show qualitatively that high entropy policies can speed up learning and improve the final solutions found. 
We empirically investigate some reasons for these improvements related to smoothing. We also show that these effects are environment-specific highlighting the challenges of policy optimization.

\subsubsection{Experimental Setup: Environments that Allow Explicit Study of Policy Entropy}
Continuous control tasks from the MuJoCo simulator \cite{todorov2012mujoco,brockman2016openai} facilitate studying the impact of entropy because we can parameterize policies using Gaussian distributions. In particular, since entropy regularization increases the entropy of a policy, we can study the impact of entropy on optimization by controlling the stochasticity of the Gaussian policy. Specifically, the entropy of a Gaussian distribution depends only on $\sigma$, and thus we control $\sigma$ explicitly to study varying levels of entropy. We use a large batch size to control for the variance reduction effects of a larger $\sigma$ \citep{zhao2011analysis}.

To keep the analysis as simple as possible, we parameterize the mean by $\theta^Ts_t$ which is known to result in good performance \citep{rajeswaran2017towards}. Since we do not have access to transition and reward dynamics, we cannot calculate the policy gradient exactly and use the \textsc{Reinforce} gradient estimator (\citet{williams1992simple}, Appendix~\ref{asec:reinforce}). To study performance during learning we consider the \emph{deterministic} performance\footnote{In this setting, the online performance does not matter, we only rely on the stochasticity for ``exploration''. We note that the deterministic performance will be better than the online stochastic performance of the policy.} by evaluating the learned policy with $\sigma=0$.

To study the landscape, we use the techniques described in Section~\ref{approach_obj_funcs} to analyze $\theta$ under different values of $\sigma$. Specifically, we obtain multiple values of $\theta$ by optimizing different values of $\sigma$. To understand how objective landscapes change, we re-evaluate interpolated slices and random perturbations under a different value of $\sigma$. We consider $\sigma=0$ to be the policy we are interested in and refer to the objective calculated as the \emph{true objective}.

\subsubsection{What is the Effect of Entropy on Learning Dynamics?}
\label{res:learning_speed}
We first show that optimizing a more stochastic policy can result in faster learning in more complicated environments and better final policies in some.

\begin{figure*}[ht]
    \begin{center}
    \vskip 0.2in
    \begin{subfigure}[b]{0.32\textwidth}
        \centering
        \includegraphics[width=\textwidth]{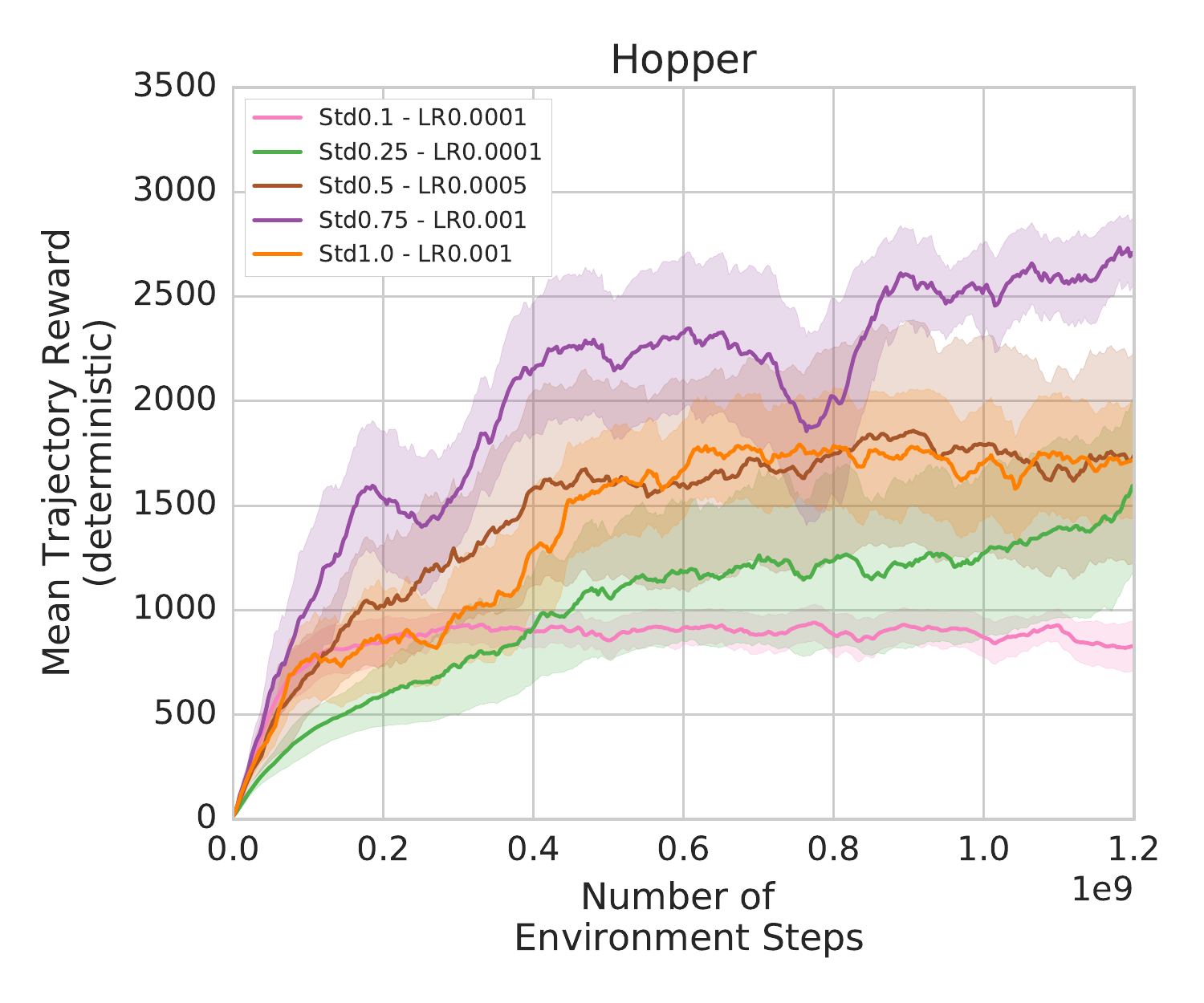}
        \caption{}
    \end{subfigure}
    \begin{subfigure}[b]{0.32\textwidth}
        \centering
        \includegraphics[width=\textwidth]{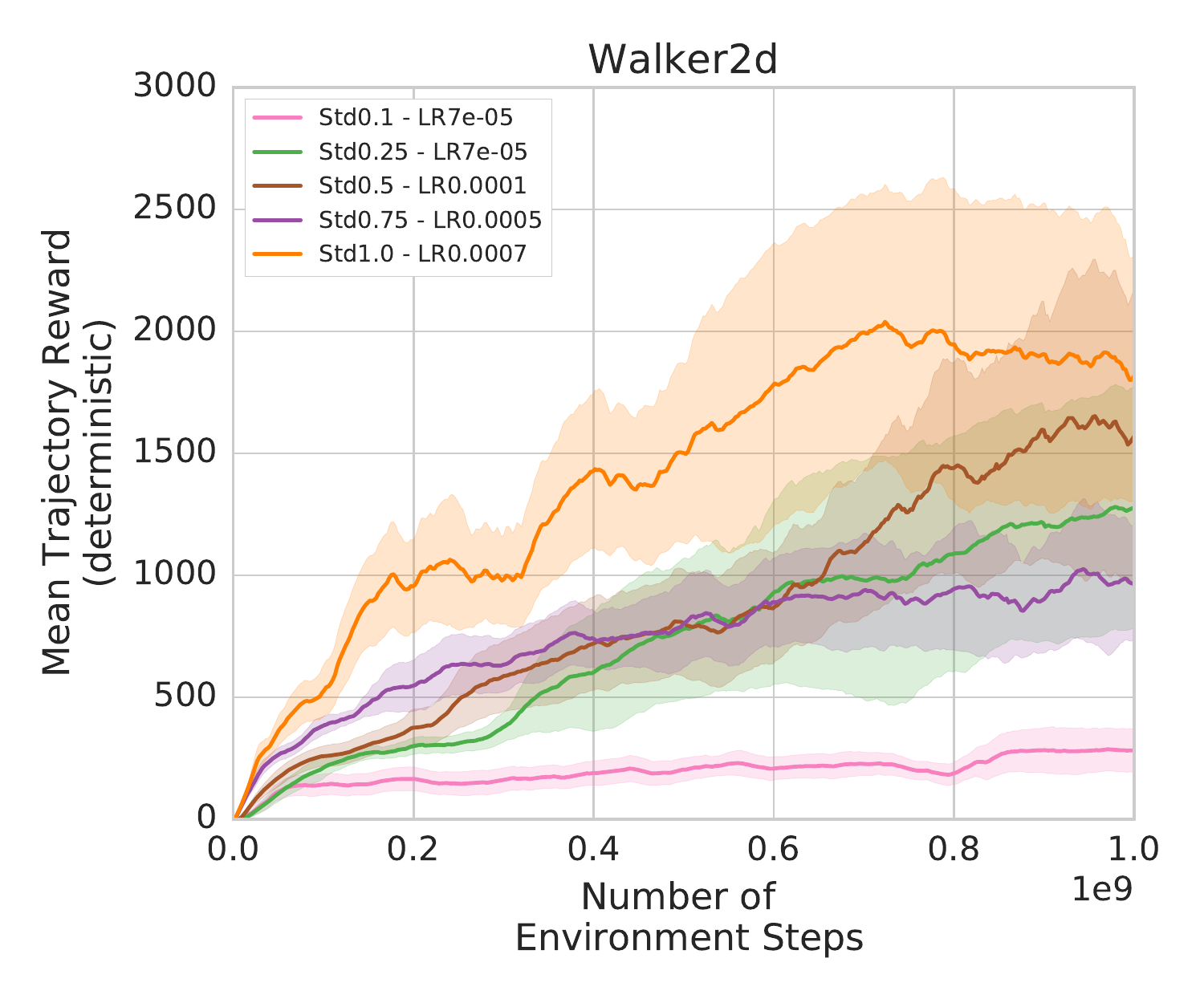}
        \caption{}
    \end{subfigure}
    \begin{subfigure}[b]{0.32\textwidth}
        \centering
        \includegraphics[width=\textwidth]{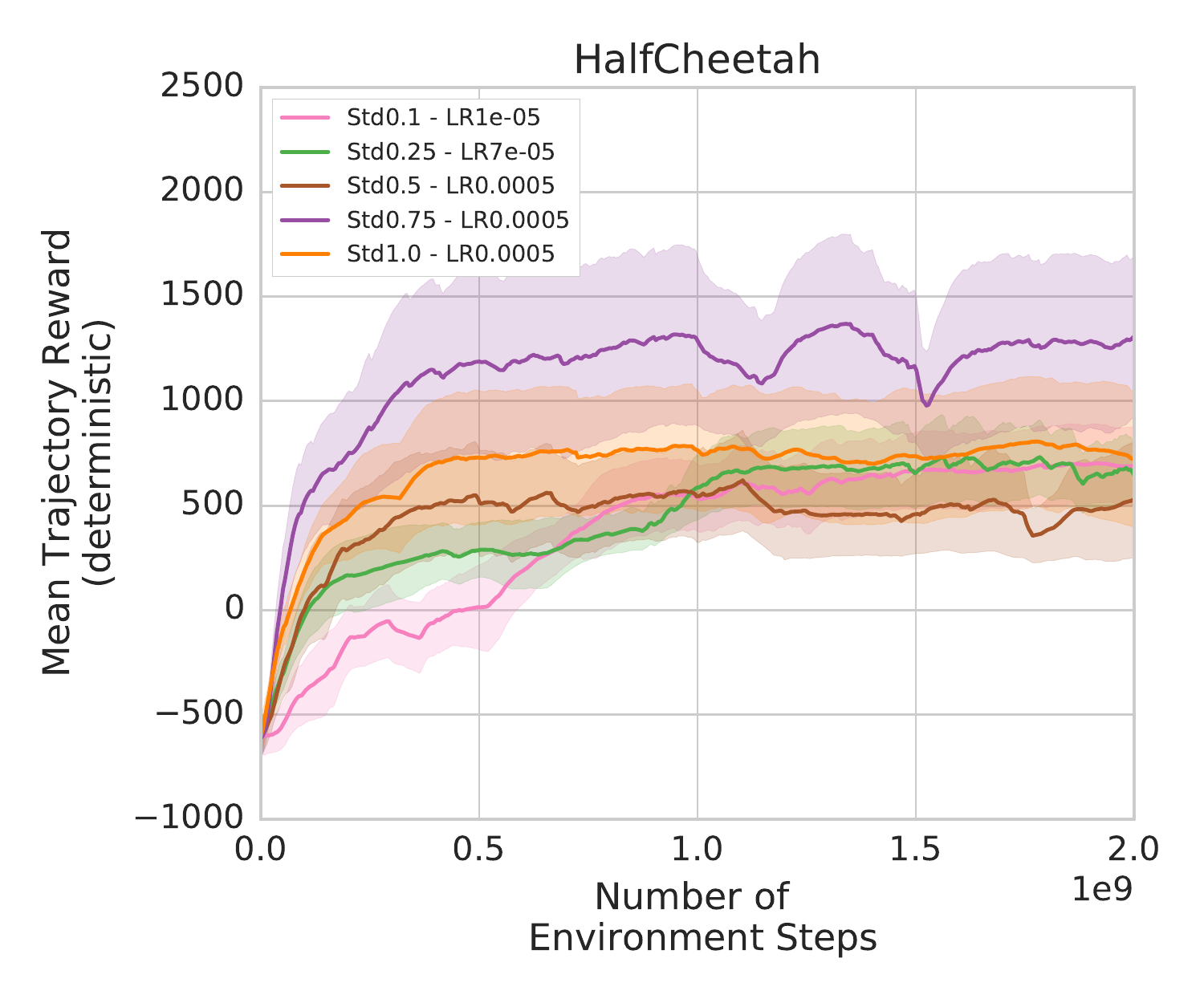}
        \caption{}
    \end{subfigure}
    \caption{
    {\small \textbf{Learning curves for policies with different entropy in continuous control tasks
    } In all environments using a high entropy policy results in faster learning. These effects are less apparent in HalfCheetah. In Hopper and Walker high entropy policies also find better final solutions. Learning rates are shown in the legends and entropy is controlled using the standard deviation. Solid curve represents the average of 5 random seeds. Shaded region represents half a standard deviation for readability. Individual learning curves are shown in Figure~\ref{sfig:hopper_individual_learning_curves} for Hopper, Figure~\ref{sfig:walker_individual_learning_curves} for Walker and Figure~\ref{sfig:halfcheetah_individual_learning_curves} for HalfCheetah. See Section~\ref{sec:learning_dynamics}~and~\ref{sec:smoothing_local_optima} for discussion.}}
    \label{fig:learning_curves}
    \end{center}
    \vskip -0.2in
\end{figure*}

In Hopper and Walker high entropy policies ($\sigma>0.1$) quickly achieve higher rewards than low entropy policies ($\sigma=0.1$) (Figure~\ref{fig:learning_curves}ab). In HalfCheetah, even though high entropy policies learn quicker (Figure~\ref{fig:learning_curves}c), the differences are less apparent and are more strongly influenced by the initialization seed (Figure~\ref{sfig:halfcheetah_individual_learning_curves}). 
In both Hopper and Walker2d, the mean reward of final policies found by optimizing high entropy policies is $2$ to 
$8$ times larger than a policy with $\sigma=0.1$ whereas, in HalfCheetah, all policies converge to a similar final reward commonly observed in the literature \citep{schulman2017proximal}.

Though statistical power to make fine-scale conclusions is limited, the qualitative trend holds: More stochastic policies perform better in terms of speed of learning and, in some environments, final policy learned. In the next two sections we investigate some reasons for these performance improvements as well as the discrepancy with HalfCheetah.

\subsubsection{Why do high entropy policies learn quickly?}
\label{sec:learning_dynamics}
We first focus on the speed of learning: A hint for the answer comes from our hyperparameter search over constant learning rates. In Hopper and Walker, the best learning rate increases consistently with entropy: The learning rate for $\sigma=1$ is $10$ times larger than for $\sigma=0.1$. At every time step, the optimal learning rate is tied to the local curvature of the loss. If the curvature changes, the best constant learning rate is the smallest one which works across all curvatures. The change in optimal learning rate cannot be solely explained by the implicit rescaling of the parameters induced by the change in $\sigma$ since the loss also increases faster with higher entropy, which would not happen with a mere scalar reparametrization. Thus, that difference in optimal learning rate suggests that adding entropy instead damps the variations of curvature along the trajectory, facilitating optimization with a constant learning rate $\eta$.

\begin{figure}[ht]
    \captionsetup[subfigure]{aboveskip=-3pt,belowskip=-3pt}
    \begin{center}
    \begin{subfigure}[b]{0.4\textwidth}
    \centering
    \includegraphics[width=\textwidth]{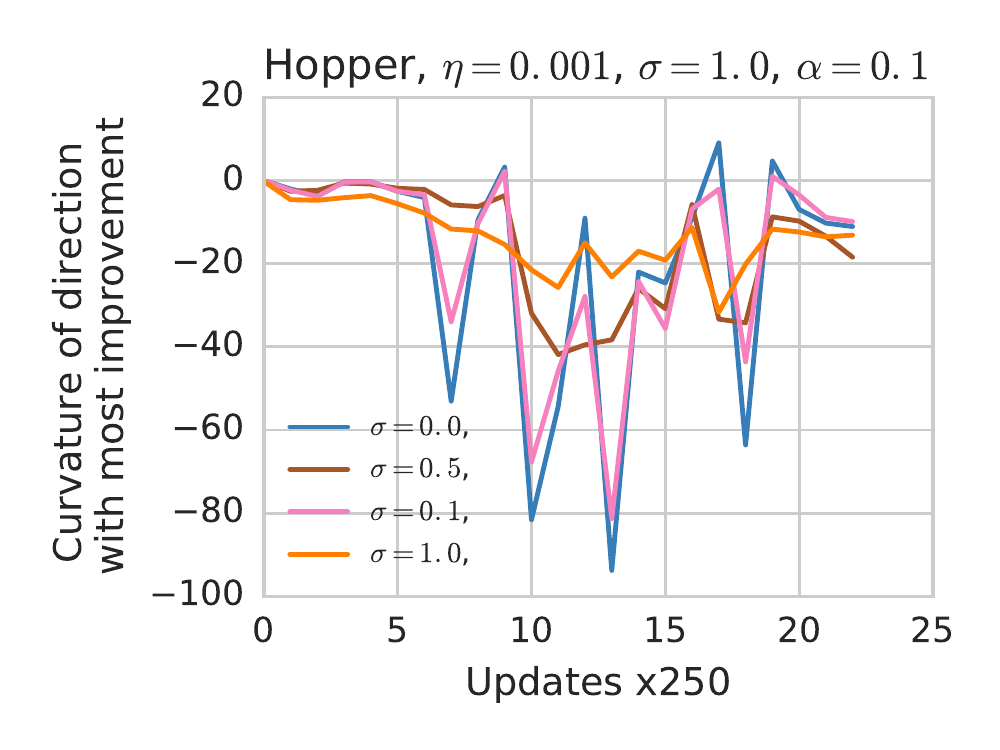}
    \caption{}
    \end{subfigure}
    \begin{subfigure}[b]{0.37\textwidth}
    \centering
    \includegraphics[width=\textwidth]{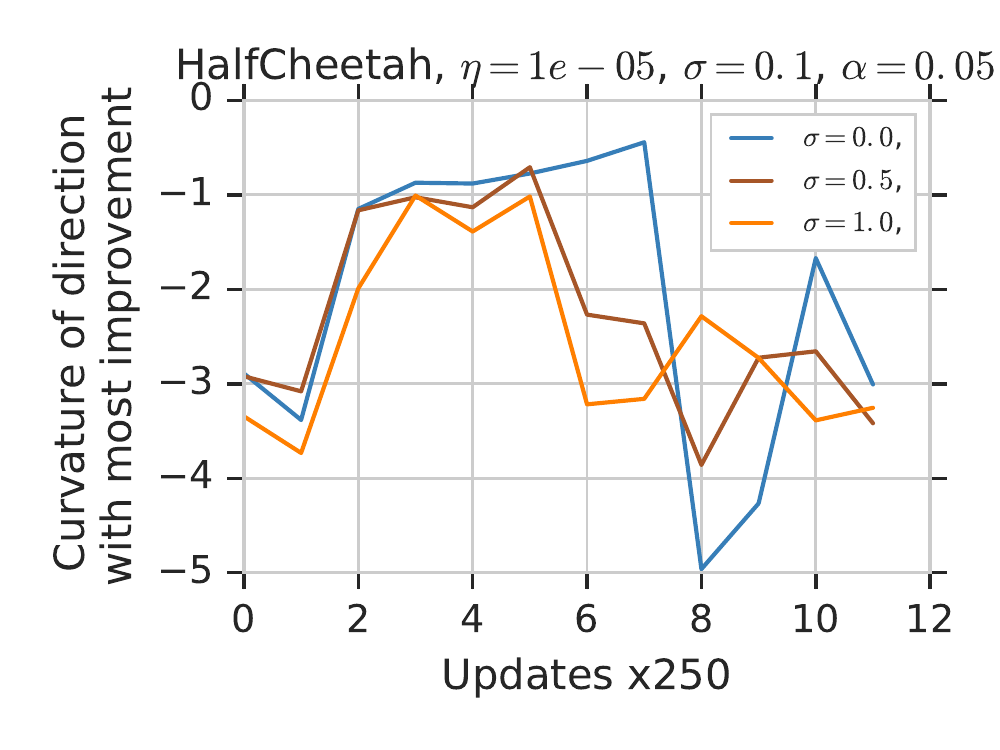}
    \caption{}
    \end{subfigure}
    \caption{\small \textbf{Curvature during training}
    (a) The curvature for the direction of most improvement in Hopper fluctuates rapidly for optimization objectives with low entropy ($\sigma\in\{0.0, 0.1\}$) compared to those of high entropy ($\sigma\in\{0.5, 1.0\})$. (b) Entropy does not seem to have any effect on HalfCheetah.
    \label{fig:curvature_fluctuation}}
    \end{center}
    \vskip -0.2in
\end{figure}

To investigate, we calculate the curvature of the objective during the first few thousand iterations of the optimization procedure. In particular, we record the curvature in a direction of improvement\footnote{We selected the direction of improvement closest to the 90th percentile which would be robust to outliers.}. As expected, curvature values fluctuate with a large amplitude for low values of $\sigma$ (Figure~\ref{fig:curvature_fluctuation}a,~\ref{sfig:curvature_fluctuation_hopper},~\ref{sfig:curvature_fluctuation_walker}). In this setting, selecting a large and constant $\eta$ might be more difficult compared to an objective induced by a policy with a larger $\sigma$. In contrast, the magnitude of fluctuations are only marginally affected by increasing $\sigma$ in HalfCheetah (Figure~\ref{fig:curvature_fluctuation}b~and~\ref{sfig:curvature_fluctuation_halfcheetah}) which might explain why using a more stochastic policy in this environment does not facilitate the use of larger learning rates.

In this section, we showed that fluctuations in the curvature of objectives decrease for more stochastic policies in some environments. The implications for these are two-fold: (1) It provides evidence for why high entropy policies facilitate the use of a larger learning rate; and (2) The impact of entropy can be highly environment specific. In the next section, we shift our focus to investigate the reasons for improved quality of \emph{final} policies found when optimizing high entropy policies.

\subsubsection{Can High Entropy Policies Reduce the Number of Local Optima in the Objective?}
\label{sec:smoothing_local_optima}
In this section, we improve our understanding of which values of $\theta$ are reachable at the end of optimization. 
We are trying to understand \emph{Why do high entropy policies learn better final solutions?} Specifically, we attempt to classify the local geometry of parameters and investigate the effect of making the policy more stochastic. We will then argue that high entropy policies induce a more connected landscape. 

\begin{figure}[t]
    \vskip 0.2in
    \begin{center}
    \begin{subfigure}[b]{0.235\textwidth}
        \centering
        \includegraphics[trim={0.60cm 0.65cm 0.65cm 0.60cm},clip,width=\textwidth]{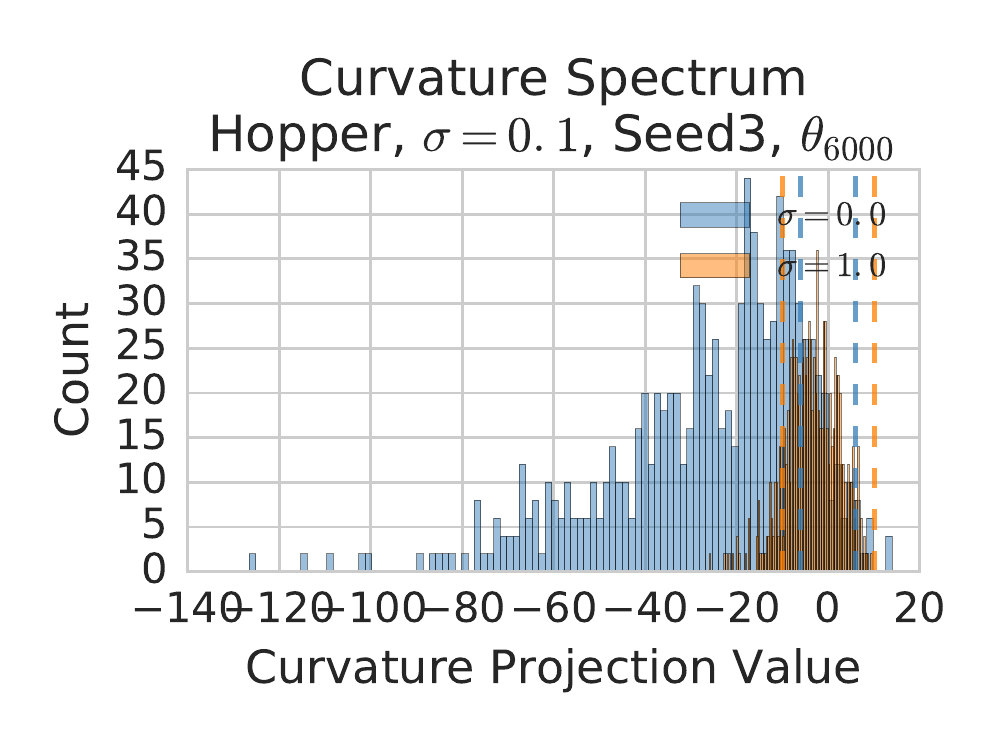}
        \caption{
        }
    \end{subfigure}
    \begin{subfigure}[b]{0.235\textwidth}
        \centering
        \includegraphics[trim={0.55cm 0.65cm 0.70cm 0.50cm},clip,width=\textwidth]{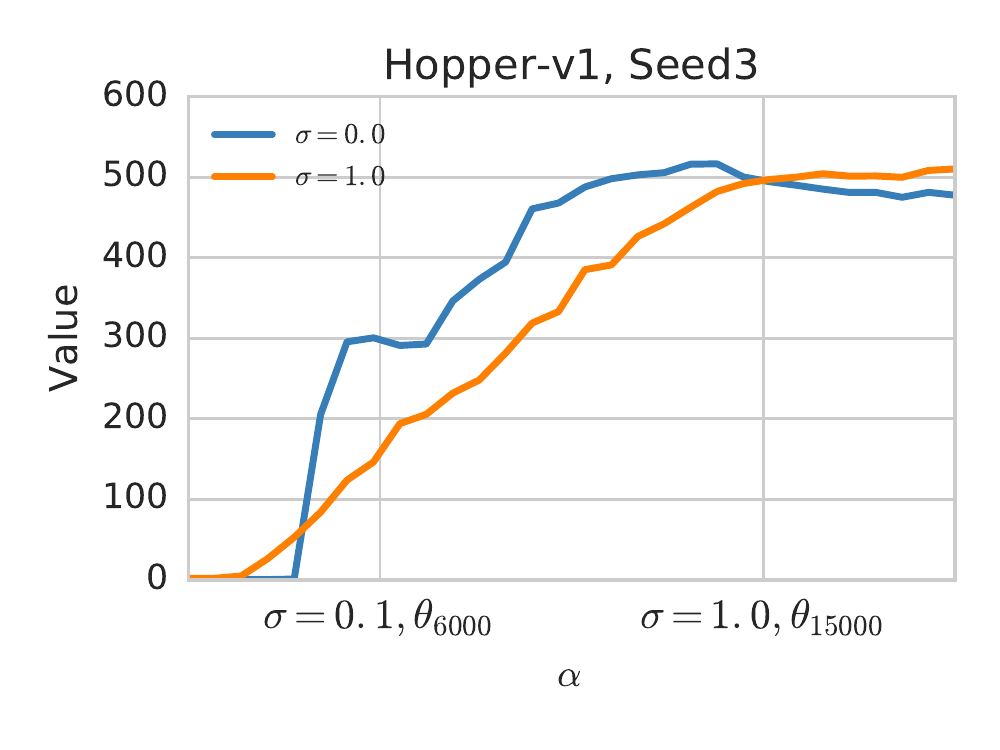}
        \caption{}
    \end{subfigure}
    \caption{{\small \textbf{Analyzing solutions in objectives given by different amounts of entropy} (a) For $\sigma=0.0$, $85\%$ of curvature values are negative. When $\sigma$ is increased to 1, nearly all curvature values are within sampling noise (indicated by dashed horizontal lines). (b) A linear interpolation shows that a monotonically increasing path to a better solution exists from the poor parameter vector. See Figures~\ref{sfig:hopper_optima_spectra}~and~\ref{sfig:walker_optima_spectra} for a different seed and in Walker respectively. See Figure~\ref{sfig:optima_negative} for negative examples.}}
    \label{fig:hopper_optima_spectra}
    \end{center}
    \vskip -0.2in
\end{figure}

Final solutions in Hopper for $\sigma=0.1$ have roughly $3$ times more directions with negative curvature than $\sigma=1.0$. This suggests that final solutions found when optimizing a high entropy policy lie in regions that are flatter and some directions \emph{might} lead to improvement. 
To understand if a more stochastic policy can facilitate an improvement from a poor solution, we visualize the local objective for increasing values of $\sigma$. Figure~\ref{fig:hopper_optima_spectra} shows this analysis for one such solution (Figure~\ref{sfig:hopper_individual_learning_curves}d) where the objective oscillates: The stochastic gradient direction is dominated by noise. For deterministic policies $84\%$ of directions have a detectable\footnote{Taking into account noise in the sampling process.} negative curvature (Figure~\ref{fig:hopper_optima_spectra}a) with the rest having near-zero curvature: The solution is likely near a local optimum. When the policy is made more stochastic, the number of directions with negative curvature reduces dramatically suggesting that the solution might be in a linear region. However, just because there are fewer directions with negative curvature, it does not imply that any of them reach good final policies.

To verify that there exists \emph{at least} one path of improvement to a good solution, we linearly interpolate between this solution and parameters for a good final policy obtained by optimizing $\sigma=1.0$ starting from the same random initialization (Figure~\ref{fig:hopper_optima_spectra}b). Surprisingly, even though this is just one very specific direction, there exists a monotonically increasing path to a better solution in the high entropy objective: If the optimizer knew the direction in advance, a simple line search would have improved upon a bad solution when using a high entropy policy. This finding extends to other pairs of parameters (Figure~\ref{sfig:hopper_optima_spectra} for Hopper and Figure~\ref{sfig:walker_optima_spectra}~and~\ref{sfig:walker_optima_spectra2} for Walker2d) but not all (Figure~\ref{sfig:optima_negative}) indicating that \emph{some} slices of the objective function may become easier to optimize and find better solutions. 

Our observations do not extend to HalfCheetah, where we were unable to find such pairs (Figure~\ref{sfig:optima_negative}c) and specifically, objectives around final solutions did not change much for different values of $\sigma$ (Figure~\ref{sfig:halfcheetah_optima_spectra}). These observations suggest that the objective landscape in HalfCheetah is not significantly affected by changing the policy entropy and explains the marginal influence of entropy in finding better solutions in this environment.

As seen before the impact of entropy on the objective function seems to be environment specific. However, in environments where the objective functions are affected by having a more stochastic policy, we have evidence that they can reduce at least a few local optima by connecting different regions of parameter space. 

\section{Related Work}
\label{sec:related_work}
There have been many visualization techniques for objective functions proposed in the last few years \citep{goodfellow2014qualitatively,li2017visualizing,draxler2018essentially}. In contrast to our work, many of these project the high dimensional objective into one or two useful dimensions. 
Our technique is closely related to those of \citet{li2017visualizing}, who used random directions for 2D interpolations, and of \citet{li2018measuring,fort2018goldilocks} who studied optimization in random hyperplanes and hyperspheres. Our work differs in that we interpolate in many more directions to summarize how the objective function changes locally. \citet{keskar2016large} used a summary metric from random perturbations to measure sharpness of an objective similar to our measure of curvature in Figure~\ref{fig:curvature_fluctuation}.

Understanding the impact of entropy on the policy optimization problem was first studied by \citet{williams1991function}. A different kind of entropy regularization has been explored in the context of deep learning: \citet{chaudhari2016entropy} show that such a penalty induces objectives with higher $\beta$-smoothness and complement our smoothness results. Recent work by \citet{neu2017unified} has shown the equivalence between the type of entropy used and a dual optimization algorithm. 

The motivation of our work is closely related to \citet{rajeswaran2017towards,henderson2018did,ilyas2018deep} in appealing to the community to study the policy optimization problem more closely. In particular, \citet{ilyas2018deep} show that in deep RL, gradient estimates can be uncorrelated with the true gradient despite having good optimization performance. This observation complements our work in saying that high variance might not be the main issue for policy optimization. The authors use 2D interpolations to show that an ascent in surrogate objectives used in PPO did not necessarily correspond to an ascent in the true objective. Our results provide a potential explanation to this phenomenon: surrogate objectives can connect regions of parameter space where the true objective might decrease.

\section{Discussion and Future Directions}
\label{sec:discussion}
\textbf{The difficulty of policy optimization.} Our work aims to redirect some research focus from the high variance issue to the study of better optimization techniques. In particular, even if we were able to perfectly estimate the gradient, policy optimization would still be difficult due to the geometry of the objective function used in RL. 

Specifically, our experiments bring to light two issues unique to policy optimization. Firstly, given that we are optimizing probability distributions, many reparameterizations can result in the same distribution. This results in objective functions that are especially susceptible to having flat regions and difficult geometries (Figure~\ref{fig:difficulty_exact}b,~\ref{sfig:walker_optima_spectra}c). There are a few solutions to this issue: As pointed out by \citet{kakade2001natural}, methods based on the natural gradient are well equipped to deal with plateaus induced by probability distributions. Alternatively, given the empirical success of natural policy gradient inspired methods like TRPO and surrogate objective methods like PPO suggests that these techniques are well motivated in RL \citep{schulman2015trust,schulman2017proximal,rajeswaran2017towards}. Such improvements are orthogonal to the noisy gradient problem and suggest that making policy optimization easier is a fruitful line of research.

Secondly, the landscape we are optimizing is problem dependent and is particularly surprising in our work. Given that the mechanics of many MuJoCo tasks are very similar, our observations on Hopper and HalfCheetah are vastly different. If our analysis was restricted to just Hopper and Walker, our conclusions with respect to entropy would have been different. This presents a challenge for both studying and designing optimization techniques. 

The MuJoCo environments considered here are deterministic given the random seed: An interesting and important extension would be to investigate other sources of noise and in general answering \emph{What aspects of the environment induce difficult objectives?} Our proposed method will likely be useful in answering at least a few such questions.

\textbf{Sampling strategies.} Our learning curves are not surprising under the mainstream interpretation of entropy regularization: A small value of $\sigma$ will not induce a policy that adequately ``explores'' the whole range of available states and actions. However, our results on HalfCheetah tell a different story: All values of $\sigma$ converged to policy with similar final reward (Figure~\ref{fig:learning_curves}c). 

Our combined results from curvature analysis and linear interpolations (Figure~\ref{fig:curvature_fluctuation}~and~\ref{fig:hopper_optima_spectra}) have shown that the geometry of the objective function is linked to the entropy of the policy being optimized. Thus using a more stochastic policy, in this case, induced by entropy regularization, facilitates the use of a larger learning rate and might provide more directions of improvement. 

Our results should hold for any technique that works by increasing policy entropy or collects data in a more uniform way\footnote{For example, the analysis in Section~\ref{sec:continuous_control_results} encompasses both the ``naive'' and ``proper'' entropy bonuses from \citet{schulman2017equivalence}.}. Investigating how other \emph{directed} or \emph{structured} sampling schemes impact the landscape will be useful to inform the design of new techniques. We conjecture that the ultimate effect of these techniques in RL is to make the objective function smoother and thus easier to optimize.

\textbf{Smoothing.} Finally, our experimental results make one suggestion: Smoothing can help learning. Therefore, \emph{How can we leverage these observations to make new algorithms?} The smoothing effect of entropy regularization, if decayed over the optimization procedure, is akin to techniques that start optimizing, easier, highly smoothed objectives and then progressively making them more complex \citep{chapelle2010gradient,gulcehre2016mollifying}. Perhaps some work should be directed on alternate smoothing techniques: \citet{santurkar2018does} suggests that techniques like batch normalization also smooth the objective function and might be able to replicate some of the performance benefits. In the context of RL, Q-value smoothing has been explored by \citet{nachum2018smoothed,fujimoto2018addressing} that resulted in performance gains for an off-policy policy optimization algorithm. 

In summary, our work has provided a new tool for and highlighted the importance of studying the underlying optimization landscape in direct policy optimization. We have shown that these optimization landscapes are highly environment-dependent making it challenging to come up with general purpose optimization algorithms. We show that optimizing policies with more entropy results in a smoother objective function that can be optimized with a larger learning rate. Finally, we identify a myriad of future work that might be of interest to the community with significant impact.

\section*{Acknowledgements}
The authors would like to thank Robert Dadashi and Saurabh Kumar for their consistent and useful discussions throughout this project; Riashat Islam, Pierre Thodoroff, Nishanth Anand, Yoshua Bengio, Sarath Chandar and the anonymous reviewers for providing detailed feedback on a draft of this manuscript; Prakash Panangaden and Clara Lacroce for a discussion on interpretations of the proposed visualization technique; Pierre-Luc Bacon for teaching the first author an alternate way to prove the policy gradient theorem; and Valentin Thomas, Pierre-Antoine Manzagol, Subhodeep Moitra, Utku Evci, Marc G. Bellemare, Fabian Pedregosa, Pablo Samuel Castro, Kelvin Xu and the entire Google Brain Montreal team for thought-provoking feedback during meetings. 

\bibliography{references}

\begin{thebibliography}{40}
\providecommand{\natexlab}[1]{#1}
\providecommand{\url}[1]{\texttt{#1}}
\expandafter\ifx\csname urlstyle\endcsname\relax
  \providecommand{\doi}[1]{doi: #1}\else
  \providecommand{\doi}{doi: \begingroup \urlstyle{rm}\Url}\fi

\bibitem[Brockman et~al.(2016)Brockman, Cheung, Pettersson, Schneider,
  Schulman, Tang, and Zaremba]{brockman2016openai}
Brockman, G., Cheung, V., Pettersson, L., Schneider, J., Schulman, J., Tang,
  J., and Zaremba, W.
\newblock Openai gym.
\newblock \emph{arXiv preprint arXiv:1606.01540}, 2016.

\bibitem[Chapelle \& Wu(2010)Chapelle and Wu]{chapelle2010gradient}
Chapelle, O. and Wu, M.
\newblock Gradient descent optimization of smoothed information retrieval
  metrics.
\newblock \emph{Information retrieval}, 13\penalty0 (3):\penalty0 216--235,
  2010.

\bibitem[Chaudhari et~al.(2017)Chaudhari, Choromanska, Soatto, LeCun, Baldassi,
  Borgs, Chayes, Sagun, and Zecchina]{chaudhari2016entropy}
Chaudhari, P., Choromanska, A., Soatto, S., LeCun, Y., Baldassi, C., Borgs, C.,
  Chayes, J., Sagun, L., and Zecchina, R.
\newblock Entropy-sgd: Biasing gradient descent into wide valleys.
\newblock \emph{International Conference on Learning Representations}, 2017.

\bibitem[Chou et~al.(2017)Chou, Maturana, and Scherer]{chou2017improving}
Chou, P.-W., Maturana, D., and Scherer, S.
\newblock Improving stochastic policy gradients in continuous control with deep
  reinforcement learning using the beta distribution.
\newblock In \emph{International Conference on Machine Learning}, pp.\
  834--843, 2017.

\bibitem[Dauphin et~al.(2014)Dauphin, Pascanu, Gulcehre, Cho, Ganguli, and
  Bengio]{dauphin2014identifying}
Dauphin, Y.~N., Pascanu, R., Gulcehre, C., Cho, K., Ganguli, S., and Bengio, Y.
\newblock Identifying and attacking the saddle point problem in
  high-dimensional non-convex optimization.
\newblock In \emph{Advances in neural information processing systems}, pp.\
  2933--2941, 2014.

\bibitem[Draxler et~al.(2018)Draxler, Veschgini, Salmhofer, and
  Hamprecht]{draxler2018essentially}
Draxler, F., Veschgini, K., Salmhofer, M., and Hamprecht, F.~A.
\newblock Essentially no barriers in neural network energy landscape.
\newblock \emph{International Conference on Machine Learning}, 2018.

\bibitem[Fort \& Scherlis(2018)Fort and Scherlis]{fort2018goldilocks}
Fort, S. and Scherlis, A.
\newblock The goldilocks zone: Towards better understanding of neural network
  loss landscapes.
\newblock \emph{Thirty-Third AAAI Conference on Artificial Intelligence
  (AAAI-19)}, 2018.

\bibitem[Fujimoto et~al.(2018)Fujimoto, van Hoof, and
  Meger]{fujimoto2018addressing}
Fujimoto, S., van Hoof, H., and Meger, D.
\newblock Addressing function approximation error in actor-critic methods.
\newblock \emph{International Conference on Machine Learning}, 2018.

\bibitem[Fujita \& Maeda(2018)Fujita and Maeda]{fujita2018clipped}
Fujita, Y. and Maeda, S.-i.
\newblock Clipped action policy gradient.
\newblock \emph{International Conference on Machine Learning}, 2018.

\bibitem[Goodfellow et~al.(2015)Goodfellow, Vinyals, and
  Saxe]{goodfellow2014qualitatively}
Goodfellow, I.~J., Vinyals, O., and Saxe, A.~M.
\newblock Qualitatively characterizing neural network optimization problems.
\newblock \emph{International Conference on Learning Representations}, 2015.

\bibitem[Greensmith et~al.(2004)Greensmith, Bartlett, and
  Baxter]{greensmith2004variance}
Greensmith, E., Bartlett, P.~L., and Baxter, J.
\newblock Variance reduction techniques for gradient estimates in reinforcement
  learning.
\newblock \emph{Journal of Machine Learning Research}, 5\penalty0
  (Nov):\penalty0 1471--1530, 2004.

\bibitem[Gulcehre et~al.(2017)Gulcehre, Moczulski, Visin, and
  Bengio]{gulcehre2016mollifying}
Gulcehre, C., Moczulski, M., Visin, F., and Bengio, Y.
\newblock Mollifying networks.
\newblock \emph{International Conference on Learning Representations 2017},
  2017.

\bibitem[Henderson et~al.(2018)Henderson, Romoff, and Pineau]{henderson2018did}
Henderson, P., Romoff, J., and Pineau, J.
\newblock Where did my optimum go?: An empirical analysis of gradient descent
  optimization in policy gradient methods.
\newblock \emph{European Workshop on Reinforcement Learning}, 2018.

\bibitem[Ilyas et~al.(2018)Ilyas, Engstrom, Santurkar, Tsipras, Janoos,
  Rudolph, and Madry]{ilyas2018deep}
Ilyas, A., Engstrom, L., Santurkar, S., Tsipras, D., Janoos, F., Rudolph, L.,
  and Madry, A.
\newblock Are deep policy gradient algorithms truly policy gradient algorithms?
\newblock \emph{arXiv preprint arXiv:1811.02553}, 2018.

\bibitem[Kakade(2001)]{kakade2001natural}
Kakade, S.
\newblock A natural policy gradient.
\newblock In \emph{Neural Information Processing Systems}, volume~14, pp.\
  1531--1538, 2001.

\bibitem[Keskar et~al.(2017)Keskar, Mudigere, Nocedal, Smelyanskiy, and
  Tang]{keskar2016large}
Keskar, N.~S., Mudigere, D., Nocedal, J., Smelyanskiy, M., and Tang, P. T.~P.
\newblock On large-batch training for deep learning: Generalization gap and
  sharp minima.
\newblock \emph{International Conference on Learning Representations 2017},
  2017.

\bibitem[Khetarpal et~al.(2018)Khetarpal, Ahmed, Cianflone, Islam, and
  Pineau]{khetarpal2018re}
Khetarpal, K., Ahmed, Z., Cianflone, A., Islam, R., and Pineau, J.
\newblock Re-evaluate: Reproducibility in evaluating reinforcement learning
  algorithms.
\newblock \emph{2nd Reproducibility in Machine Learning Workshop at ICML 2018},
  2018.

\bibitem[Konda \& Tsitsiklis(2000)Konda and Tsitsiklis]{konda2000actor}
Konda, V.~R. and Tsitsiklis, J.~N.
\newblock Actor-critic algorithms.
\newblock In \emph{Advances in neural information processing systems}, pp.\
  1008--1014, 2000.

\bibitem[Le~Cun et~al.(1991)Le~Cun, Kanter, and Solla]{le1991eigenvalues}
Le~Cun, Y., Kanter, I., and Solla, S.~A.
\newblock Eigenvalues of covariance matrices: Application to neural-network
  learning.
\newblock \emph{Physical Review Letters}, 66\penalty0 (18):\penalty0 2396,
  1991.

\bibitem[Li et~al.(2018{\natexlab{a}})Li, Farkhoor, Liu, and
  Yosinski]{li2018measuring}
Li, C., Farkhoor, H., Liu, R., and Yosinski, J.
\newblock Measuring the intrinsic dimension of objective landscapes.
\newblock \emph{International Conference on Learning Representations (ICLR)},
  2018{\natexlab{a}}.

\bibitem[Li et~al.(2018{\natexlab{b}})Li, Xu, Taylor, and
  Goldstein]{li2017visualizing}
Li, H., Xu, Z., Taylor, G., and Goldstein, T.
\newblock Visualizing the loss landscape of neural nets.
\newblock \emph{International Conference on Learning Representations, Workshop
  Track}, 2018{\natexlab{b}}.

\bibitem[Mandt et~al.(2017)Mandt, Hoffman, and Blei]{mandt2017stochastic}
Mandt, S., Hoffman, M.~D., and Blei, D.~M.
\newblock Stochastic gradient descent as approximate bayesian inference.
\newblock \emph{The Journal of Machine Learning Research}, 18\penalty0
  (1):\penalty0 4873--4907, 2017.

\bibitem[Mnih et~al.(2016)Mnih, Badia, Mirza, Graves, Lillicrap, Harley,
  Silver, and Kavukcuoglu]{mnih2016asynchronous}
Mnih, V., Badia, A.~P., Mirza, M., Graves, A., Lillicrap, T., Harley, T.,
  Silver, D., and Kavukcuoglu, K.
\newblock Asynchronous methods for deep reinforcement learning.
\newblock In \emph{International conference on machine learning}, pp.\
  1928--1937, 2016.

\bibitem[Nachum et~al.(2018)Nachum, Norouzi, Tucker, and
  Schuurmans]{nachum2018smoothed}
Nachum, O., Norouzi, M., Tucker, G., and Schuurmans, D.
\newblock Smoothed action value functions for learning gaussian policies.
\newblock \emph{International Conference on Machine Learning}, 2018.

\bibitem[Neu et~al.(2017)Neu, Jonsson, and G{\'o}mez]{neu2017unified}
Neu, G., Jonsson, A., and G{\'o}mez, V.
\newblock A unified view of entropy-regularized markov decision processes.
\newblock \emph{arXiv preprint arXiv:1705.07798}, 2017.

\bibitem[Rajeswaran et~al.(2017)Rajeswaran, Lowrey, Todorov, and
  Kakade]{rajeswaran2017towards}
Rajeswaran, A., Lowrey, K., Todorov, E.~V., and Kakade, S.~M.
\newblock Towards generalization and simplicity in continuous control.
\newblock In \emph{Advances in Neural Information Processing Systems}, pp.\
  6550--6561, 2017.

\bibitem[Santurkar et~al.(2018)Santurkar, Tsipras, Ilyas, and
  Madry]{santurkar2018does}
Santurkar, S., Tsipras, D., Ilyas, A., and Madry, A.
\newblock How does batch normalization help optimization?(no, it is not about
  internal covariate shift).
\newblock \emph{arXiv preprint arXiv:1805.11604}, 2018.

\bibitem[Schulman et~al.(2015{\natexlab{a}})Schulman, Levine, Abbeel, Jordan,
  and Moritz]{schulman2015trust}
Schulman, J., Levine, S., Abbeel, P., Jordan, M., and Moritz, P.
\newblock Trust region policy optimization.
\newblock In \emph{International Conference on Machine Learning}, pp.\
  1889--1897, 2015{\natexlab{a}}.

\bibitem[Schulman et~al.(2015{\natexlab{b}})Schulman, Moritz, Levine, Jordan,
  and Abbeel]{schulman2015high}
Schulman, J., Moritz, P., Levine, S., Jordan, M., and Abbeel, P.
\newblock High-dimensional continuous control using generalized advantage
  estimation.
\newblock \emph{arXiv preprint arXiv:1506.02438}, 2015{\natexlab{b}}.

\bibitem[Schulman et~al.(2017{\natexlab{a}})Schulman, Chen, and
  Abbeel]{schulman2017equivalence}
Schulman, J., Chen, X., and Abbeel, P.
\newblock Equivalence between policy gradients and soft q-learning.
\newblock \emph{arXiv preprint arXiv:1704.06440}, 2017{\natexlab{a}}.

\bibitem[Schulman et~al.(2017{\natexlab{b}})Schulman, Wolski, Dhariwal,
  Radford, and Klimov]{schulman2017proximal}
Schulman, J., Wolski, F., Dhariwal, P., Radford, A., and Klimov, O.
\newblock Proximal policy optimization algorithms.
\newblock \emph{arXiv preprint arXiv:1707.06347}, 2017{\natexlab{b}}.

\bibitem[Silver et~al.(2014)Silver, Lever, Heess, Degris, Wierstra, and
  Riedmiller]{silver2014deterministic}
Silver, D., Lever, G., Heess, N., Degris, T., Wierstra, D., and Riedmiller, M.
\newblock Deterministic policy gradient algorithms.
\newblock In \emph{International Conference on Machine Learning}, 2014.

\bibitem[Sutton et~al.(2000)Sutton, McAllester, Singh, and
  Mansour]{sutton2000policy}
Sutton, R.~S., McAllester, D.~A., Singh, S.~P., and Mansour, Y.
\newblock Policy gradient methods for reinforcement learning with function
  approximation.
\newblock In \emph{Advances in neural information processing systems}, pp.\
  1057--1063, 2000.

\bibitem[Todorov et~al.(2012)Todorov, Erez, and Tassa]{todorov2012mujoco}
Todorov, E., Erez, T., and Tassa, Y.
\newblock Mujoco: A physics engine for model-based control.
\newblock In \emph{International Conference on Intelligent Robots and Systems
  (IROS)}, pp.\  5026--5033. IEEE, 2012.

\bibitem[Tucker et~al.(2018)Tucker, Bhupatiraju, Gu, Turner, Ghahramani, and
  Levine]{tucker2018mirage}
Tucker, G., Bhupatiraju, S., Gu, S., Turner, R.~E., Ghahramani, Z., and Levine,
  S.
\newblock The mirage of action-dependent baselines in reinforcement learning.
\newblock \emph{International Conference on Learning Representations (Workshop
  Track)}, 2018.

\bibitem[Watkins \& Dayan(1992)Watkins and Dayan]{watkins1992q}
Watkins, C.~J. and Dayan, P.
\newblock Q-learning.
\newblock \emph{Machine learning}, 8\penalty0 (3-4):\penalty0 279--292, 1992.

\bibitem[Williams(1992)]{williams1992simple}
Williams, R.~J.
\newblock Simple statistical gradient-following algorithms for connectionist
  reinforcement learning.
\newblock \emph{Machine learning}, 8\penalty0 (3-4):\penalty0 229--256, 1992.

\bibitem[Williams \& Peng(1991)Williams and Peng]{williams1991function}
Williams, R.~J. and Peng, J.
\newblock Function optimization using connectionist reinforcement learning
  algorithms.
\newblock \emph{Connection Science}, 3\penalty0 (3):\penalty0 241--268, 1991.

\bibitem[Xiao et~al.(2017)Xiao, Rasul, and Vollgraf]{xiao2017fashion}
Xiao, H., Rasul, K., and Vollgraf, R.
\newblock Fashion-mnist: a novel image dataset for benchmarking machine
  learning algorithms.
\newblock \emph{arXiv preprint arXiv:1708.07747}, 2017.

\bibitem[Zhao et~al.(2011)Zhao, Hachiya, Niu, and Sugiyama]{zhao2011analysis}
Zhao, T., Hachiya, H., Niu, G., and Sugiyama, M.
\newblock Analysis and improvement of policy gradient estimation.
\newblock In \emph{Advances in Neural Information Processing Systems}, pp.\
  262--270, 2011.

\end{thebibliography}
\bibliographystyle{icml2019}

\newpage
\clearpage
\onecolumn
\appendix

\renewcommand\thefigure{\thesection\arabic{figure}}    
\setcounter{figure}{0}
\setcounter{section}{18}
\section{Appendix}

\begin{figure*}[b]
    \centering
    \begin{subfigure}[b]{0.45\textwidth}
        \centering
        \includegraphics[width=\textwidth]{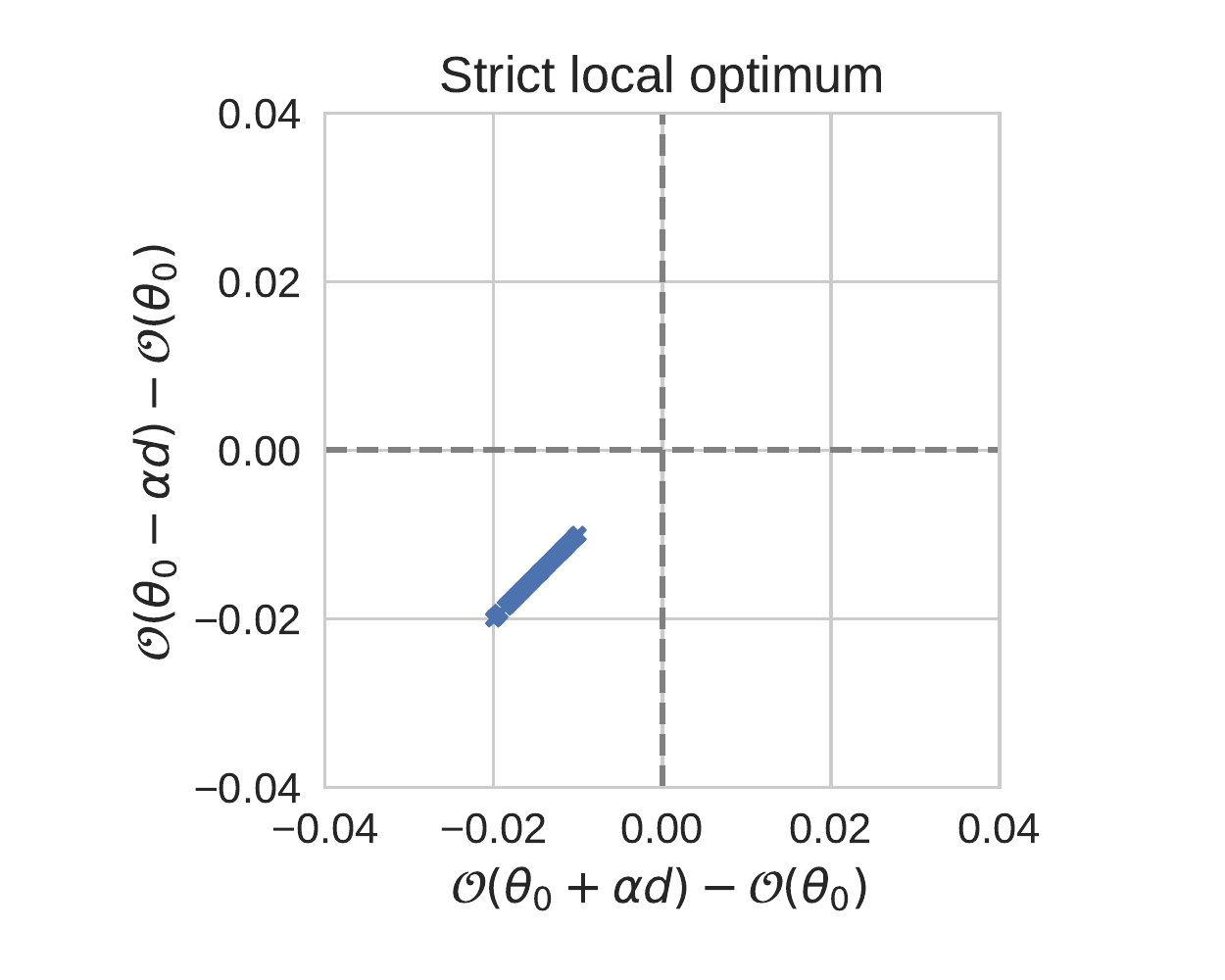}
        \caption{$-x^2+0y^2$ at $(x,y)=(0,0)$}
    \end{subfigure}
    \begin{subfigure}[b]{0.45\textwidth}
        \centering
        \includegraphics[width=\textwidth]{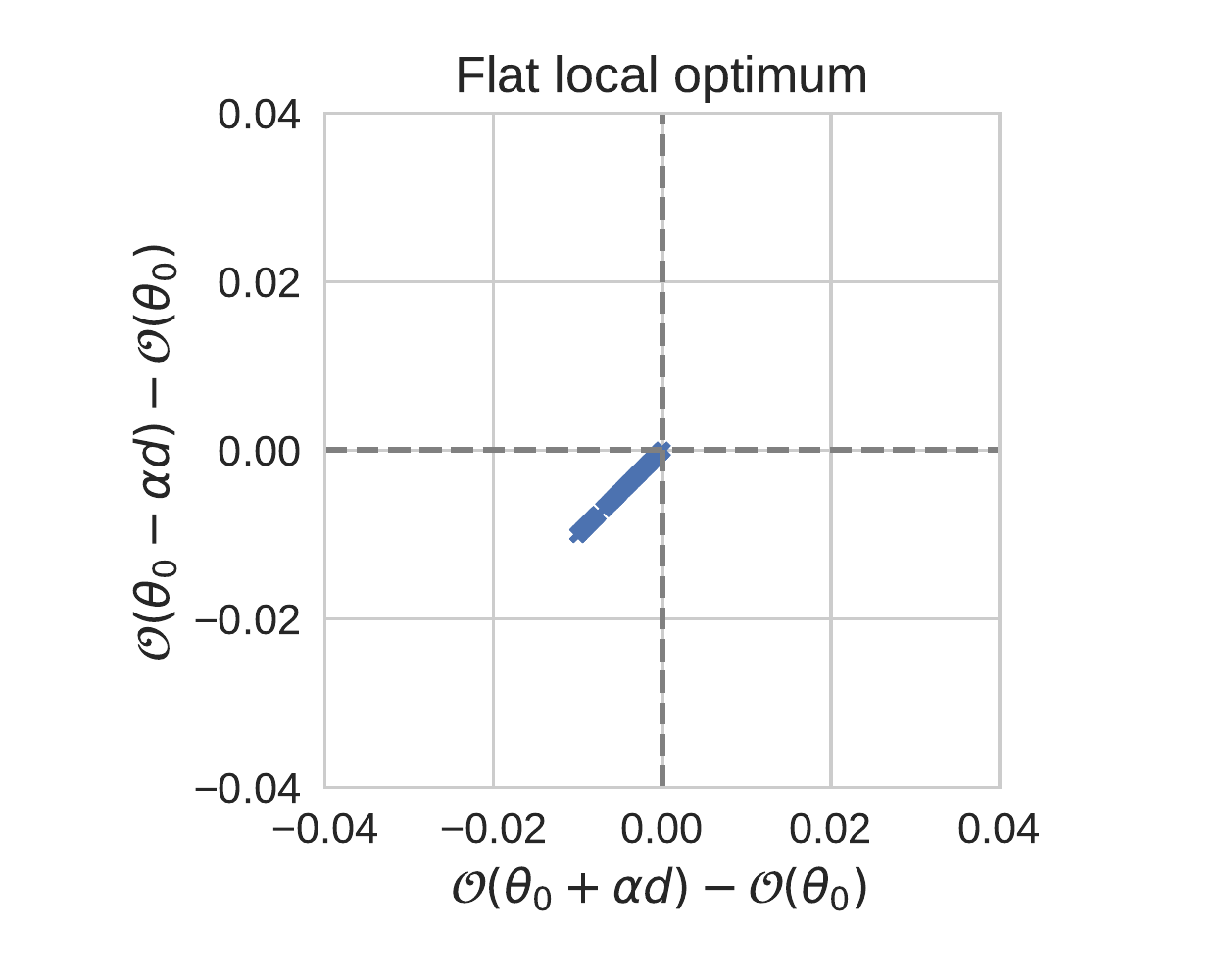}
        \caption{$-x^2-2y^2$ at $(x,y)=(0,0)$}
    \end{subfigure}
    \caption{Example visualizations of the random perturbation method of local optima in simple loss functions. Scatter plots can distinguish between strict local optimum (where all directions are negative and have negative curvature) with a flat optimum (where some directions might have 0 curvature.}
    \label{sfig:LM_example_visualizations}
\end{figure*}

\begin{figure*}[b]
    \centering
    \begin{subfigure}[b]{0.32\textwidth}
        \centering
        \includegraphics[width=\textwidth]{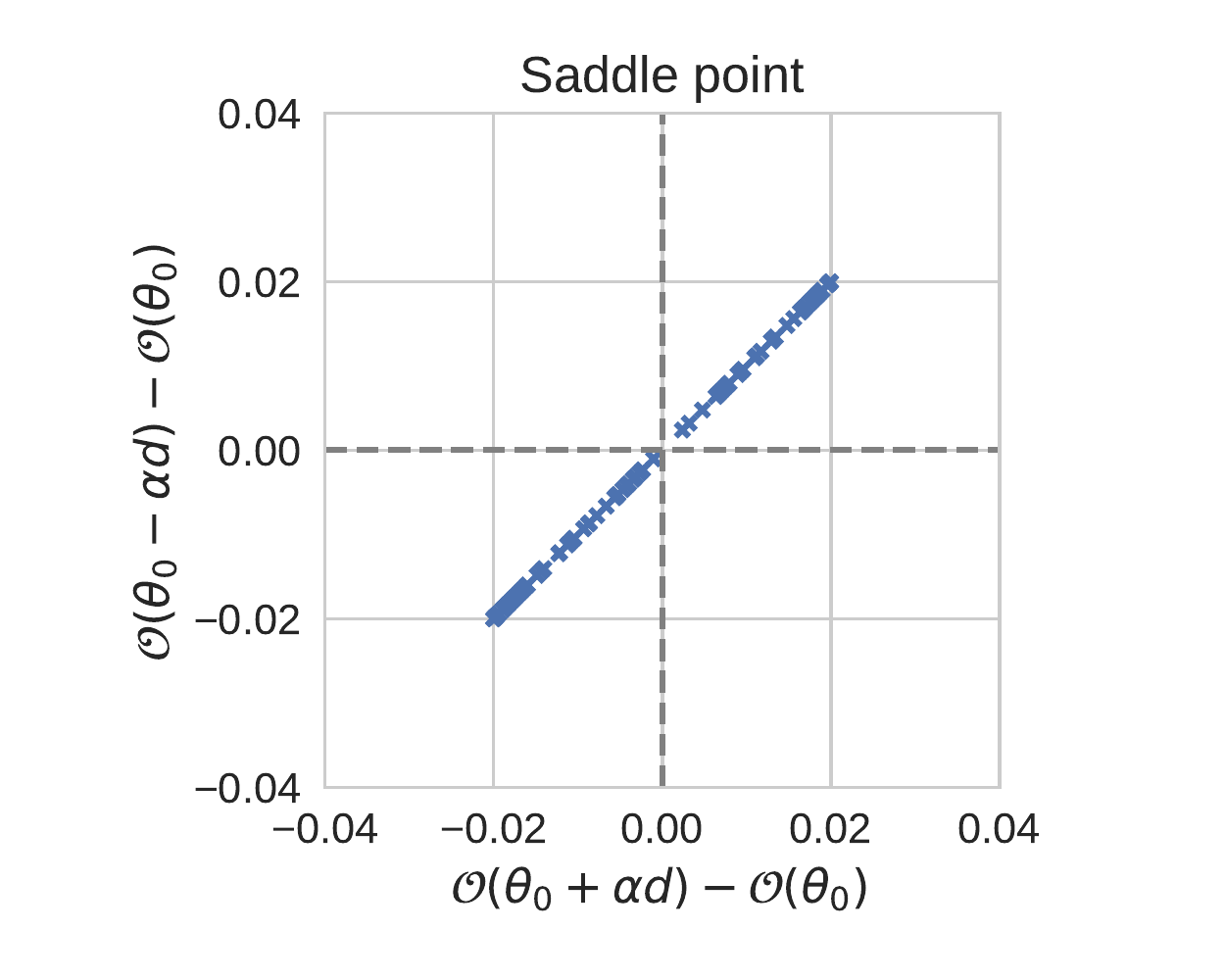}
        \caption{$-x^2+y^2$ at $(x,y)=(0,0)$}
    \end{subfigure}
    \begin{subfigure}[b]{0.32\textwidth}
        \centering
        \includegraphics[width=\textwidth]{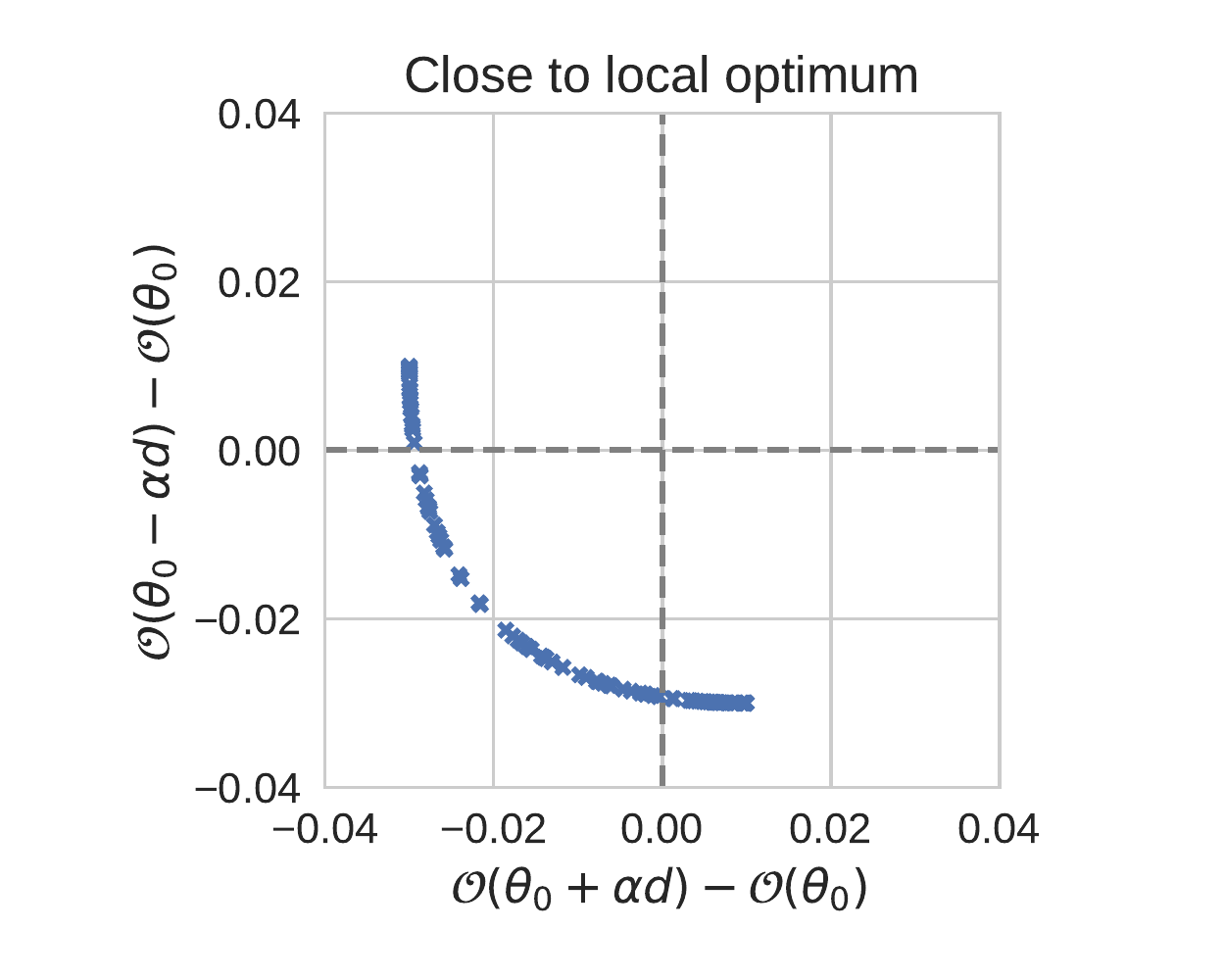}
        \caption{$-x^2-y^2$ at $(x,y)=(0.1, 0.0)$}
    \end{subfigure}
    \begin{subfigure}[b]{0.32\textwidth}
        \centering
        \includegraphics[width=\textwidth]{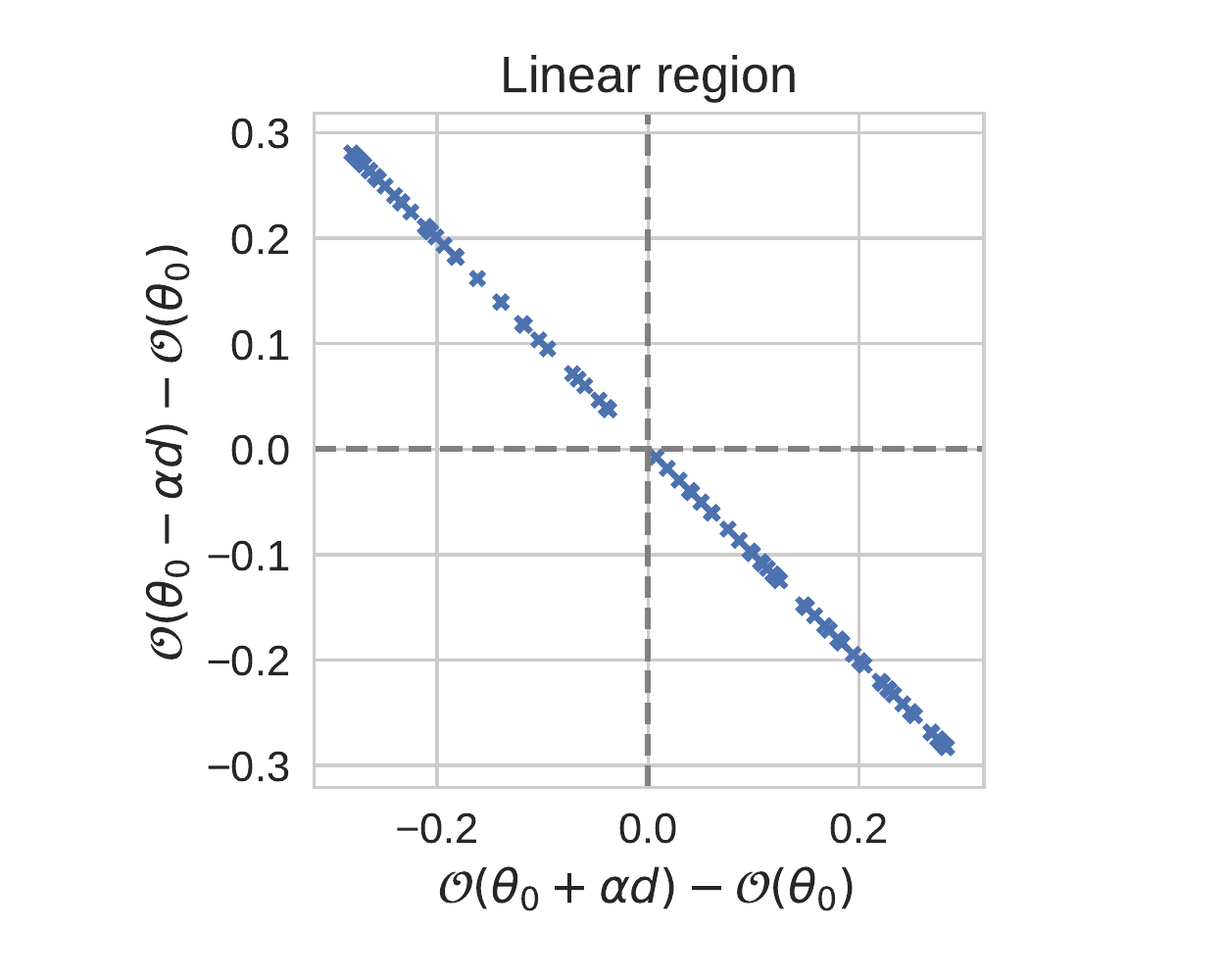}
        \caption{$-2x+2y$ at $(x,y)=(0,0)$}
    \end{subfigure}
    
    \caption{Example visualizations of the random perturbation method of saddle points, linear regions in simple loss functions.}
    \label{sfig:linear_example_visualizations}
\end{figure*}

\begin{figure*}[ht]
    \vskip 0.2in
    \begin{center}
    \begin{subfigure}[b]{0.44\textwidth}
        \centering
        \includegraphics[width=\textwidth]{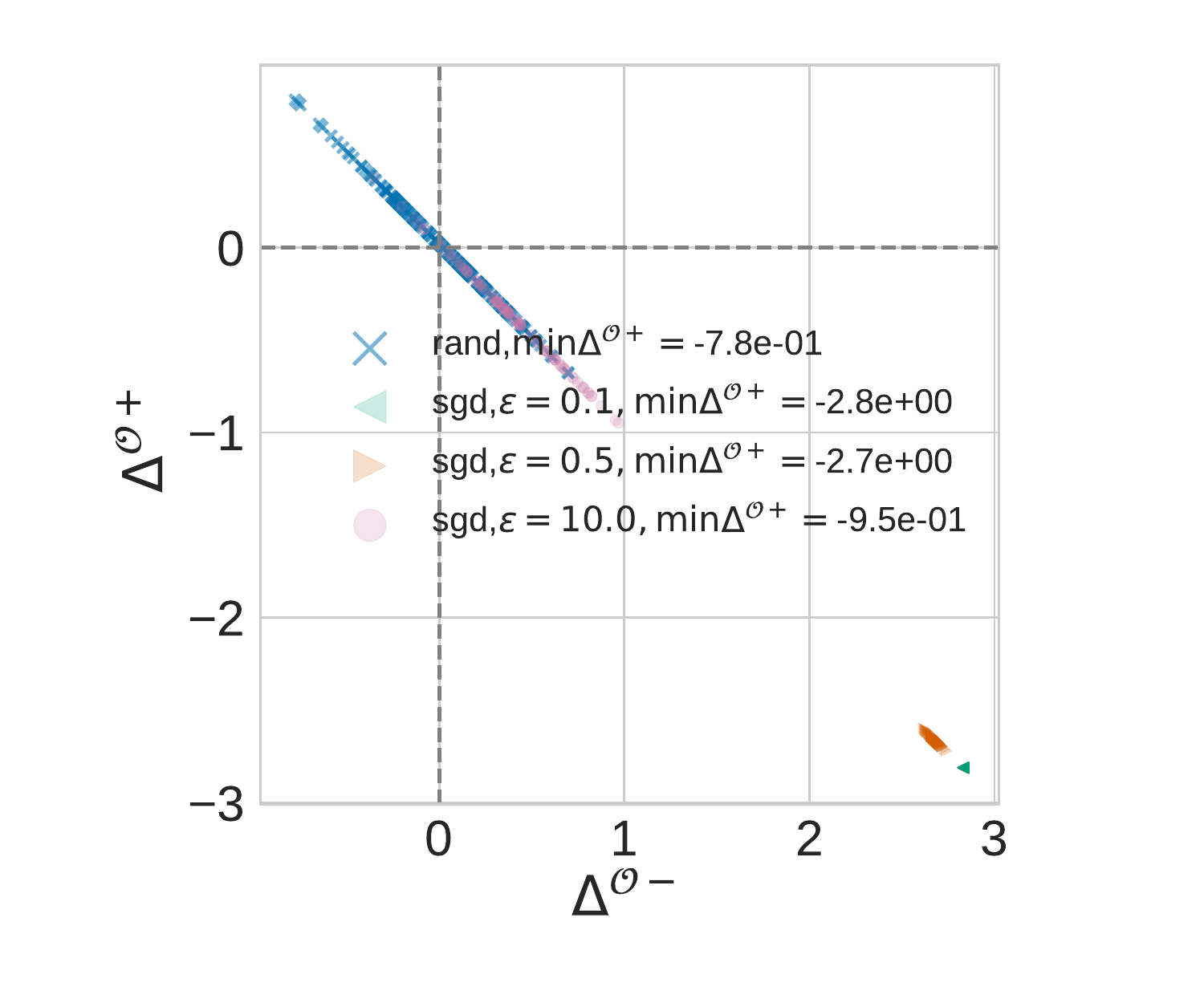}
        \caption{$k_1=50, k_2=50$}
    \end{subfigure}
    \begin{subfigure}[b]{0.44\textwidth}
        \centering
        \includegraphics[width=\textwidth]{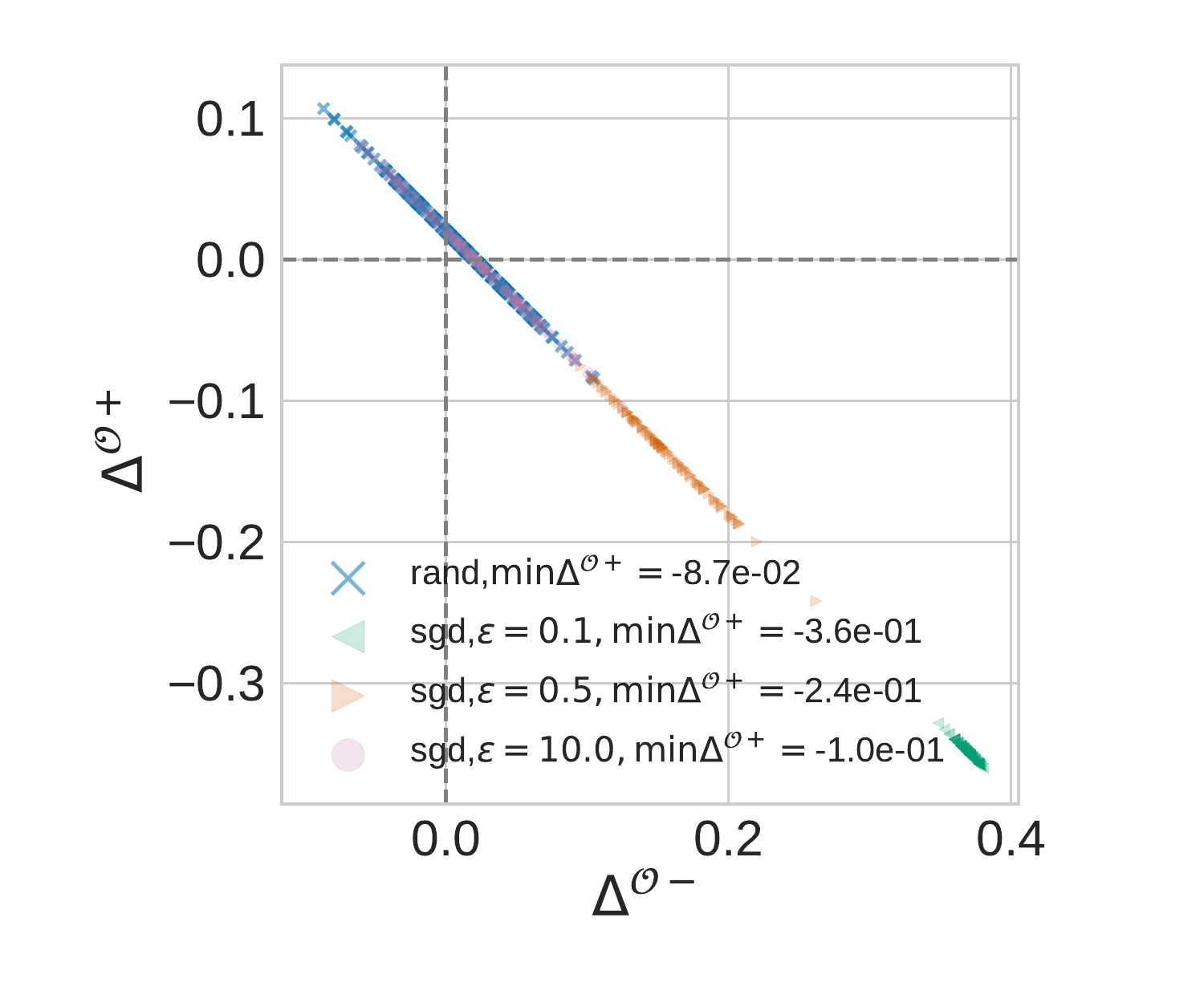}
        \caption{$k_1=99, k_2=1$}
    \end{subfigure}
    \begin{subfigure}[b]{0.44\textwidth}
        \centering
        \includegraphics[width=\textwidth]{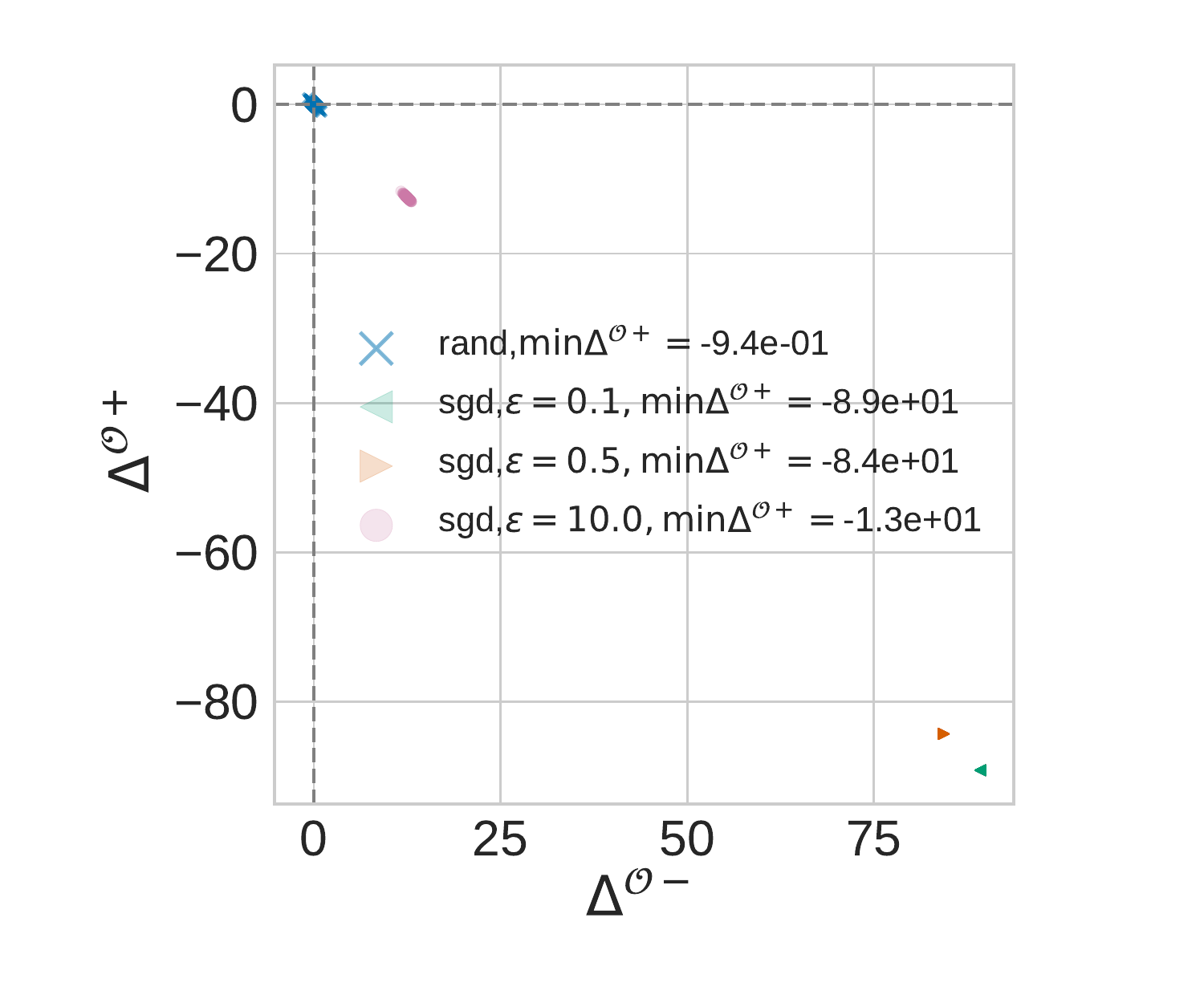}
        \caption{$k_1=50000, k_2=50000$}
    \end{subfigure}
    \begin{subfigure}[b]{0.44\textwidth}
        \centering
        \includegraphics[width=\textwidth]{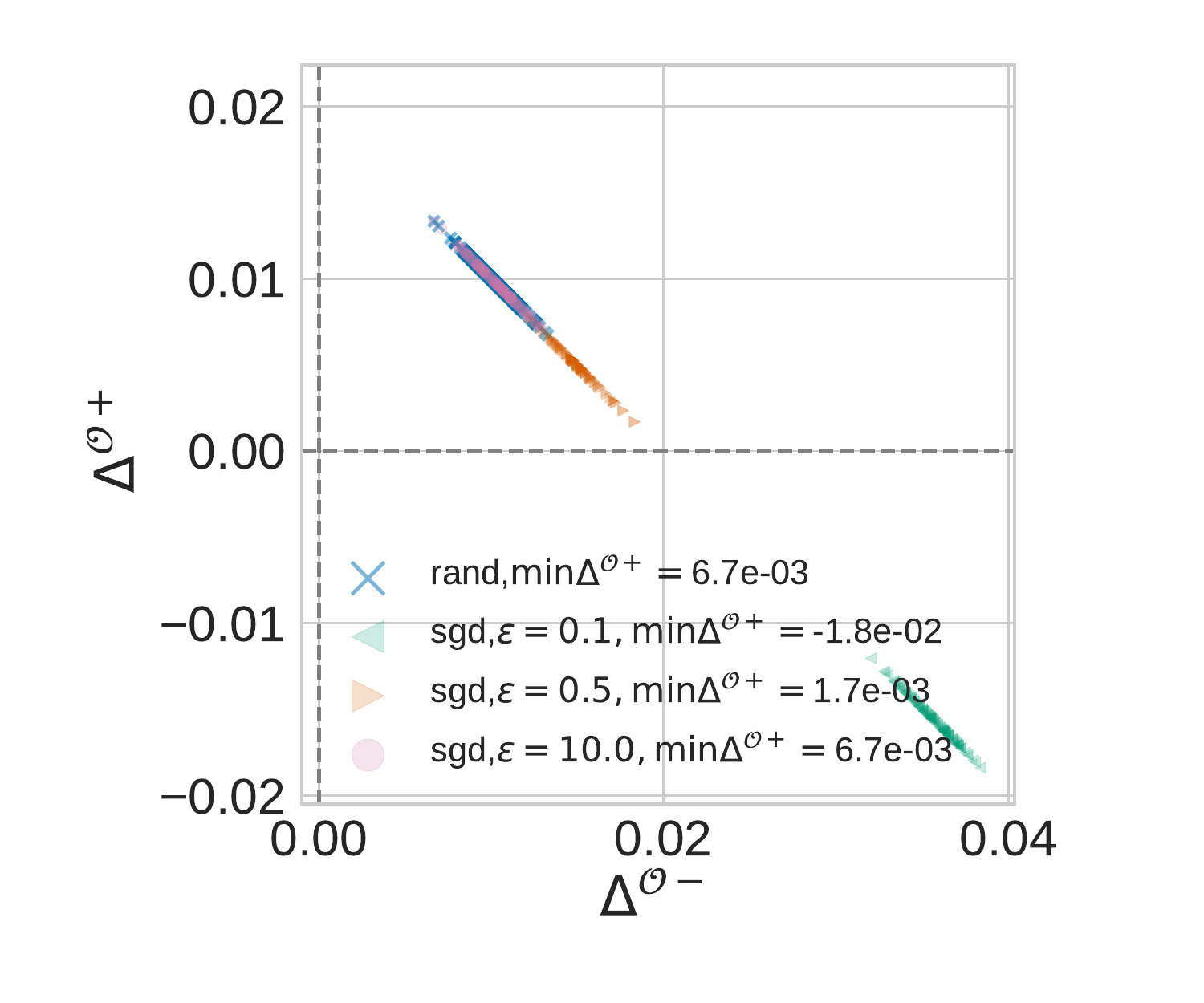}
        \caption{$k_1=99999, k_2=1$}
    \end{subfigure}
    \caption{{\small \textbf{Assessing the limitations of random perturbations.} We repeat the sampling procedure using directions given by stochastic gradients rather than random directions. For small models, $k_1+k_2=100$, our method can correctly recover all directions of descent. For larger models, $k_1+k_2=10000$, we find that when the number of descent directions, $k_2$, is small both methods will miss this direction. It is only if the gradient noise $\epsilon$ is small will it capture the descent direction. We numerically verify that running stochastic gradient descent from this point does not lead to a solution in a reasonable number of iterations.}}
    \label{sfig:random_perturbations_limits}
    \end{center}
    \vskip -0.2in
\end{figure*}

\begin{figure*}[ht]
    \vskip 0.2in
    \begin{center}
    \begin{subfigure}[b]{0.24\textwidth}
        \centering
        \includegraphics[width=\textwidth]{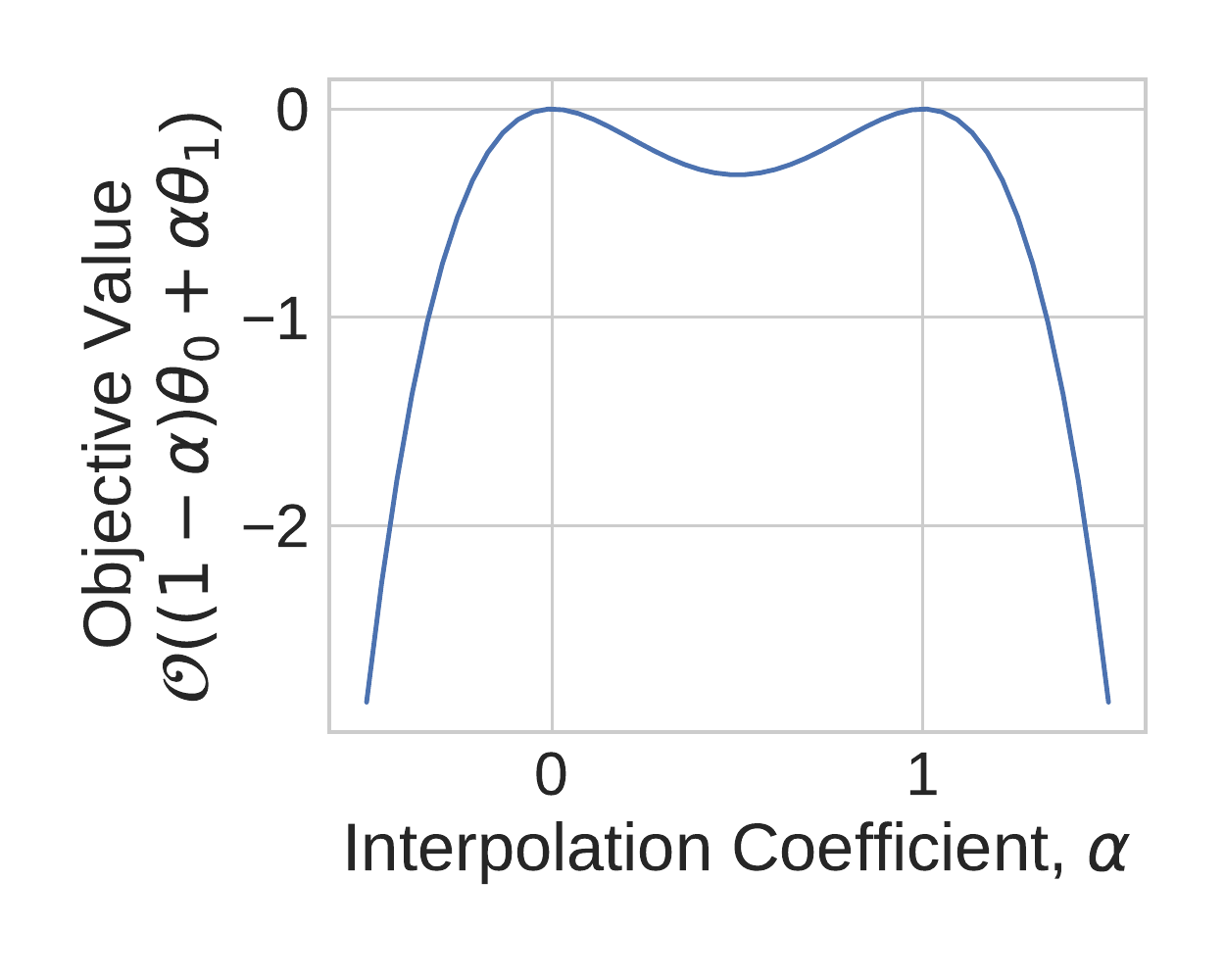}
        \caption{Linear Interpolation}
    \end{subfigure}
    \begin{subfigure}[b]{0.24\textwidth}
        \centering
        \includegraphics[width=\textwidth]{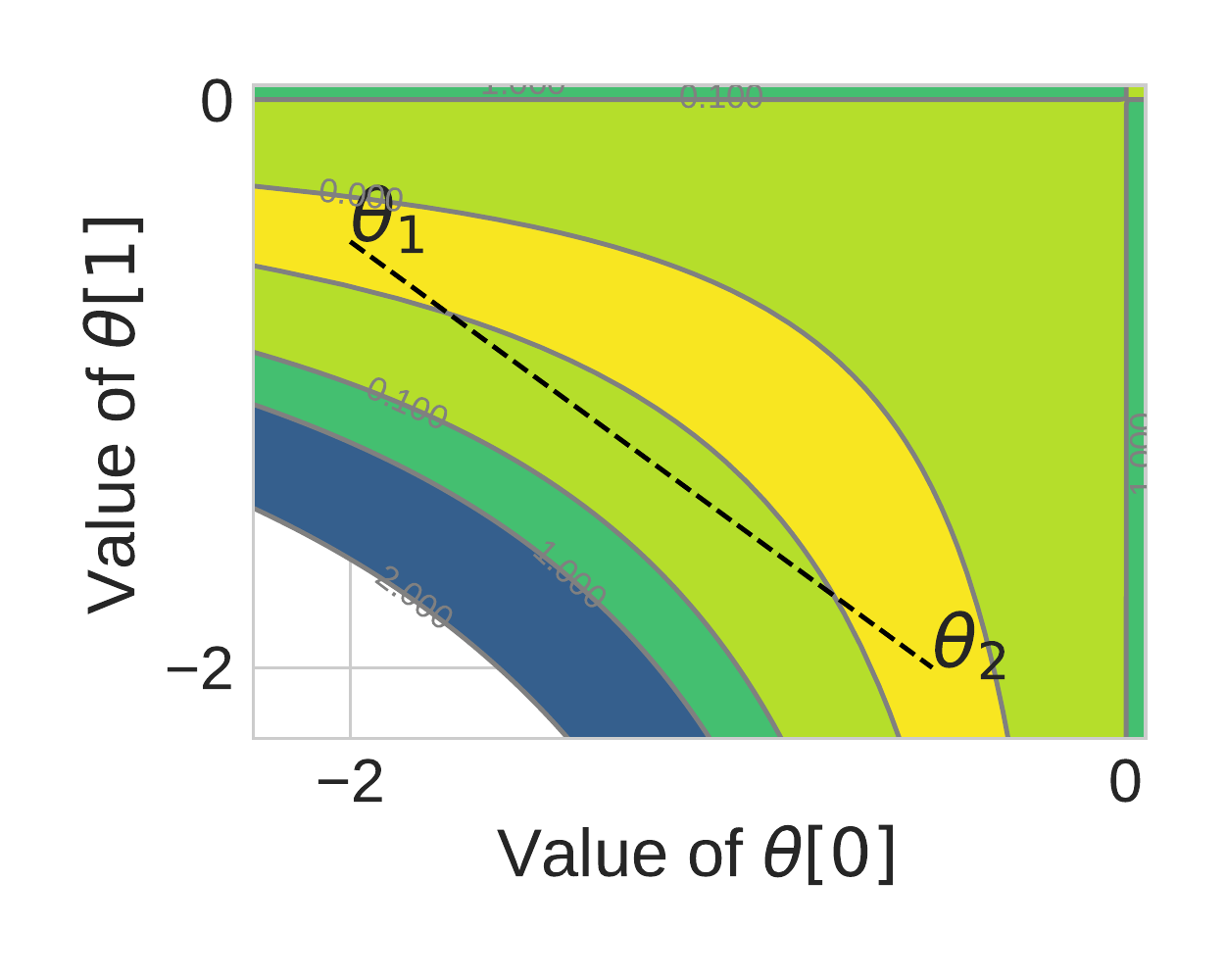}
        \caption{2D Plot of $\Oh$}
    \end{subfigure}
    \begin{subfigure}[b]{0.24\textwidth}
        \centering
        \includegraphics[width=\textwidth]{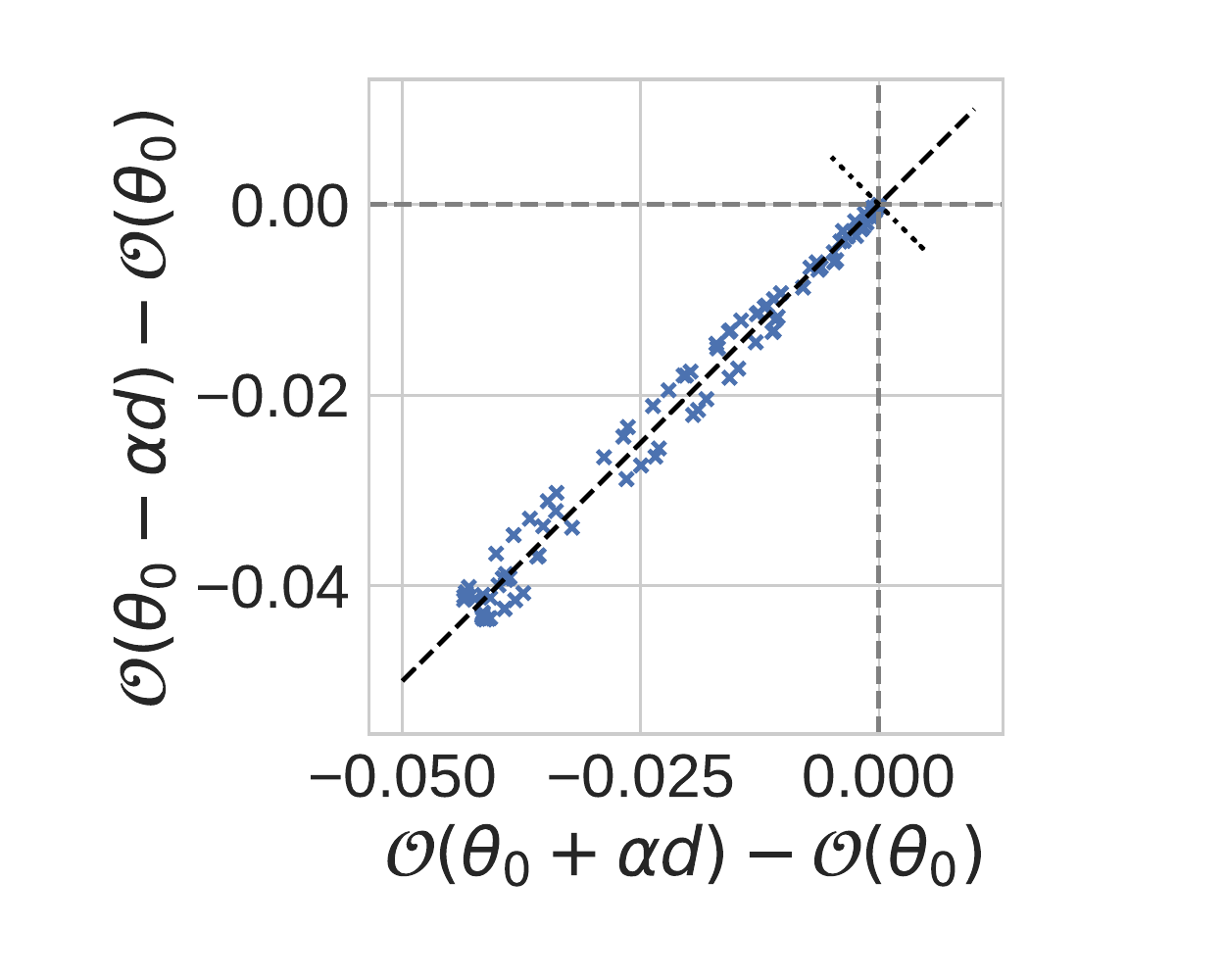}
        \caption{$\{\Delta^{\Oh^+}, \Delta^{\Oh^-}\}$ at $\theta_0$}
    \end{subfigure}
    \begin{subfigure}[b]{0.24\textwidth}
        \centering
        \includegraphics[width=\textwidth]{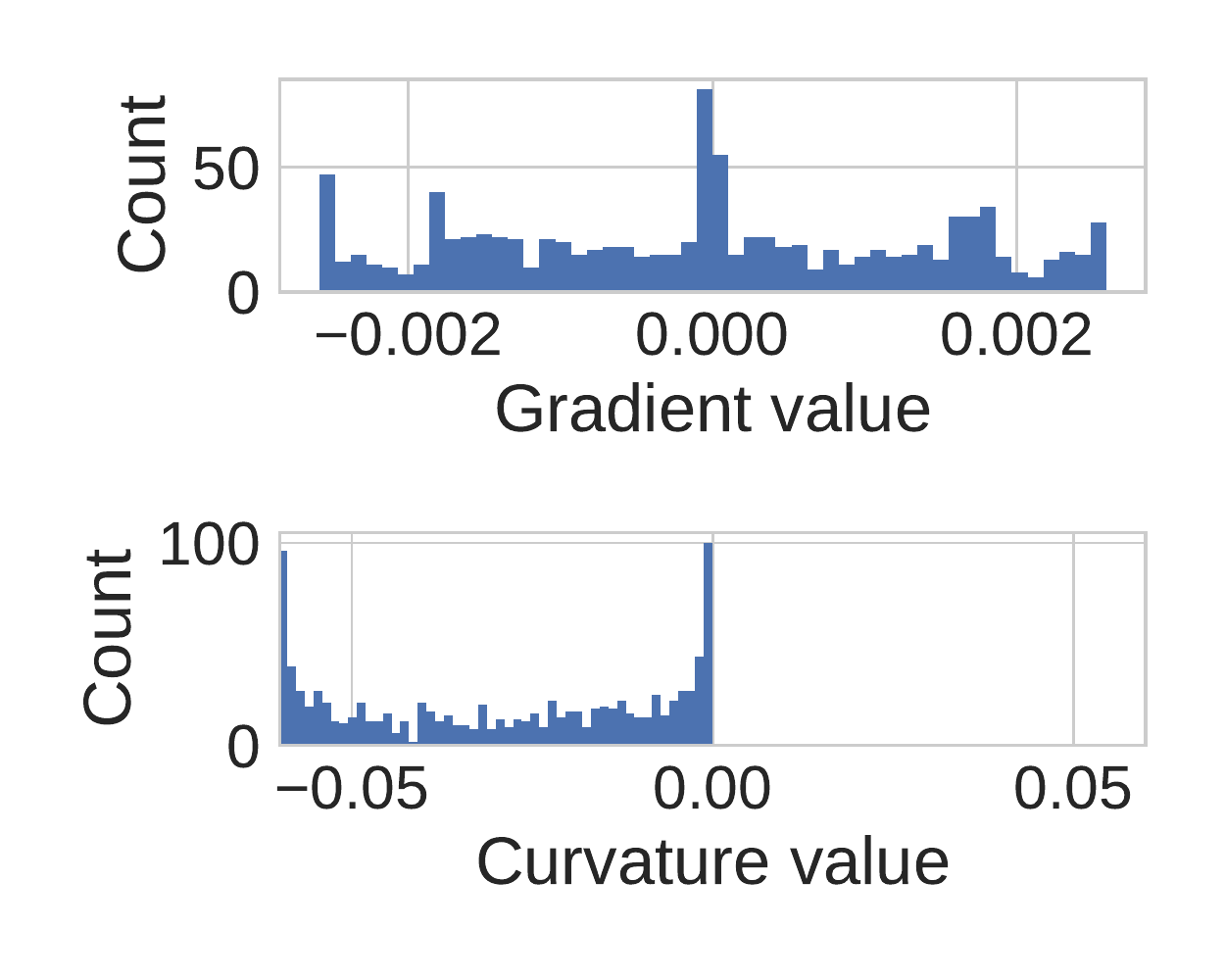}
        \caption{Projection of (c)}
    \end{subfigure}
    \caption{{\small \textbf{Summary of the methods on $\Oh(\theta)=-(1-\theta[0]\theta[1])^2$} (a) A linear interpolation between two local maxima, $\theta_0=(-0.5, -2)$ and $\theta_1=(-2, -0.5)$ suggests that these maxima are isolated. (b) A contour plot of $\Oh$ shows that the linear interpolation (dashed) goes through a region of high $\Oh$, and $\theta_0$ and $\theta_1$ are not isolated but are connected by a low value of $\Oh$. (c) Random perturbations around $\theta_0$ show that many directions lead to a decrease in $\Oh$ indicating a local maxima and some directions have near zero change indicating flatness. (d) Projecting the points from (c) onto the two axes (dotted) gives us density plots for the gradient and curvature. The values on the gradient spectra are close to zero, indicating it is a critical point. The curvature spectra shows some negative curvature (local maximum) and some zero curvature (flatness). See Section~\ref{approach_obj_funcs} for detailed explanation.}}
    \label{fig:method}
    \end{center}
    \vskip -0.2in
\end{figure*}

\begin{figure}[b]
    \centering
    \includegraphics[width=0.5\textwidth]{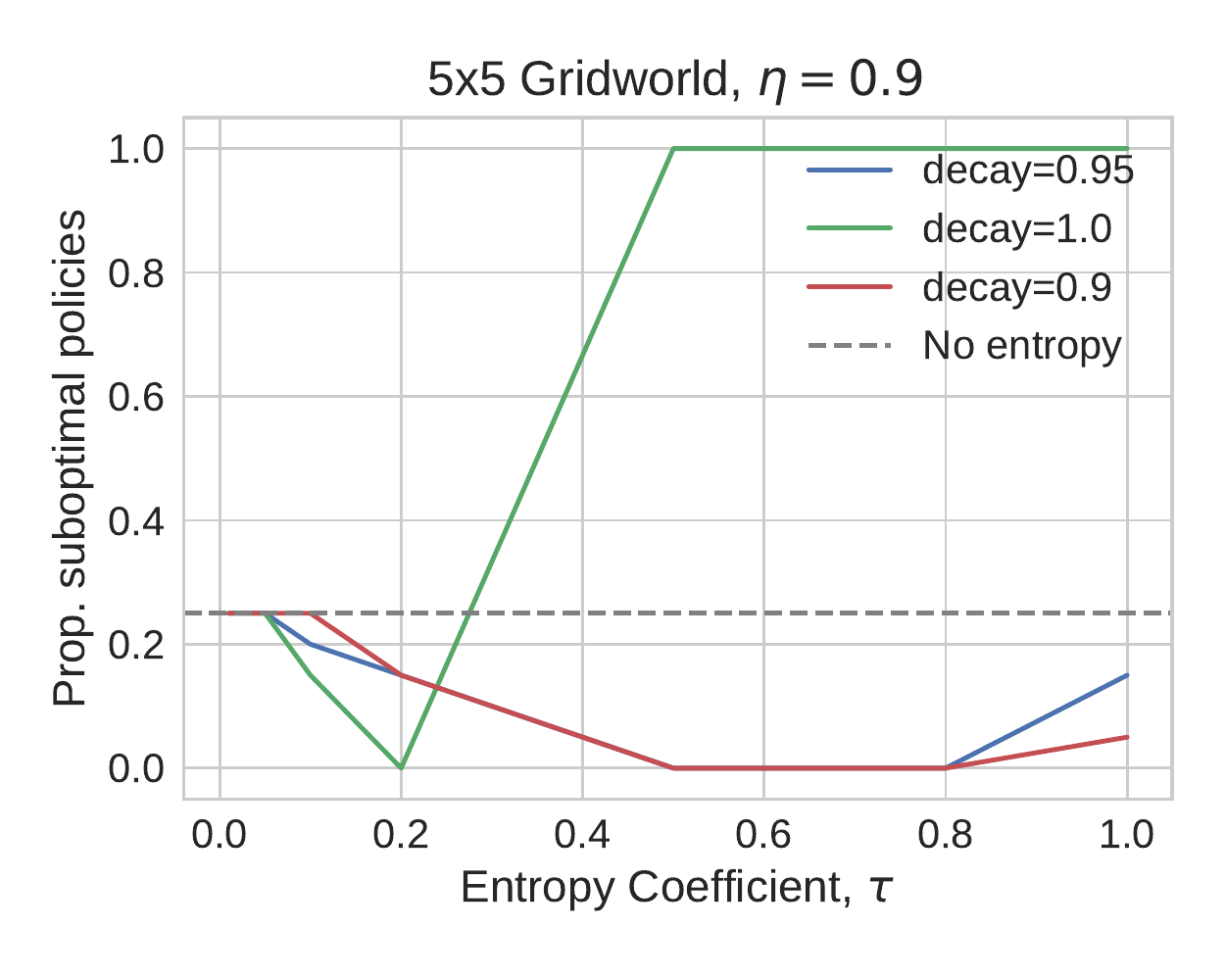}
    \caption{Proportion of sub-optimal solutions found for different entropy coefficients $\tau$ and decay factors. Using an entropy coefficient helps learn the optimal policy.}
    \label{sfig:entropy_solutions}
\end{figure}

\begin{figure}
    \centering
     \begin{subfigure}[b]{0.45\textwidth}
        \centering
        \includegraphics[width=\textwidth]{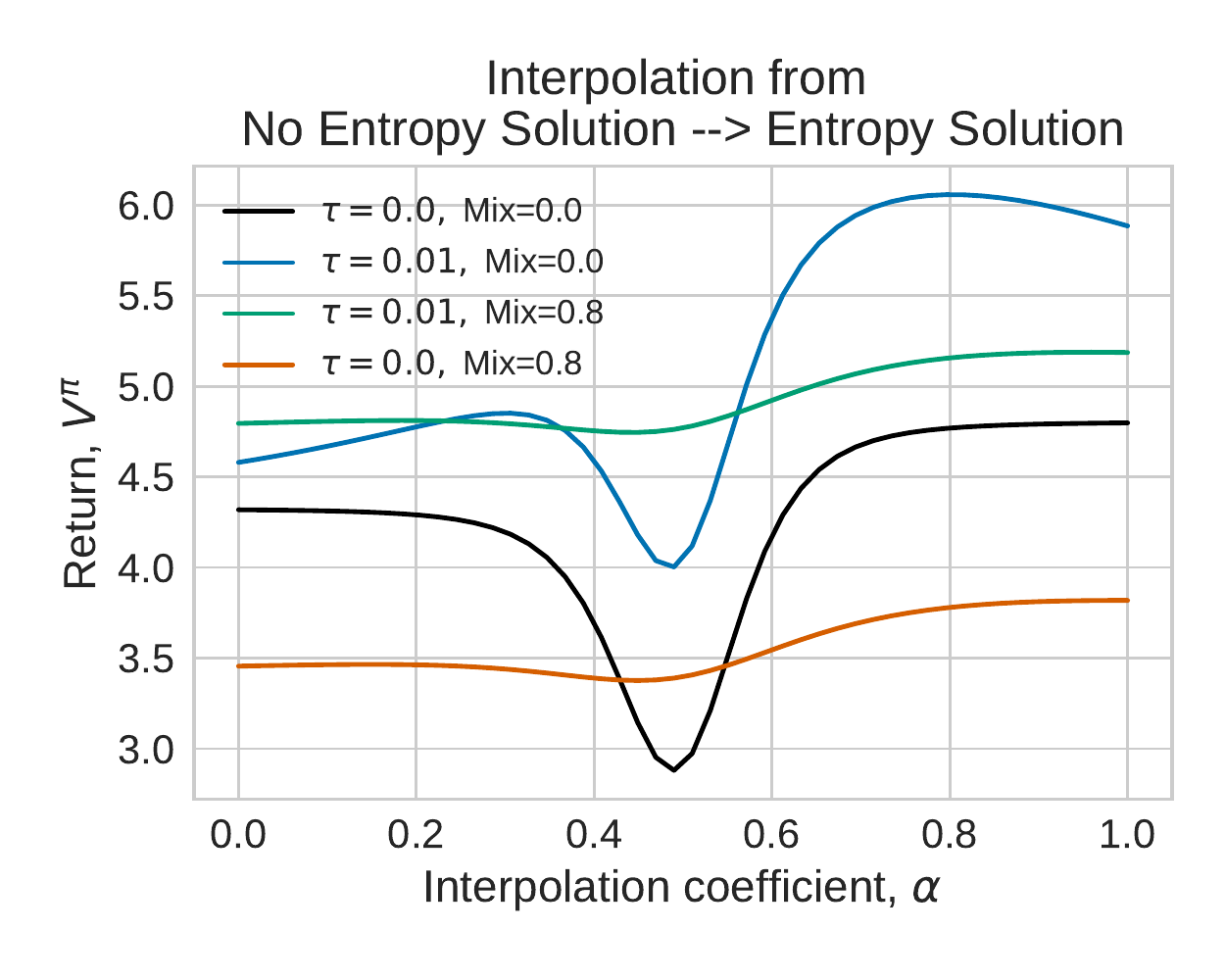}
        \caption{}
    \end{subfigure}
     \begin{subfigure}[b]{0.45\textwidth}
        \centering
        \includegraphics[width=\textwidth]{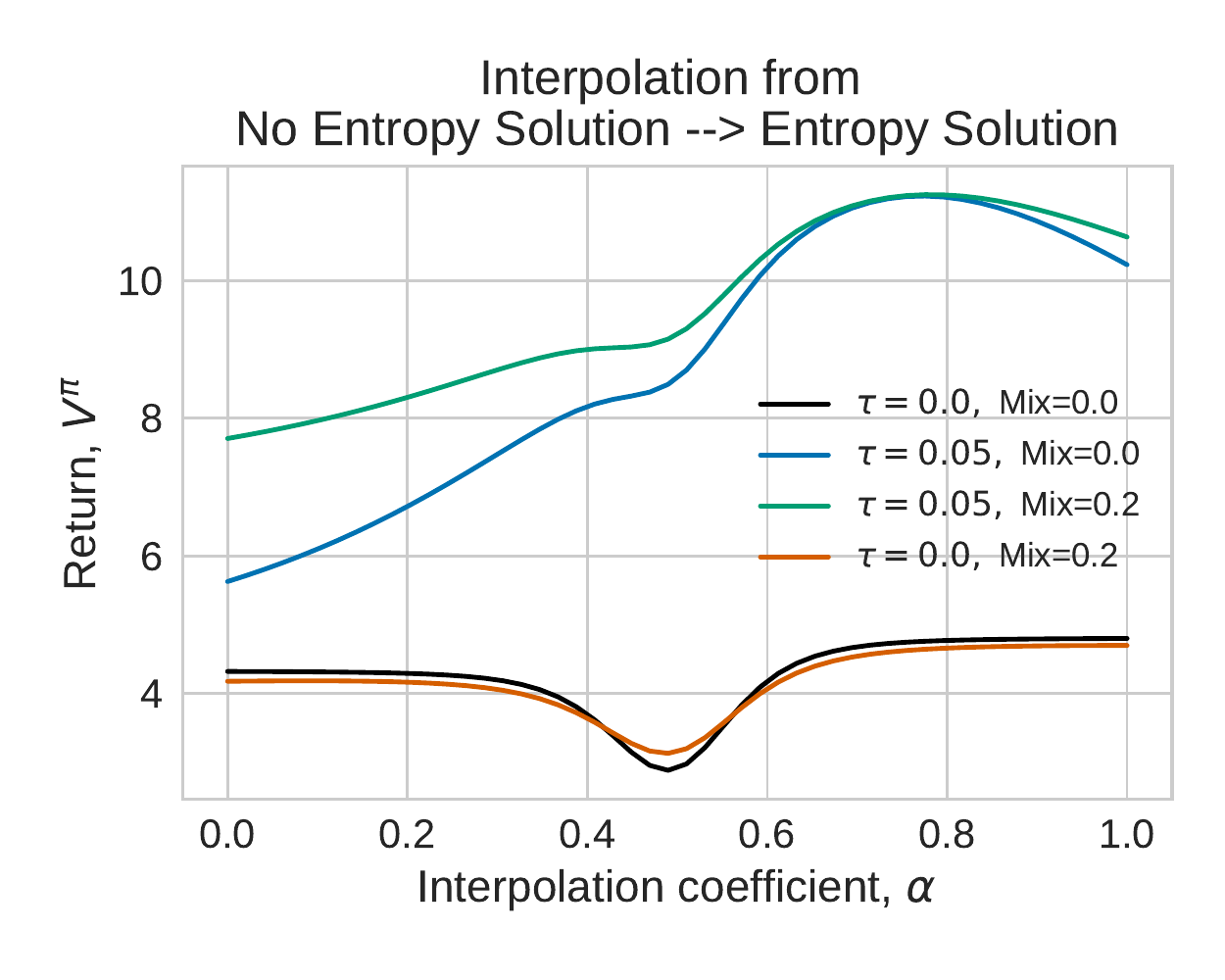}
        \caption{}
    \end{subfigure}
    \caption{Visualizing the objective function for different combinations of $\tau$ and minimum policy entropy (mix) in the interpolation between solutions found when optimizing with and without entropy entropy regularization.}
    \label{sfig:difficulty_exact}
\end{figure}

\begin{figure*}[b]
        \centering
        \includegraphics[angle=90,height=\textheight]{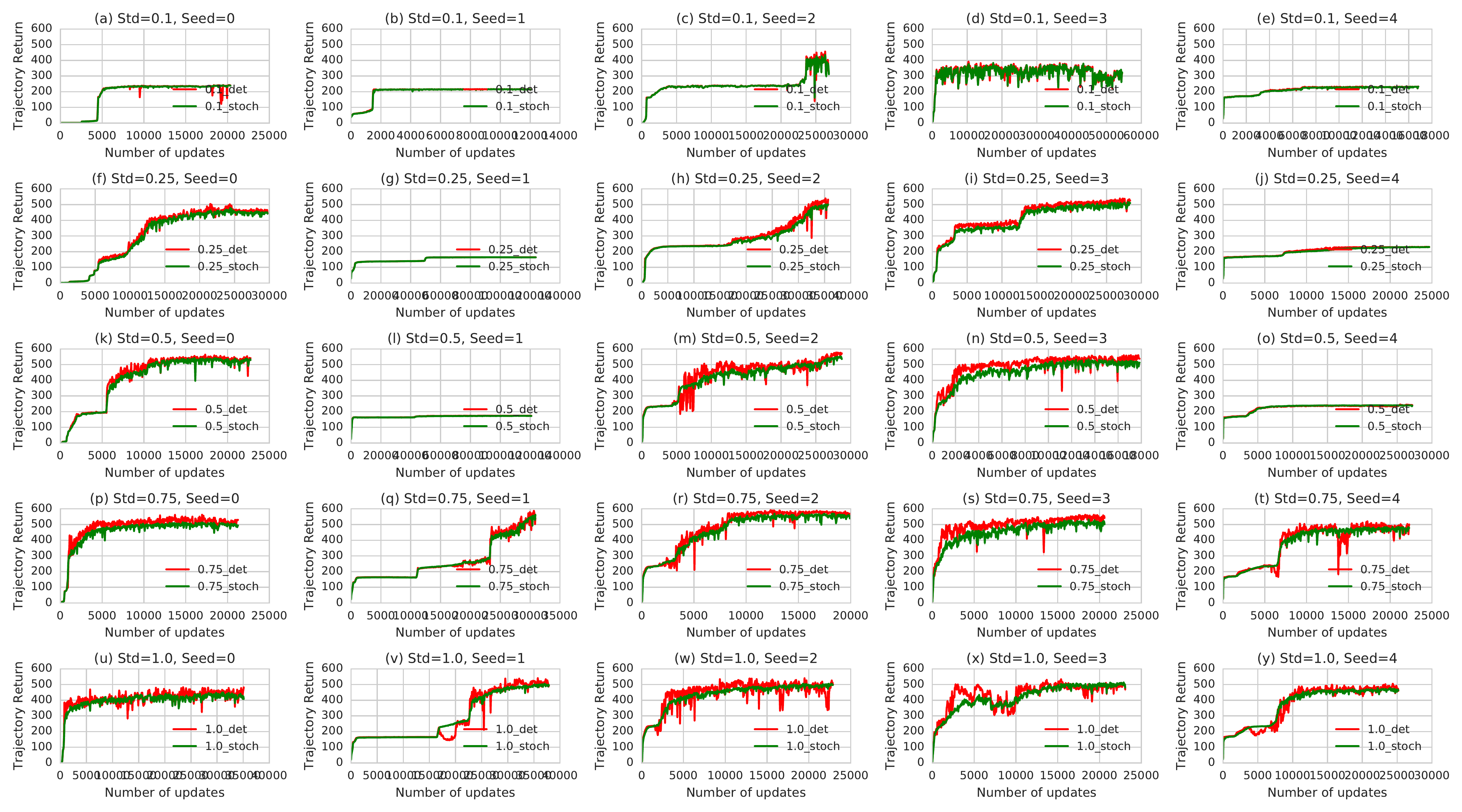}
        \caption{Individual learning curves for Hopper}
        \label{sfig:hopper_individual_learning_curves}
\end{figure*}
\begin{figure*}[b]
        \centering
        \includegraphics[angle=90,height=\textheight]{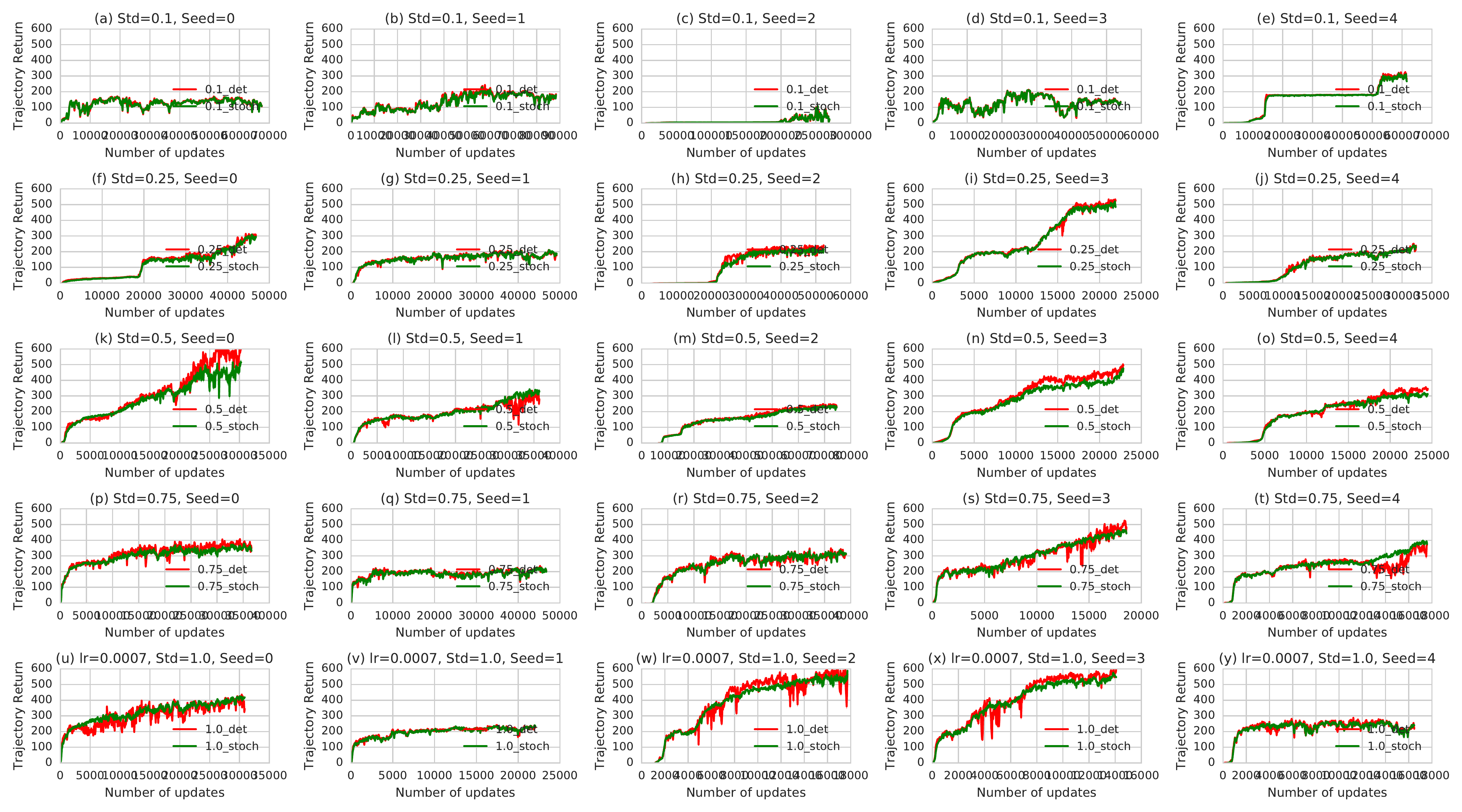}
        \caption{Individual learning curves for Walker}
        \label{sfig:walker_individual_learning_curves}
\end{figure*}

\begin{figure*}[b]
        \centering
        \includegraphics[angle=90,height=\textheight]{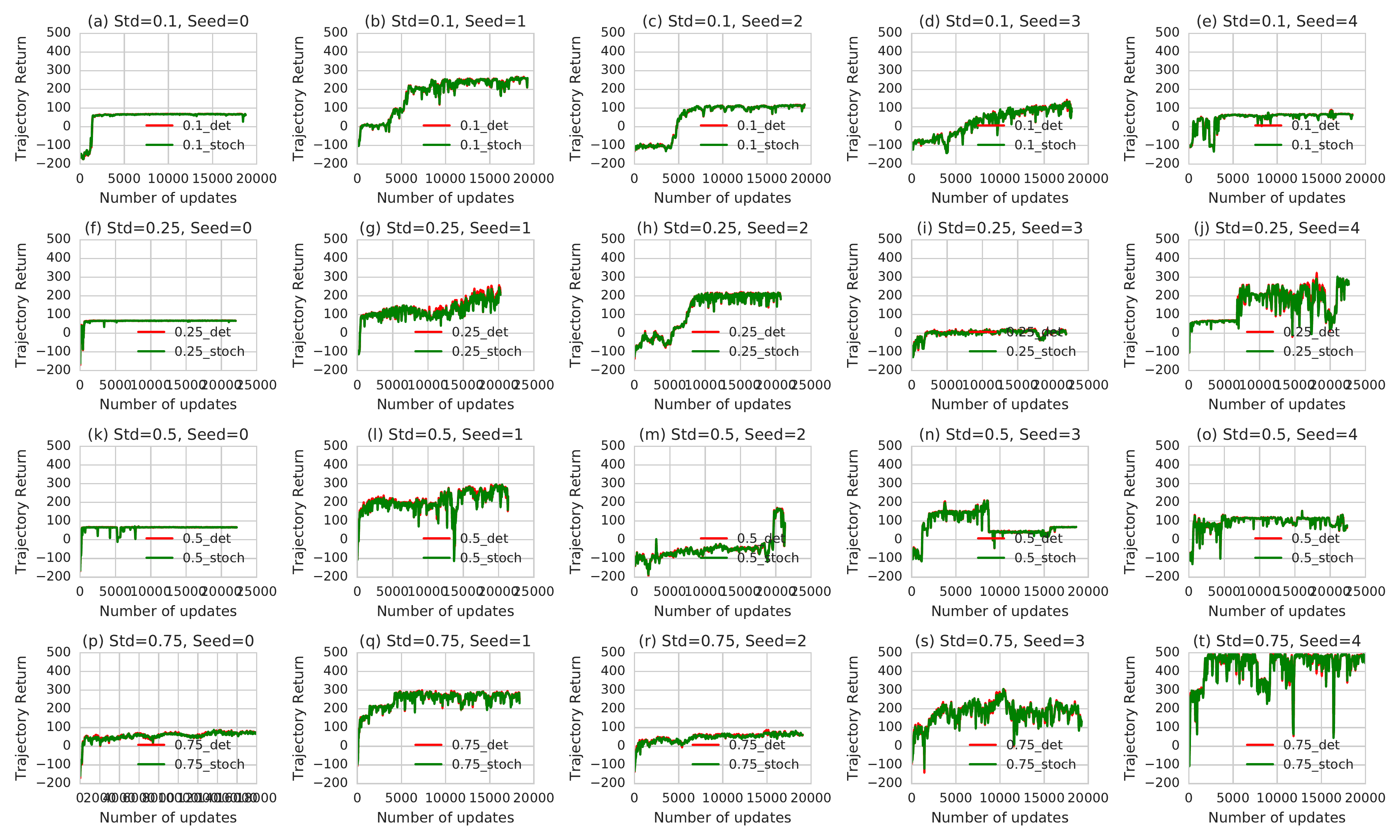}
        \caption{Individual learning curves for HalfCheetah}
        \label{sfig:halfcheetah_individual_learning_curves}
\end{figure*}

\begin{figure}
    \centering
     \begin{subfigure}[b]{0.22\textwidth}
        \centering
        \includegraphics[width=\textwidth]{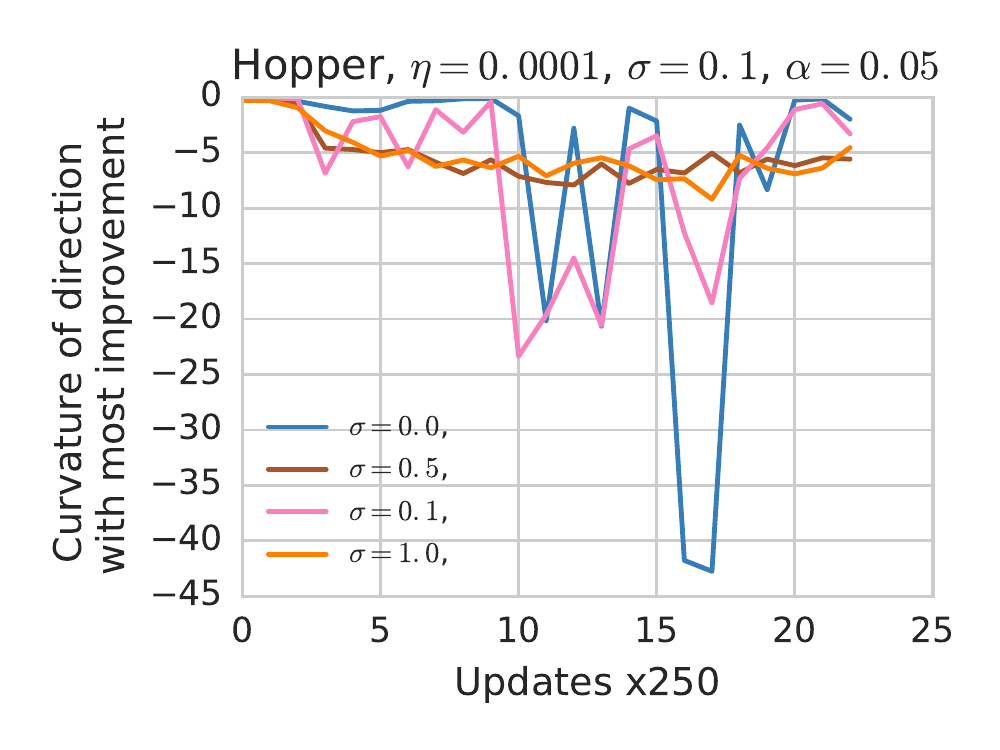}
        \caption{}
    \end{subfigure}
     \begin{subfigure}[b]{0.22\textwidth}
        \centering
        \includegraphics[width=\textwidth]{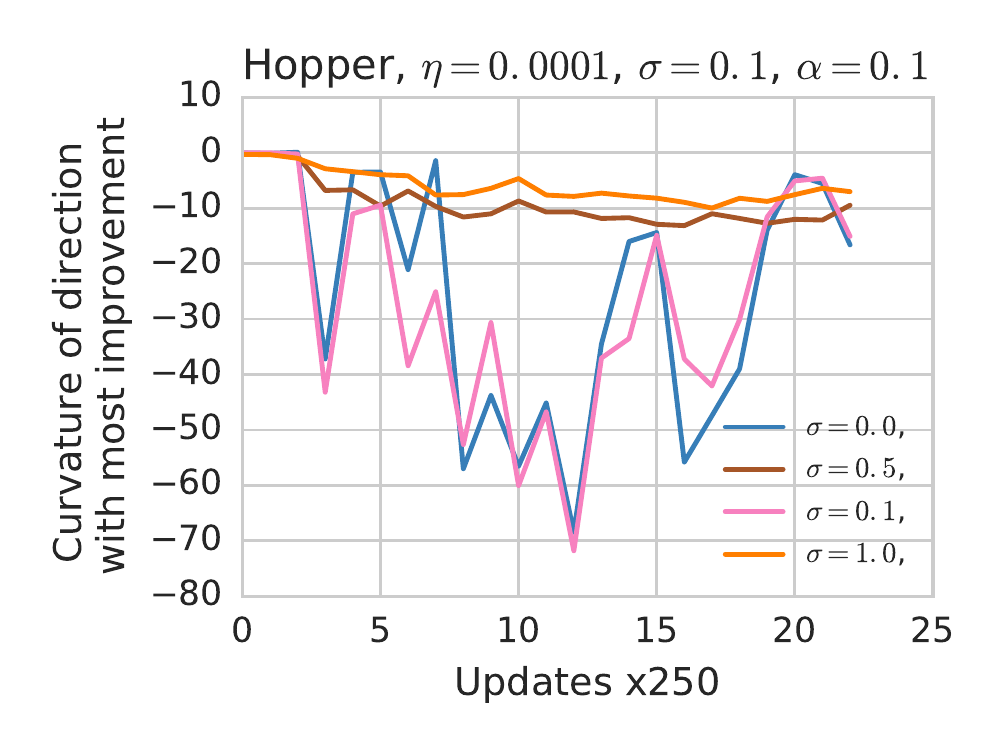}
        \caption{}
    \end{subfigure}
    \begin{subfigure}[b]{0.22\textwidth}
        \centering
        \includegraphics[width=\textwidth]{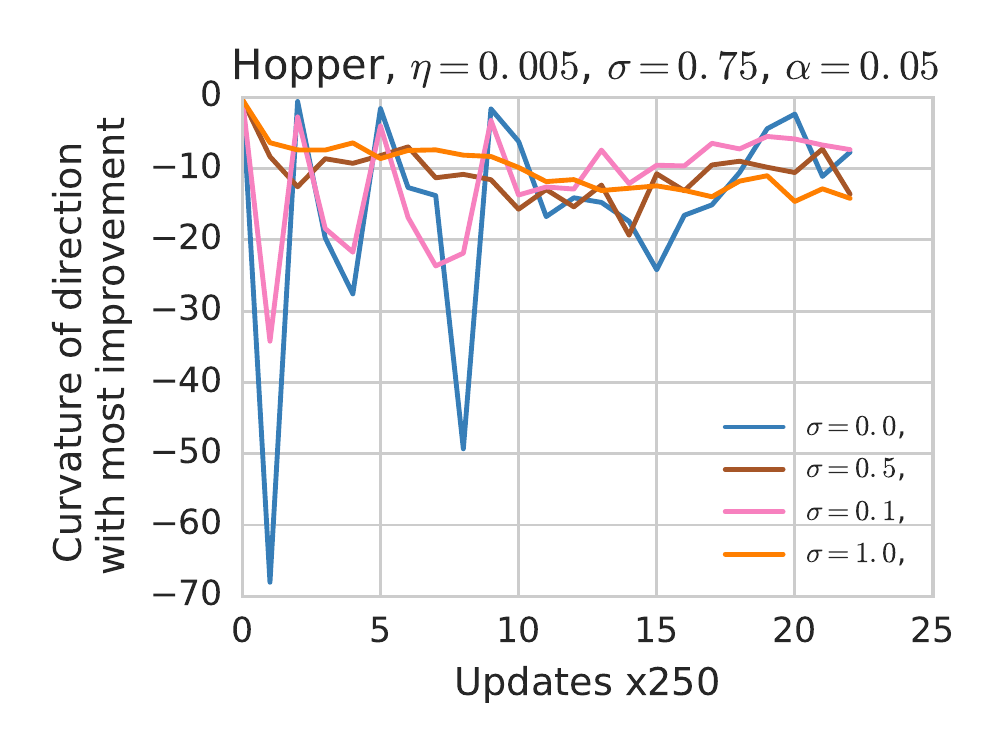}
        \caption{}
    \end{subfigure}
     \begin{subfigure}[b]{0.22\textwidth}
        \centering
        \includegraphics[width=\textwidth]{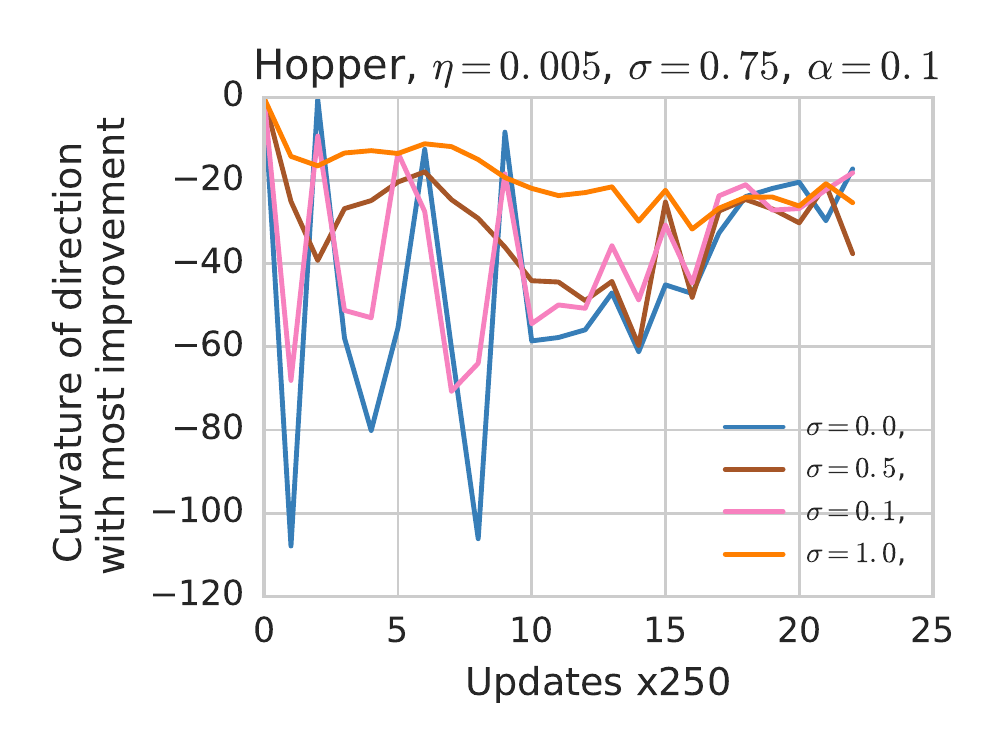}
        \caption{}
    \end{subfigure}
    \caption{Curvature for the direction with the most improvement during the optimization for different seeds and standard deviations in Hopper.}
    \label{sfig:curvature_fluctuation_hopper}
\end{figure}

\begin{figure}
    \centering
     \begin{subfigure}[b]{0.22\textwidth}
        \centering
        \includegraphics[width=\textwidth]{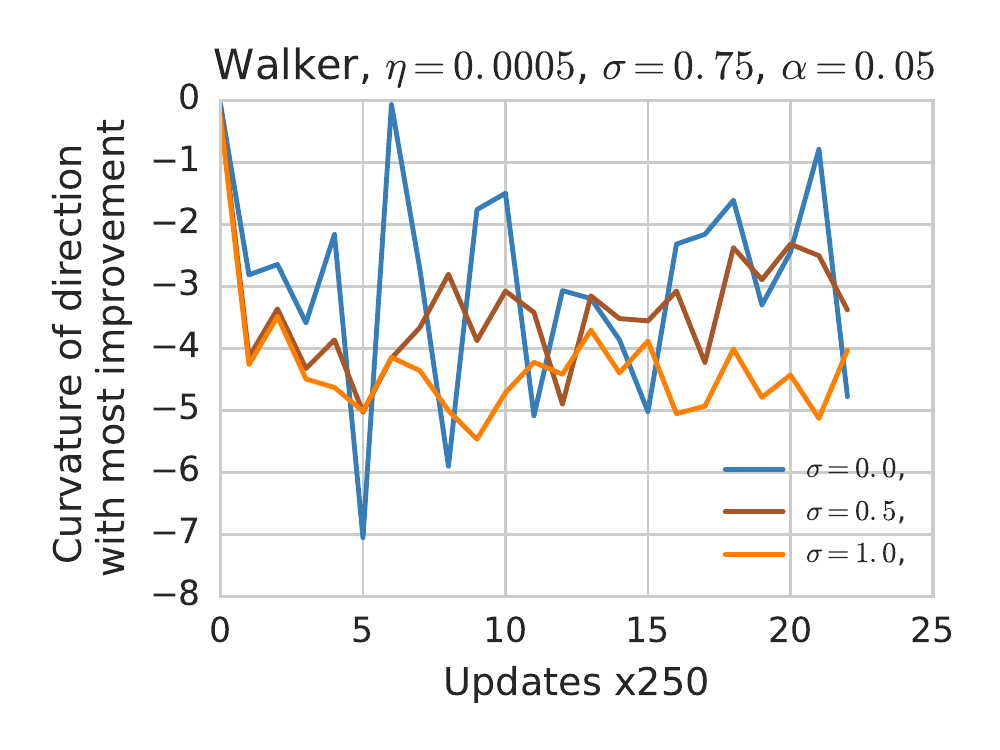}
        \caption{}
    \end{subfigure}
     \begin{subfigure}[b]{0.22\textwidth}
        \centering
        \includegraphics[width=\textwidth]{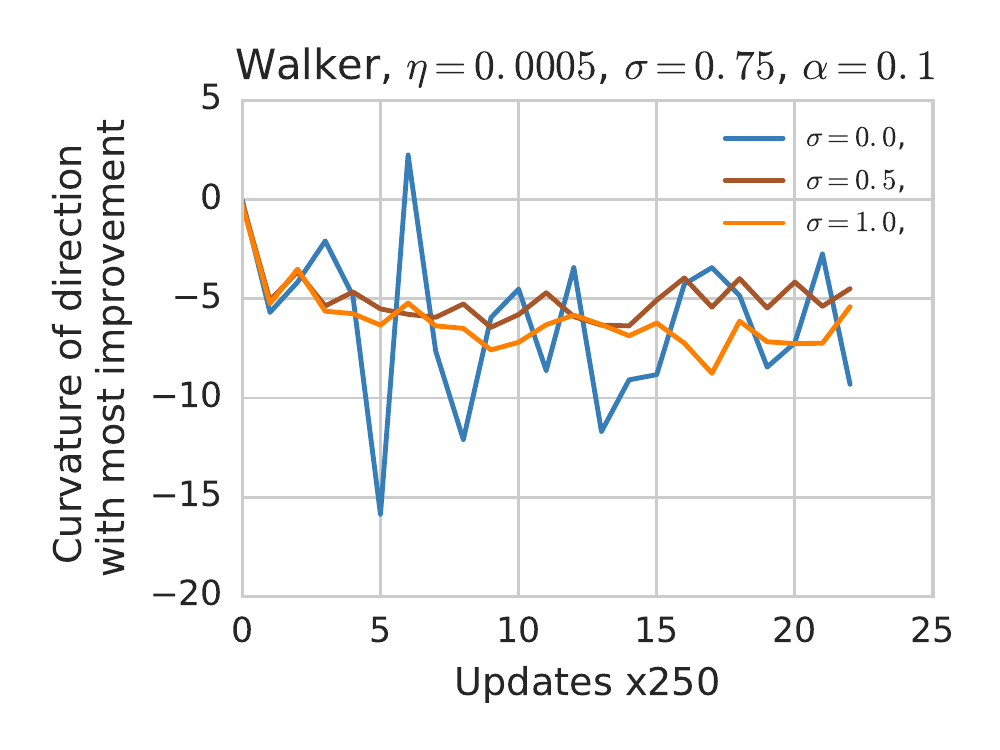}
        \caption{}
    \end{subfigure}
     \begin{subfigure}[b]{0.22\textwidth}
        \centering
        \includegraphics[width=\textwidth]{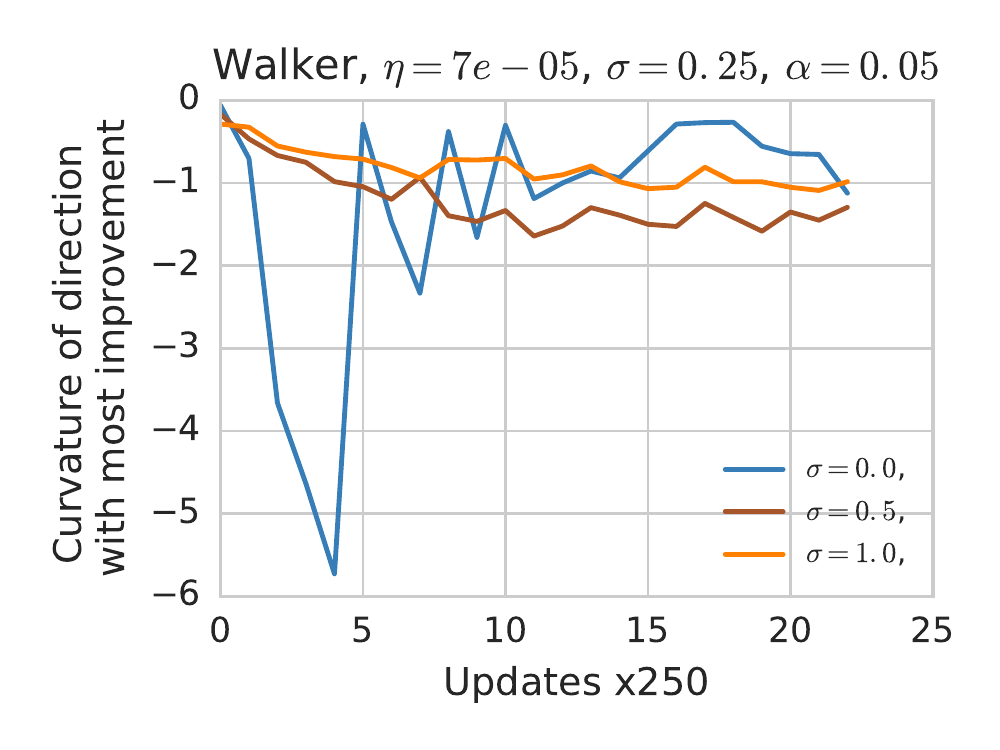}
        \caption{}
    \end{subfigure}
     \begin{subfigure}[b]{0.22\textwidth}
        \centering
        \includegraphics[width=\textwidth]{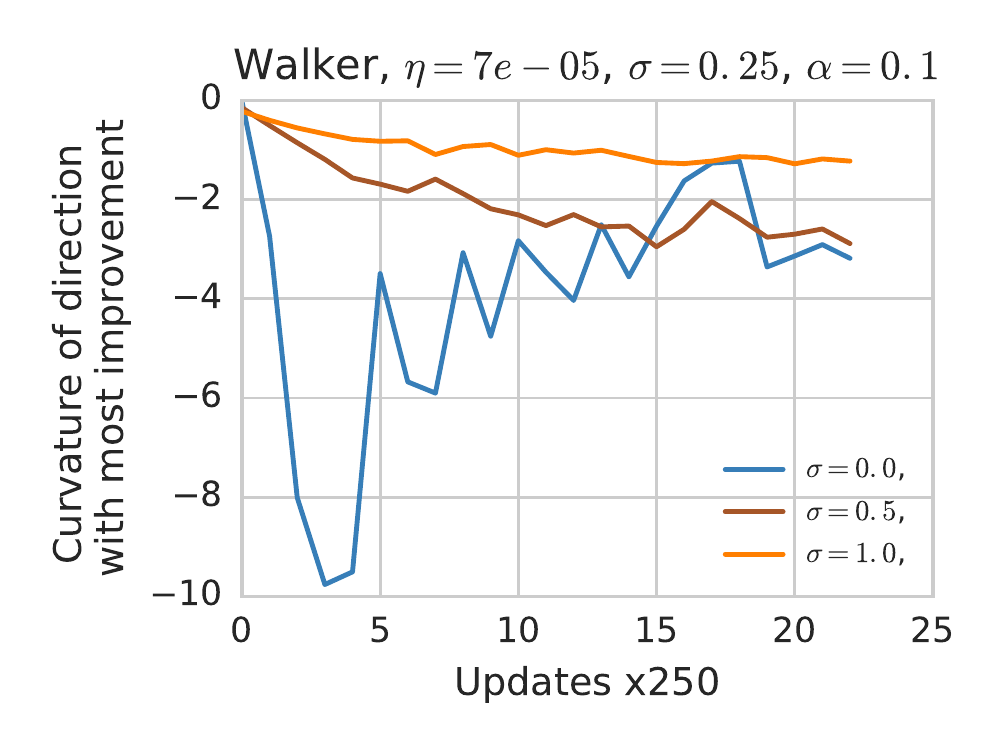}
        \caption{}
    \end{subfigure}
    \caption{Curvature for the direction with the most improvement during the optimization for different seeds and standard deviations in Walker2d.}
    \label{sfig:curvature_fluctuation_walker}
\end{figure}

\begin{figure}
    \centering
    \hspace*{-0.1\textwidth}
     \begin{subfigure}[b]{0.22\textwidth}
        \centering
        \includegraphics[width=\textwidth]{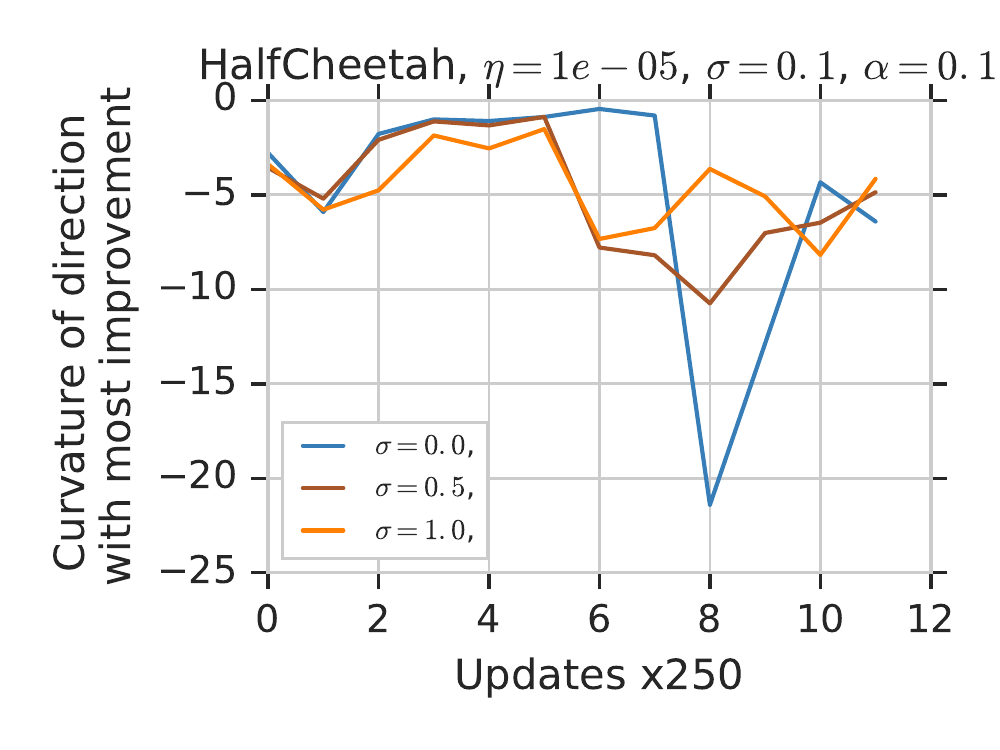}
        \caption{}
    \end{subfigure}
     \begin{subfigure}[b]{0.22\textwidth}
        \centering
        \includegraphics[width=\textwidth]{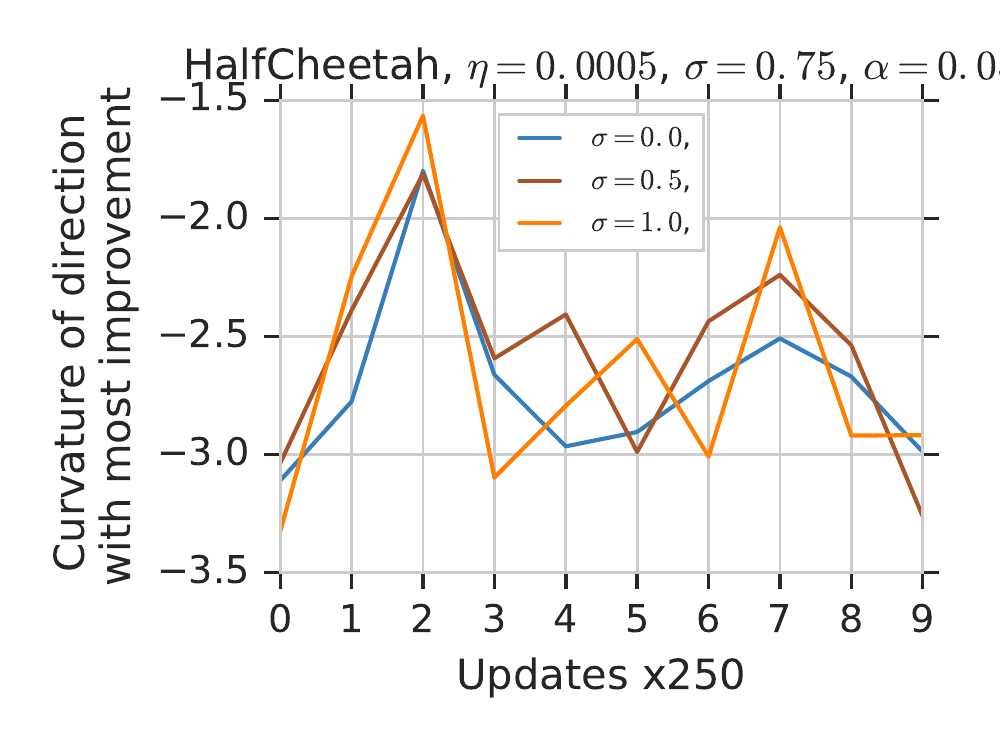}
        \caption{}
    \end{subfigure}
     \begin{subfigure}[b]{0.22\textwidth}
        \centering
        \includegraphics[width=\textwidth]{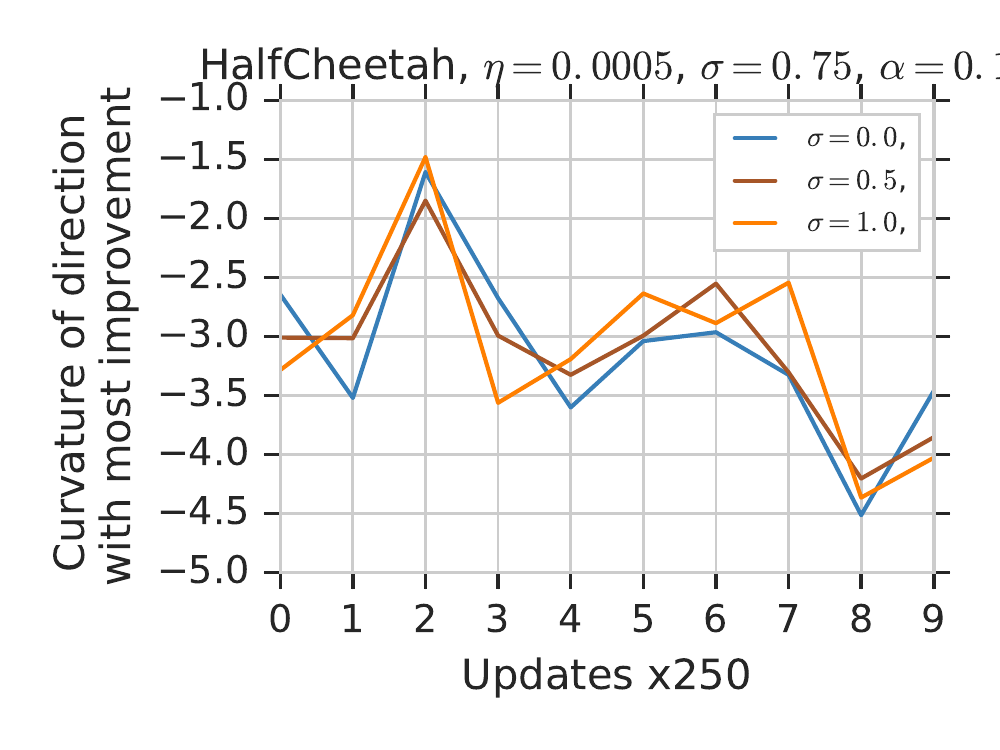}
        \caption{}
    \end{subfigure}
    \caption{Curvature for the direction with the most improvement during the optimization for different seeds and standard deviations in HalfCheetah.}
    \label{sfig:curvature_fluctuation_halfcheetah}
\end{figure}

\begin{figure}[b]
        \centering
        \includegraphics[width=0.5\textwidth]{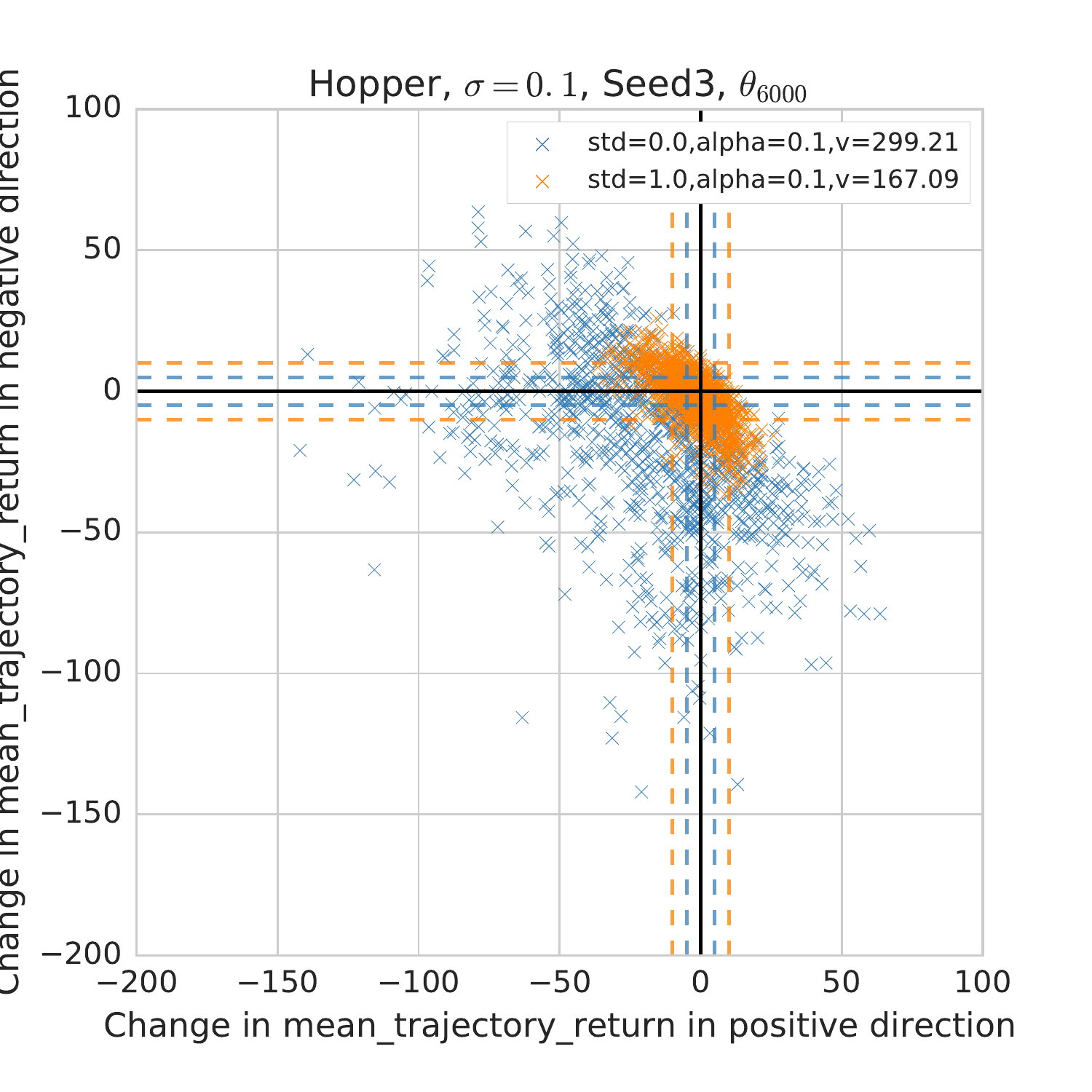}
        \caption{Scatter plot for randomly drawn directions for the solution shown in Figure~\ref{fig:hopper_optima_spectra}. For $\sigma=0$, most directions are negative and they all have near zero or negative curvature. For $\sigma=1$, there are fewer negative directions, and more importantly less negative curvature: Indicating an almost linear region. This linear region can be seen in the 1D interpolation (Figure~\ref{fig:hopper_optima_spectra}c)}
        \label{sfig:hopper_scatter}
\end{figure}

\begin{figure}
    \centering
    \hspace*{-0.05\textwidth}
    \begin{subfigure}[b]{0.33\textwidth}
        \centering
        \includegraphics[width=\textwidth]{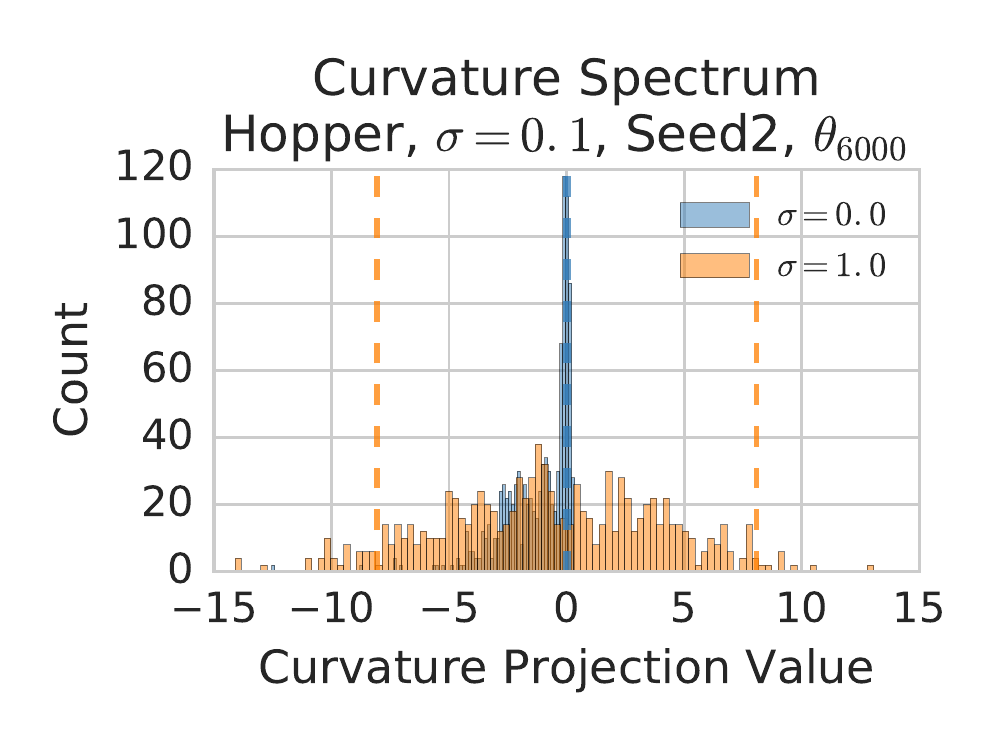}
        \caption{
        }
    \end{subfigure}
    \begin{subfigure}[b]{0.33\textwidth}
        \centering
        \includegraphics[width=\textwidth]{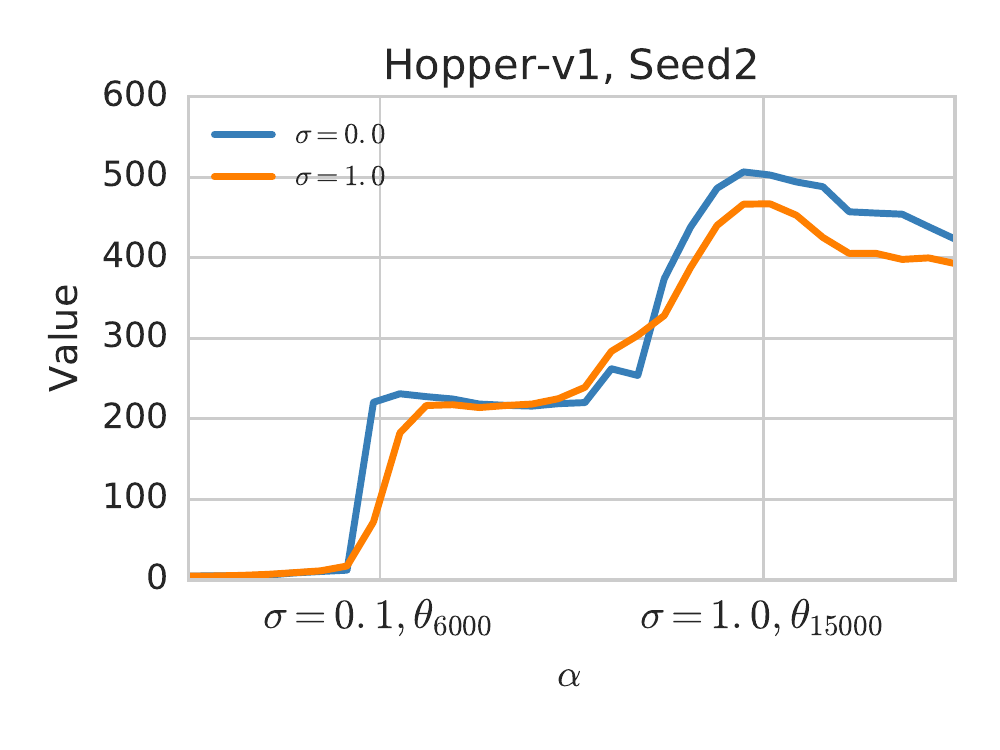}
        \caption{}
    \end{subfigure}
    \caption{Curvature spectra and linear interpolation for a solution in Hopper. 
    (a) For $\sigma=0$ most directions have a significant negative curvature implying that this solution is near a local optimum. For $\sigma=1$ all curvature values are indistinguishable from noise. (b) An increasing path to a better solution exists but might be non-trivial for a line search to follow.}
    \label{sfig:hopper_optima_spectra}
\end{figure}
\begin{figure}[ht]
    \centering
    \hspace*{-0.05\textwidth}
    \begin{subfigure}[b]{0.33\textwidth}
        \centering
        \includegraphics[width=\textwidth]{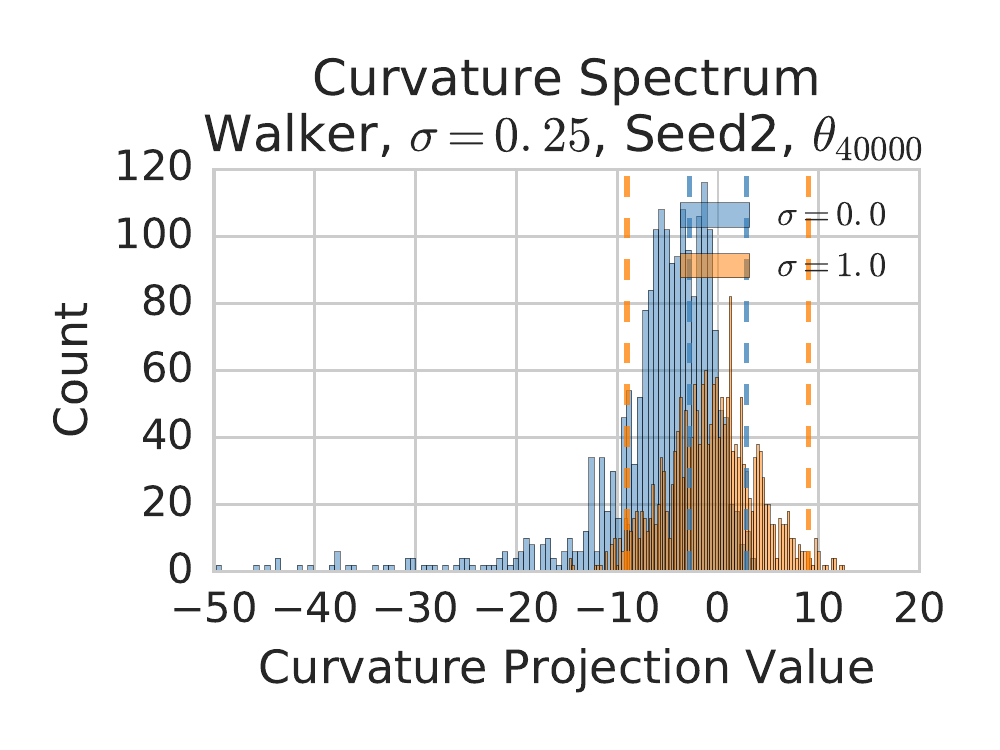}
        \caption{
        }
    \end{subfigure}
    \begin{subfigure}[b]{0.33\textwidth}
        \centering
        \includegraphics[width=\textwidth]{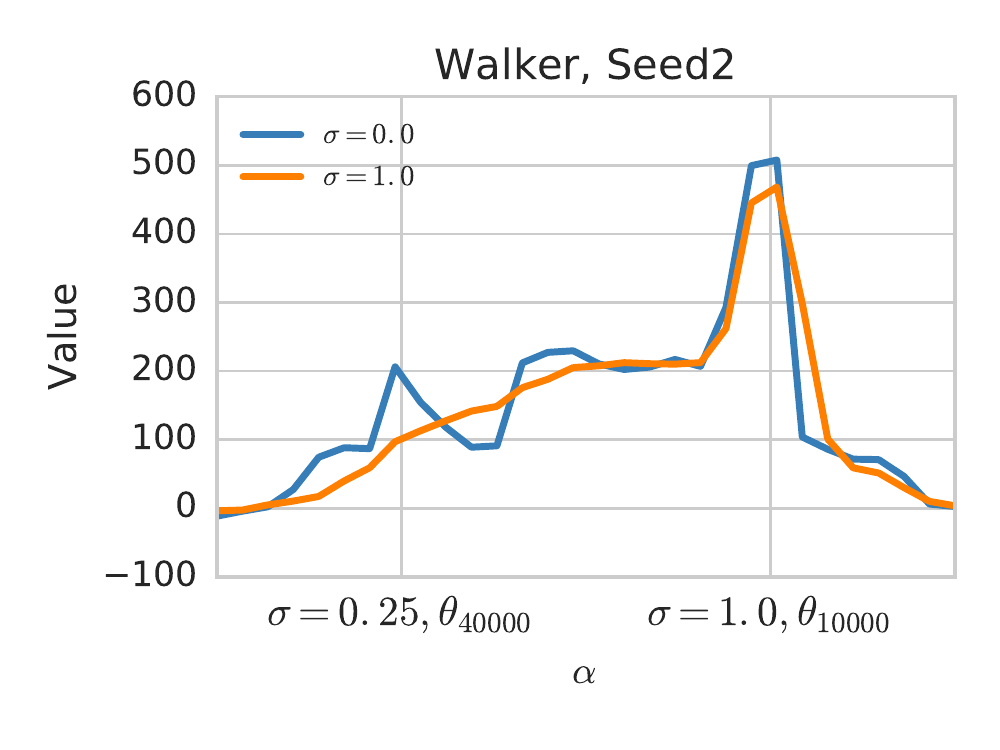}
        \caption{}
    \end{subfigure}
    \caption{Curvature spectra and linear interpolation for solutions in Walker2d. 
    (a) For $\sigma=0$ most directions have a significant negative curvature implying that this solution is near a local optimum. For $\sigma=1$ all curvature values are indistinguishable from noise. (c) A monotonically increasing path to a better solution exists if we knew the direction to a solution a-priori.}
    \label{sfig:walker_optima_spectra}
\end{figure}
\begin{figure}[ht]
    \centering
    \hspace*{-0.05\textwidth}
    \hspace*{-0.05\textwidth}
    \begin{subfigure}[b]{0.33\textwidth}
        \centering
        \includegraphics[width=\textwidth]{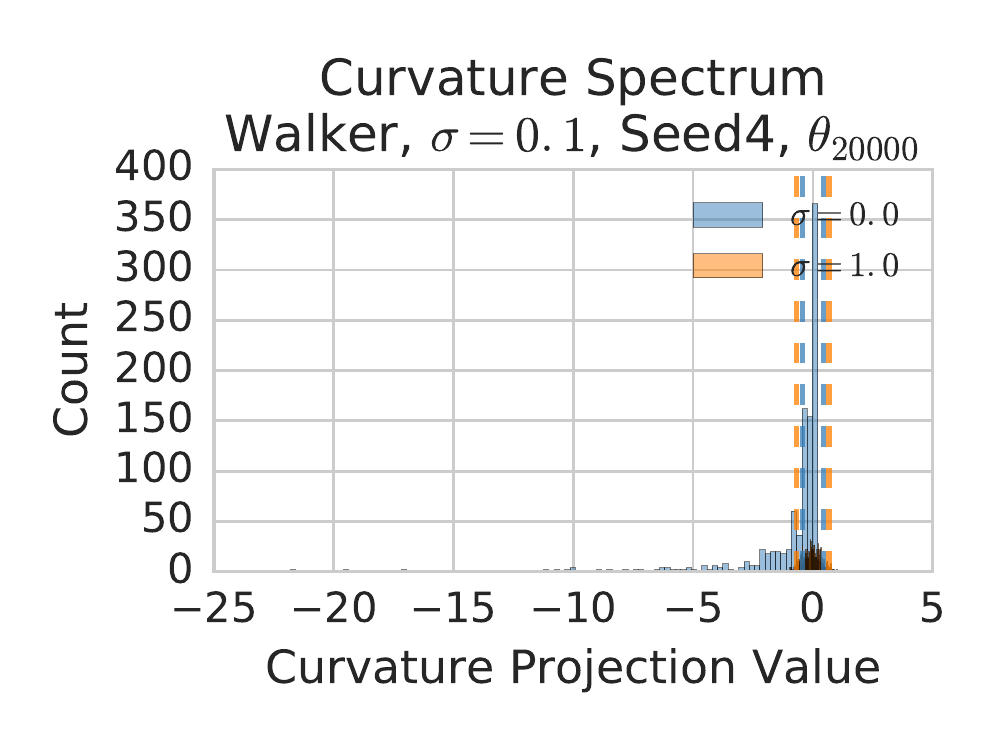}
        \caption{}
    \end{subfigure}
    \begin{subfigure}[b]{0.33\textwidth}
        \centering
        \includegraphics[width=\textwidth]{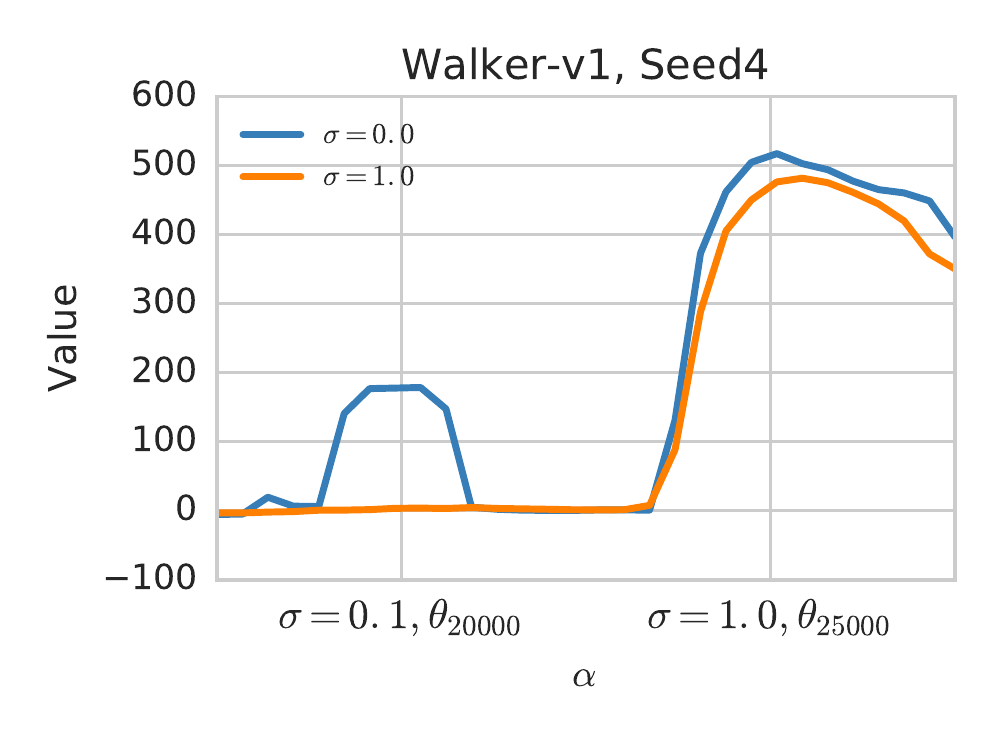}
        \caption{}
    \end{subfigure}
    \caption{Curvature spectra and linear interpolation for solutions in Walker2d. 
    (a) For $\sigma=0$ the local objective has mostly negative curvature, but when increased to $\sigma=1$ there is no curvature. Putting these two results together means that the solution ends up being in a flat region. (b) A linear interpolation confirms this observation in 
    one dimension.}
    \label{sfig:walker_optima_spectra2}
\end{figure}

\begin{figure}
    \centering
    \hspace*{-0.05\textwidth}
    \begin{subfigure}[b]{0.33\textwidth}
        \centering
        \includegraphics[width=\textwidth]{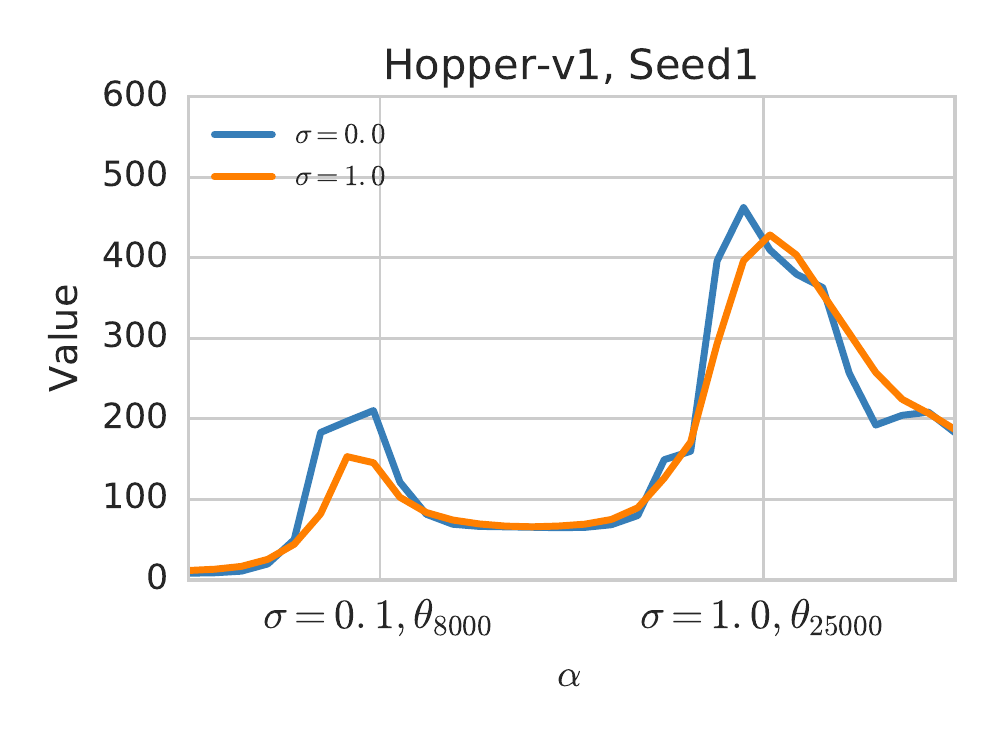}
        \caption{}
    \end{subfigure}
    \begin{subfigure}[b]{0.33\textwidth}
        \centering
        \includegraphics[width=\textwidth]{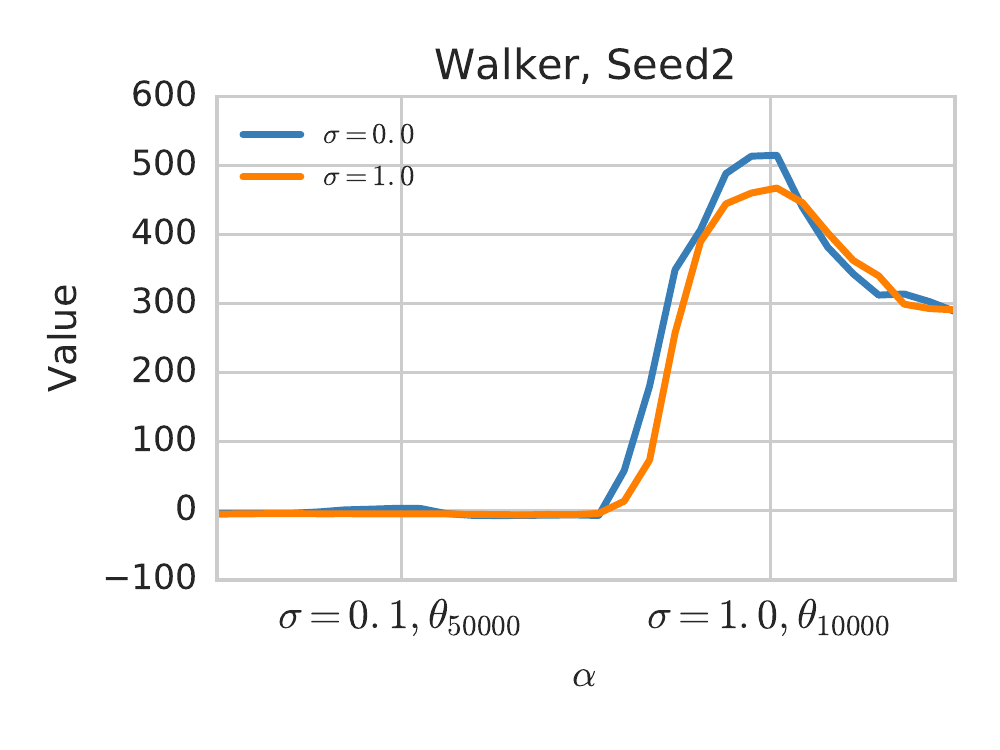}
        \caption{}
    \end{subfigure}
    \begin{subfigure}[b]{0.33\textwidth}
        \centering
        \includegraphics[width=\textwidth]{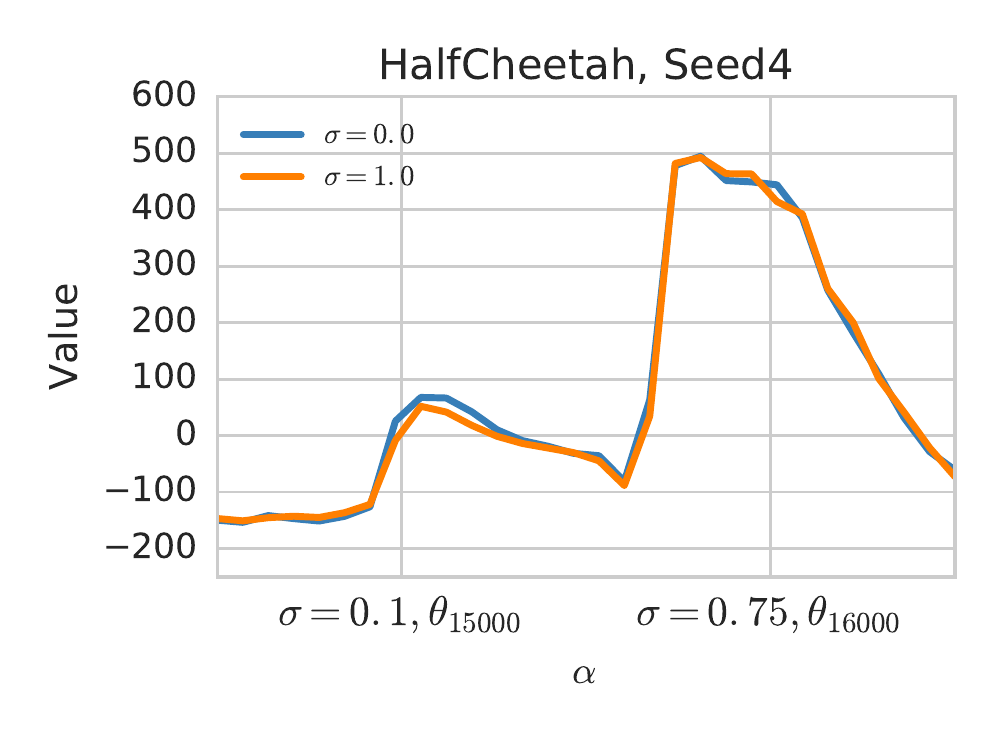}
        \caption{}
    \end{subfigure}
    \caption{Negative examples for linear interpolations between solutions. Interpolations between these solutions do not show a monotonically increasing path under the high entropy objective suggesting that though high entropy objectives might connect some local optima, they do not connect all.}
    \label{sfig:optima_negative}
\end{figure}

\begin{figure*}
    \centering
    \hspace*{-0.05\textwidth}
    \begin{subfigure}[b]{0.22\textwidth}
        \centering
        \includegraphics[width=\textwidth]{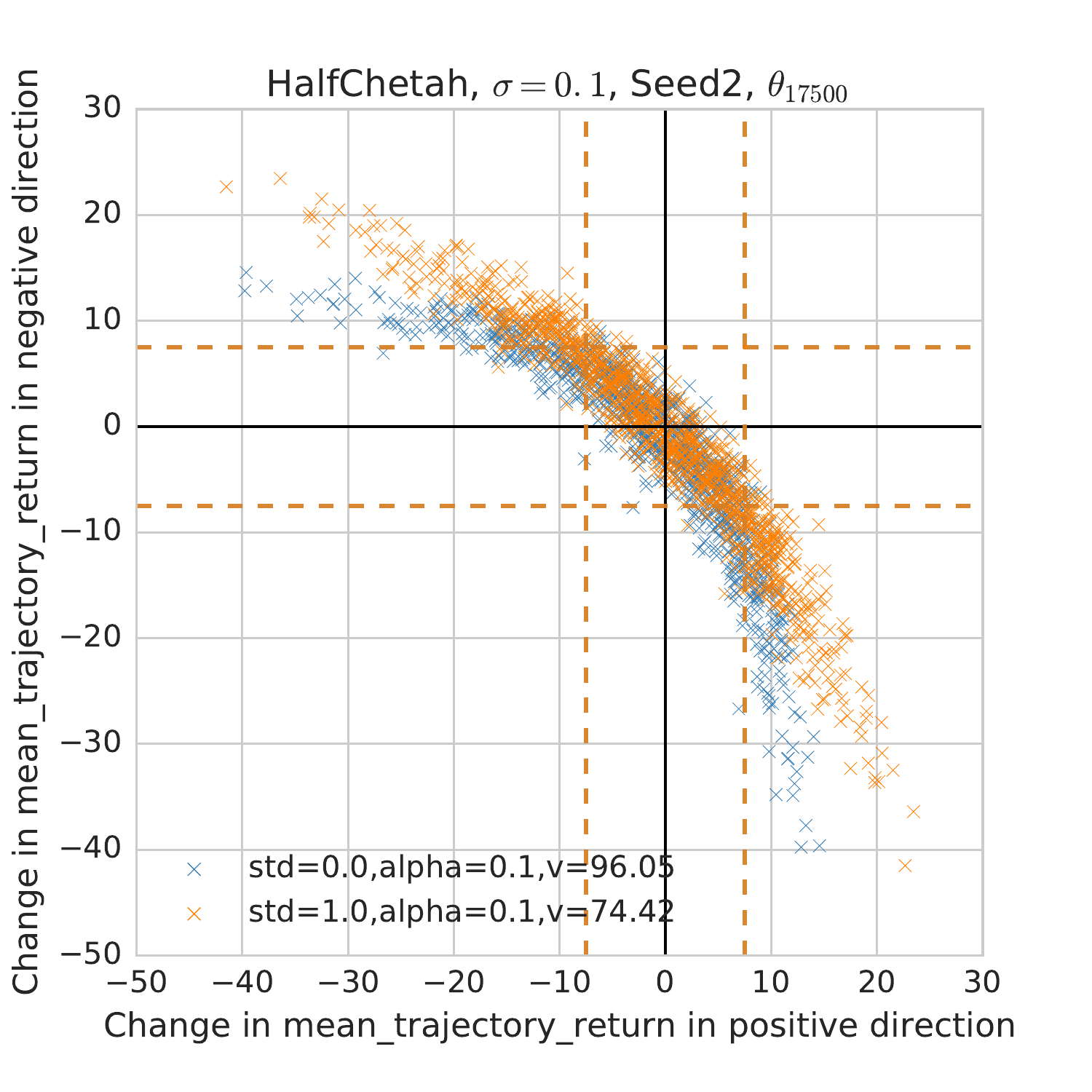}
        \caption{}
    \end{subfigure}
    \begin{subfigure}[b]{0.22\textwidth}
        \centering
        \includegraphics[width=\textwidth]{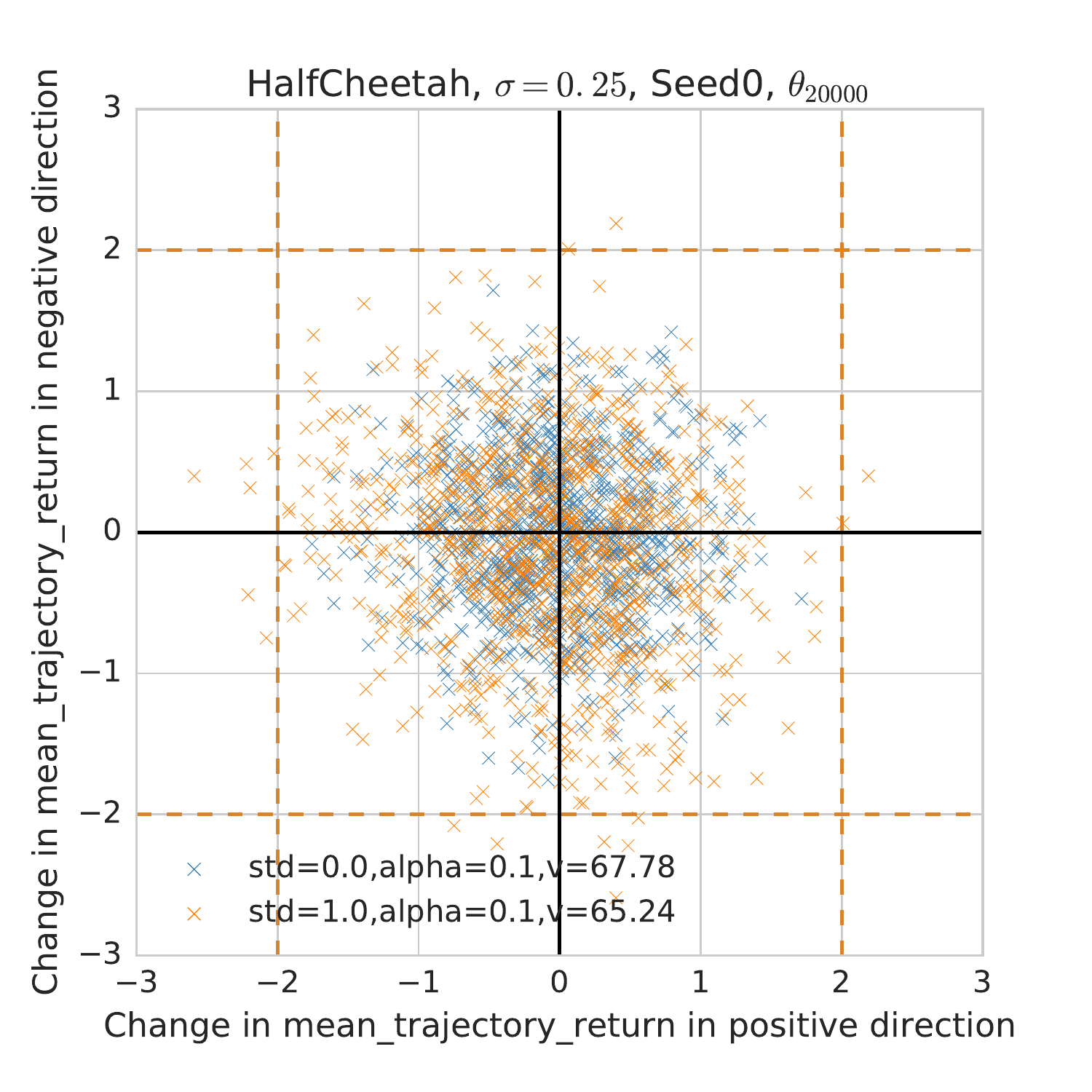}
        \caption{}
    \end{subfigure}
    \begin{subfigure}[b]{0.22\textwidth}
        \centering
        \includegraphics[width=\textwidth]{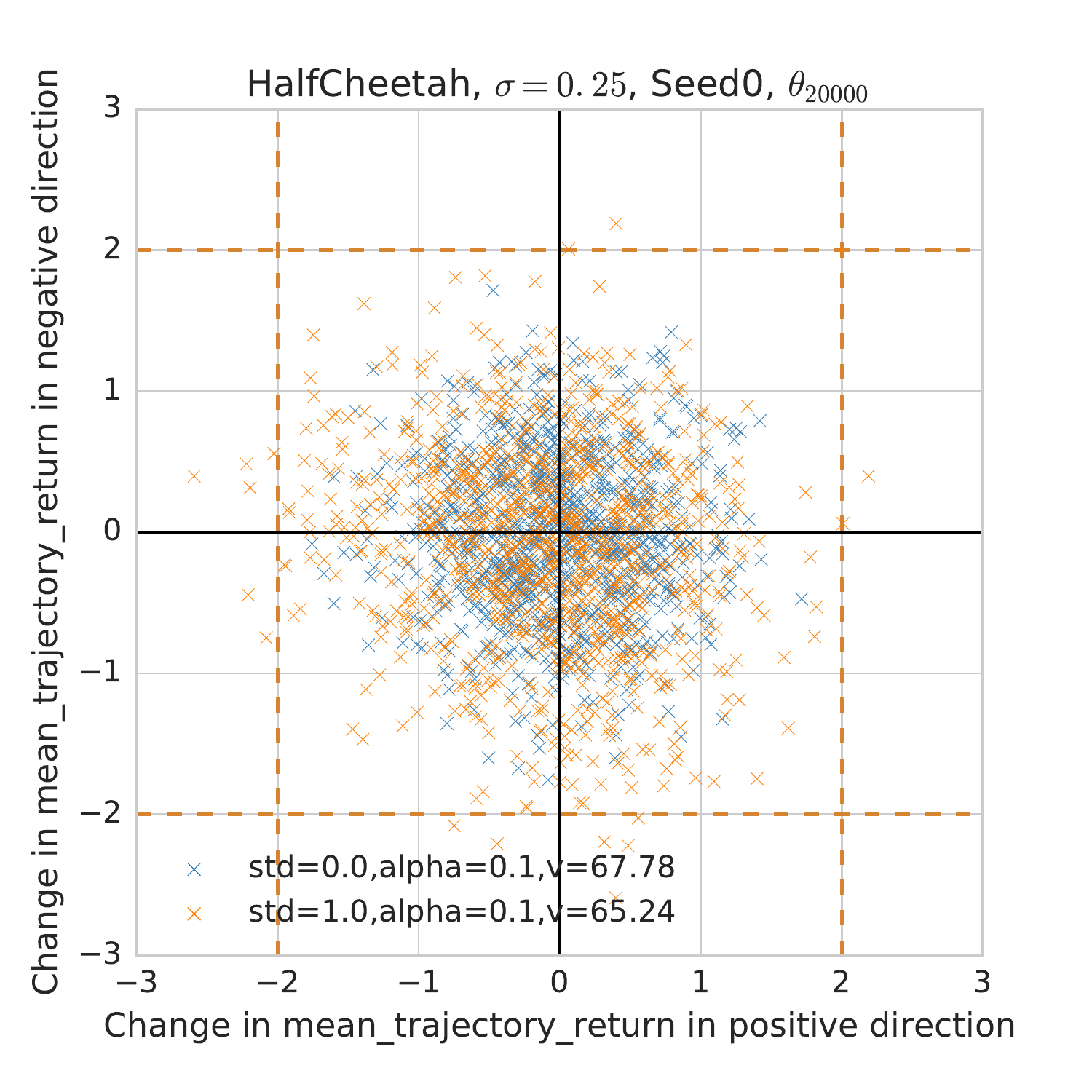}
        \caption{}
    \end{subfigure}
    \begin{subfigure}[b]{0.22\textwidth}
        \centering
        \includegraphics[width=\textwidth]{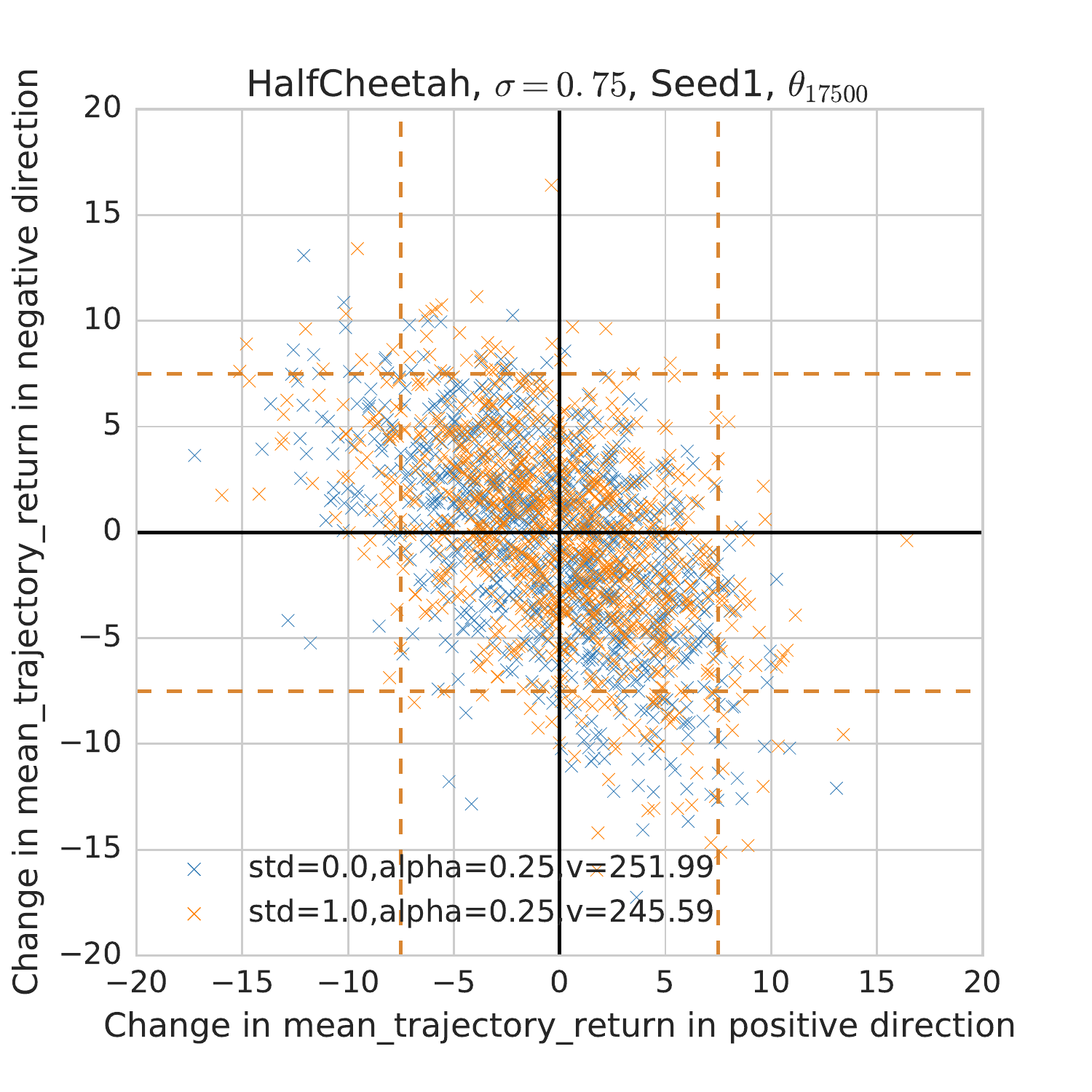}
        \caption{}
    \end{subfigure}
    \caption{Local objective functions do not change much in HalfCheetah when $\sigma$ is increased.}
    \label{sfig:halfcheetah_optima_spectra}
\end{figure*}

\subsection{More details about visualizing objective functions using random perturbations}
\label{sec:appendix_vis}
We introduced a novel technique for visualizing objective functions by using random perturbations. Understanding the benefits and limitations is key to knowing when this method will be useful.


\subsubsection{Benefits of random perturbations}
\label{asec:objs_vis_benefits}
\begin{enumerate}
    \item Since our technique is only bounded by the number of samples we wish to obtain, it allows us to scale beyond regimes where computing eigenvalues of $H$ might be computationally expensive. In particular, our method does not require computing any gradients and is amenable to massive parallelization.
    \item Our random perturbations capture a lot of information about the local geometry of an objective function. In this work we discuss two possible summarizations that capture information about the gradient and Hessian. Other summarizations may exist that capture different geometrical and topological properties of the objective function around this point. 
\end{enumerate}

\subsubsection{Derivation for Equation~\ref{eqn:projected_final_answer}}
\label{asec:obj_vis}
Here we derive the form for projections onto the two diagonal axes $x=y$ and $x=-y$. Assume $\Oh(\theta)\approx a^T\theta + \frac{1}{2}\theta^TH\theta$. Now
\begin{align}
        \Oh(\theta_0 +\alpha d) = {} & a^T (\theta_0 + \alpha d) + \frac{1}{2}(\theta_0+\alpha d)^T H(\theta_0+\alpha d)\\
        = {} & a^T\theta_0 +\alpha a^T d + \frac{\alpha}{2}[\theta_0^T H d + d^T H\theta_0] + \frac{\alpha^2}{2}d^T Hd + \frac{1}{2}\alpha \theta_0^TH\theta_0 \nonumber \\
        = {} & \Oh(\theta_0) + \alpha a^T d + \frac{\alpha^2}{2}d^THd +  \theta_0^TH d
\end{align}
Therefore:
\begin{align}
    \Delta^{\Oh+} &= \Oh(\theta_0+\alpha d) - \Oh(\theta_0) \\
    &= \alpha a^T d + \frac{\alpha^2}{2}d^THd + \alpha \theta_0^THd
\end{align}
and similarly, 
\begin{align}
    \Delta^{\Oh-} &= \Oh(\theta_0-\alpha d) - \Oh(\theta_0) \\
    &= -\alpha a^T d + \frac{\alpha^2}{2}d^THd - \alpha\theta_0^THd
\end{align}
Now doing the projection onto the diagonal axes we get: 
\begin{equation}
    \Delta^{\Oh+} + \Delta^{\Oh-} = \alpha^2 d^T H d
\end{equation}
which gives us information about the Hessian in direction $d$ and 
\begin{align}
    \Delta^{\Oh+} - \Delta^{\Oh-} &= 2\alpha a ^T d + 2\alpha \theta_0^T H d \\
    &= 2\alpha (a + \theta_0 H)^Td\\
    &= 2\alpha\nabla \Oh(\theta_0)^T d
\end{align}
which gives us information about the gradient in that direction.

By repeating this procedure and obtaining many samples, and can thus get an understanding of how $\Oh$ changes in many directions around $\theta_0$.

\subsubsection{Limitations}
Consider $\Oh(\theta)=-\sum_{i=1}^{k_1} \theta_i^2 + -\sum_{i=k_1+1}^{k_1 + k_2} (\theta_i-2)^2$ where at $\theta=\vec{0}$ the function is locally optimal in $k_1$ directions but there are $k_2$ ascent directions that improve $\Oh$. To get an idea of the extent of this limitation, the loss function is perturbed based on directions given by a stochastic gradient\footnote{Stochastic gradients are simulated by adding Gaussian noise with co-variance $\epsilon 2I_{k_1+k_2}$ to the true gradient \citep{mandt2017stochastic}} in contrast to random directions. 
When the total number of dimensions, $k_1+k_2$, is small, random perturbations can accurately capture all directions regardless of the relative magnitudes of $k_1$ and $k_2$ (Figure~\ref{sfig:random_perturbations_limits}ab). When the number of dimensions, $k_1+k_2$, is large and the number of improvement directions, $k_1=k_2$, both methods discover the ascent directions (Figure~\ref{sfig:random_perturbations_limits}c). However, when $k_1 \gg k_2$, both random perturbations and stochastic gradients miss the ascent directions unless noise is small (Figure~\ref{sfig:random_perturbations_limits}d). 

\subsection{Derivation of Entropy-augmented exact policy gradient (Equation~\ref{eqn:policy_gradient_thm_entropy})}
\label{asec:derivation_exact_pg}
In this section we derive the exact gradient updates used in Section~\ref{sec:high_variance_issue} for the entropy regularized objective. This derivation differs from but has the same solution as \cite{sutton2000policy} when $\tau=0$. Recall that the objective function is given by:
\begin{equation}
    V^\pi(s_0) = \sum_a \pi(a|s=s_0) Q^\pi(s_0,a)
\end{equation}
where $V^\pi(s_0)$ is the expected discounted sum of rewards from the starting state. We can substitute the definition of $Q^\pi(s,a) = \big[r(s,a) + \tau \mathbb{H}(\pi(\cdot|s)) + \gamma\sum_{s'} P(s'|s=s_0,a) V^\pi(s')\big]$ to obtain a recursive formulation of the objective.
{\small\begin{align}
    V^\pi(s_0) = {} & \sum_a \pi(a|s=s_0) \big[r(s_0,a) + \tau \mathbb{H}(\pi(\cdot|s_0)) \nonumber\\
    & + \gamma\sum_{s'} P(s'|s=s_0,a) V^\pi(s')\big]
\end{align}}
If our policy $\pi$ is parameterized by $\theta$ we can take the gradient of this objective function so that we can use it in a gradient ascent algorithm:
{\small\begin{equation}
    \frac{d}{d\theta}V^\pi(s) = \frac{d}{d\theta}\sum_a \pi(a|s)Q^\pi(s,a)
\end{equation}}
By using the product rule we have that:
{\small\begin{equation}
    \frac{d}{d\theta}V^\pi(s) = \sum_a Q^\pi(s,a)\frac{d}{d\theta}\pi(a|s) + \sum_a\pi(a|s)\frac{d}{d\theta}Q^\pi(s,a)
\end{equation}}
We can now focus on the term $\frac{dQ^\pi(s,a)}{d\theta}$:
\begin{align}
    \frac{d}{d\theta}Q^\pi(s,a) = {} &\frac{d}{d\theta}\big[r(s,a) + \tau \mathbb{H}(\pi(\cdot|s)) + \gamma\sum_{s'} P(s'|s,a) V^\pi(s')\big] \nonumber \\
    = {} &\frac{d}{d\theta}\tau \mathbb{H}(\pi(\cdot|s)) + \gamma\sum_{s'} P(s'|s,a) \frac{d}{d\theta}V^\pi(s')
\end{align}
We can substitute the last equation in our result from the product rule expansion:
{\small \begin{align}
        \frac{d}{d\theta}V^\pi(s) = {} & \sum_a Q^\pi(s,a)\frac{d}{d\theta}\pi(a|s) + \sum_a\pi(a|s)\bigg[\frac{d}{d\theta}\tau \mathbb{H}(\pi(\cdot|s)) + \gamma\sum_{s'} P(s'|s,a) \frac{d}{d\theta}V^\pi(s')\bigg] 
\end{align}}
We can use the fact that $\frac{d}{d\theta}\pi(a|s) = \pi(a|s)\frac{d}{d\theta}\log\pi(a|s)$ to simplify some terms:
\begin{align}
    \frac{d}{d\theta}V^\pi(s) = {} & \sum_a \pi(a|s)\bigg[ Q^\pi(s,a)\frac{d}{d\theta}\log\pi(a|s)+ \frac{d}{d\theta}\tau \mathbb{H}(\pi(\cdot|s)) \bigg] + \sum_a\pi(a|s)\gamma\sum_{s'} P(s'|s,a) \frac{d}{d\theta}V^\pi(s') 
\end{align}
We can now consider the term $Q^\pi(s,a)\frac{d}{d\theta}\log\pi(a|s) + \frac{d}{d\theta}\tau \mathbb{H}(\pi(\cdot|s))$ as a ``cumulant'' or augmented reward $\hat{r}(s,a)$. Let us define $r^\pi(s)=\sum_a \pi(a|s)\hat{r}(a,s)$ and $r^\pi$ the vector form containing the values $r^\pi(s)$ and $g^\pi$ the vector form of $\frac{d}{d\theta}V^\pi$ for each state. We also define $P^\pi(s',s) = \sum_a\pi(a|s)\sum_s' p(s'|s,a)$ as the transition matrix representing the probability of going from $s\rightarrow s'$. If we write everything in matrix form we get that:
\begin{equation}
    g^\pi = r^\pi + \gamma P^\pi g^\pi
\end{equation}
This is a Bellman equation and we can solve it using the matrix inverse:
\begin{equation}
    g^\pi = \frac{r^\pi}{(I-\gamma P^\pi)}
\end{equation}
Written explicitly this is:
{\small \begin{align}
    \frac{dV^\pi(s)}{d\theta} = {} & \sum_t \gamma^t P(s_t=s|s_0) \sum_a \pi(a|s)\bigg[Q^\pi(s,a)\frac{d}{d\theta}\log\pi(a|s) + \frac{d}{d\theta}\tau \mathbb{H}(\pi(\cdot|s)) \bigg]
\end{align}}
To get the correct loss, we extract the term corresponding to $s_0$:
{\small \begin{align}
    e_{s0}^T(I-\gamma P^\pi)^{-1}\sum_a \pi(a|s)\bigg[ & Q^\pi(s,a)\frac{d}{d\theta}\log\pi(a|s) + \frac{d}{d\theta}\tau \mathbb{H}(\pi(\cdot|s)) \bigg]
\end{align}}
We make this loss suitable for automatic differentiation by placing a ``stop gradient'' in the appropriate locations:
{\small \begin{align}
& e_{s0}^T(I-\gamma P^\pi)^{-1}  \sum_a 
STOP(\pi(a|s))\bigg[\log\pi(a|s) STOP(Q^\pi(s,a)) + \tau \mathbb{H}(\pi(\cdot|s)) \bigg]
\end{align}}
The code that implements the above loss is provided here: \href{https://goo.gl/D3g4vE}{https://goo.gl/D3g4vE}

\subsubsection{\textsc{REINFORCE} gradient estimator}
\label{asec:reinforce}

In most environments, we do not have access to the exact transition and reward dynamics needed to calculate $d^\pi(s)$. Therefore, the gradient of $\Oh_{ER}$, given by the policy gradient theorem,
\begin{equation}
\nabla_\theta \Oh_{ER}(\theta) = \int_{s}d^{\pi_\theta}(s) \int_a\nabla_\theta\pi_\theta(a|s)Q^{\pi_\theta}(s,a) \text{d}a\text{d}s    
\end{equation}
cannot be evaluated directly. The \textsc{Reinforce} \citep{williams1992simple} estimator is derived by considering the fact that $\nabla_\theta \pi_\theta(s|a) = \pi_\theta(s|a)\nabla_\theta\log\pi_\theta(s|a)$, allowing us to estimate $\nabla\Oh_{ER}$ using Monte-Carlo samples:
\begin{equation}
    \nabla \Oh_{ER}(\theta)\approx \frac{1}{N}\sum_{n}\sum_{s^n_t,a^n_t\sim\pi} \nabla\log\pi(a^n_t|s^n_t) G_t
\end{equation}
where $G_t$ is the Monte-Carlo estimate for $Q^\pi(a_t, s_t)$. We use $N=128$ and the batch average baseline to reduce variance in the estimator and to account of the confounding variance reduction effect of $\sigma$ in the case of Gaussian policies \citep{zhao2011analysis}.


\subsection{Open source implementation details and reproducibility instructions}
\subsubsection{Objective function analysis demonstration}

We provide a demonstration of our random perturbation method (Section~\ref{appr:pertubations}) in a Colab notebook using toy landscapes as well as FashionMNIST \citep{xiao2017fashion}\footnote{Landscape analysis demo:
\href{https://goo.gl/nXEDXJ}{https://goo.gl/nXEDXJ}
}.

\subsubsection{Reinforcement Learning Experiments}
Our Gridworld is implemented in the easyMDP package\footnote{easyMDP \href{https://github.com/zafarali/emdp}{https://github.com/zafarali/emdp}
} which provides access to quantities needed to calculate the analytic gradient. The experiments are reproduced in a Colab with embedded instructions\footnote{
Exact policy gradient experiments: 
\href{https://goo.gl/D3g4vE}{https://goo.gl/D3g4vE}.
}.

Our high dimensional experiments used the Hopper-v1, Walker2d-v1 and HalfCheetah-v1 continuous control environments from OpenAI Gym \citep{brockman2016openai} based on Mujoco \cite{todorov2012mujoco}.  The REINFORCE algorithm is implemented in Tensorflow Eager\footnote{Algorithm:
\href{https://goo.gl/ZbtLLV}{https://goo.gl/ZbtLLV}.
}\footnote{Launcher script:
\href{https://goo.gl/dMgkZm}{https://goo.gl/dMgkZm}.
}.  Learning curves are generated based on the deterministic evaluation $\sigma=0$ to ensure policies trained using different standard deviations can be compared. Evaluation rollouts are independent of trajectories collected during training \citep{khetarpal2018re}.

To do thorough objective function analysis, it is necessary to store the parameters of the model every few updates. Once the optimization is complete and parameters have been obtained we provide a script that does linear interpolations between two parameters\footnote{Interpolation experiments: 
\href{https://goo.gl/CGVPvG}{https://goo.gl/CGVPvG}}. Different standard deviations can be given to investigate the objective function for policies with different amounts of entropy.

Similarly, we also provide the script that does random perturbations experiment around one parameter\footnote{Random perturbation experiments:
\href{https://goo.gl/vY7gYK}{https://goo.gl/vY7gYK}}. To scale up and collect a large number of samples, we recommend running this script multiple times in parallel as evaluations in random directions can be done independently of one another. We used $\approx1000$ evaluations per parameter vector. Each evaluation used 512 rollouts.

We also provide a small library to create plots that can easily be imported into a Colab\footnote{Analysis tools:
\href{https://goo.gl/DMbkZA}{https://goo.gl/DMbkZA}
}.

\subsection{Limitations of the Analysis}
In this section we describe some limitations of our work:
\begin{enumerate}
    \item{The high entropy policies we train are especially susceptible to over-reliance on the fact that actions are clipped before being executed in the environment. This phenomenon has been documented before in \cite{chou2017improving,fujita2018clipped}. Beta policies and TanhGaussian policies are occasionally used to deal with the boundaries naturally. In this work we chose to use the simplest formulation possible: the Gaussian policy. In the viewpoint of the optimization problem it still maximizes the objective. Since all relevant continuous control environments use clipping, we were careful to ensure our policies were not completely clipped in this work and that $\sigma$ was always smaller than the length of the window of values that would not be clipped. We do not expect clipping to have a significant impact on our observations with respect to smoothing behaviours of high entropy policies.}
    \item{Recent new work \citep{ilyas2018deep} has shown that in the sample size we have used to visualize the landscape, kinks and bumps are expected but get smoother with larger sample sizes. Though our batch size is higher than most RL methods but not as high as \cite{ilyas2018deep}, it captures what day-to-day algorithms face. We were careful to ensure our evaluations has a small standard error.}
\end{enumerate}


\end{document}